\pdfoutput=1
\documentclass[10pt,a4paper,twocolumn]{article}
\usepackage[T1]{fontenc}
\usepackage[utf8]{inputenc}
\usepackage{mathptmx}
\usepackage[scaled=.92]{helvet}
\usepackage{amsmath,amssymb}
\usepackage[a4paper,left=18mm,right=18mm,top=19mm,bottom=20mm,columnsep=7mm]{geometry}
\usepackage{graphicx,booktabs,tabularx,array,microtype}
\usepackage[table]{xcolor}
\usepackage[numbers,sort&compress]{natbib}
\usepackage{xurl}
\usepackage{caption,titlesec,fancyhdr,flushend}
\usepackage{placeins}
\usepackage{cuted}
\usepackage[colorlinks=true,citecolor=Link,linkcolor=Link,urlcolor=Link,breaklinks=true]{hyperref}
\definecolor{Ink}{HTML}{1D394D}
\definecolor{Link}{HTML}{176F7A}
\definecolor{Light}{HTML}{EEF3F7}
\titleformat{\section}{\large\bfseries\raggedright\color{Ink}}{\thesection}{.6em}{}
\titleformat{\subsection}{\normalsize\bfseries\raggedright}{\thesubsection}{.5em}{}
\titlespacing*{\section}{0pt}{2.2ex plus .5ex minus .2ex}{.9ex plus .2ex}
\titlespacing*{\subsection}{0pt}{1.8ex plus .4ex minus .2ex}{.7ex plus .2ex}
\newcolumntype{Y}{>{\raggedright\arraybackslash}X}
\newcommand{\doi}[1]{\href{https://doi.org/#1}{\nolinkurl{#1}}}

\newcommand{\OC}{\mathrm{orthogonally\_confirmed}}
\hypersetup{pdftitle={Embodied AI for Scientific Discovery: From Physical Action to Scientific Evidence},pdfsubject={Closed-loop AI for science, autonomous experimentation and reproducible evidence}}
\begin{document}

\twocolumn[{

\noindent{\sffamily\bfseries\small\color{Link} REVIEW ARTICLE}\par\vspace{7pt}

{\raggedright\fontsize{21}{24}\selectfont\bfseries\color{Ink}Embodied AI for Scientific Discovery: From Physical Action to Scientific Evidence\par}
\vspace{8pt}
{\fontsize{10.5}{13}\selectfont
Wenhao Xu\textsuperscript{1},
Changwei Wang\textsuperscript{2},
Shunpeng Chen\textsuperscript{3},
Yukun Song\textsuperscript{3},
Sifan Zhou\textsuperscript{4},
Kaize Luo\textsuperscript{5},
Shibiao Xu\textsuperscript{3},
Yuehua Zhao\textsuperscript{6,*},
and Rongtao Xu\textsuperscript{7,*}\par
}
\vspace{6pt}
{\footnotesize\color{black!80}
\textsuperscript{1}Zhengzhou Police University, Zhengzhou, China\quad
\textsuperscript{2}Qilu University of Technology (Shandong Academy of Sciences) and Shandong Computer Science Center, Jinan, China\\
\textsuperscript{3}School of Artificial Intelligence, Beijing University of Posts and Telecommunications, Beijing, China\quad
\textsuperscript{4}Southeast University, Nanjing, China\\
\textsuperscript{5}Southern University of Science and Technology, Shenzhen, China\quad
\textsuperscript{6}College of Electrical Engineering, North China University of Science and Technology, Tangshan, China\\
\textsuperscript{7}Institute of Automation, Chinese Academy of Sciences, Beijing, China\quad
\textsuperscript{*}Corresponding authors: yuehua@ncst.edu.cn, xurongtao2022@gmail.com\par
}

\vspace{13pt}

\noindent\textbf{Abstract}\quad Scientific experiments are increasingly selected, executed and interpreted by learned systems that act through physical interfaces. Robot learning builds general physical competence, whereas AI for science chooses which experiment to run, and each reports a different endpoint. This review treats their interface as one problem, embodied AI for science, where the experiment itself is partially observed: specimen state, apparatus state and the measurement process are latent, actions are costly and often irreversible, and missingness can depend on execution and outcome. We organize the literature by three closures that operate at different points: operational closure (the experiment realized is the one intended), epistemic closure (the result changes a belief or the next experiment) and evidential closure (the record supports a claim). We read each workflow through a record-linked evidence cycle---Record, Diagnose, Calibrate, Confirm---as analytic functions with feedback. A taxonomy crosses these closures with six delegated decisions and five embodiment classes. A formulation separates commanded, realized and observed conditions and treats diagnosis as an action with explicit resource, risk and sample-change costs. We compare experiment-selection methods by terminal decision, review perception, language and protocol compilation, visuomotor policies, world models, recovery and transfer by what each must emit as evidence, and compare chemistry, materials, biology, microscopy and field workflows by what their records establish. An appendix provides a structured evidence map of 28 workflows, 26 physical and two software-boundary: it categorizes physical sample routing, fixed-inline platforms, and software-boundary systems, and characterizes reported validation pathways alongside claim-level confirmation requirements. We propose evaluation that separates actions, valid experiments, campaigns and confirmed claims, and state open problems as controlled comparisons with measurable failure conditions. This structured narrative review connects experimental decisions, realized conditions and claim assessment, and derives criteria for method comparison and prospective evaluation.\par

\vspace{8pt}

\par

\vspace{12pt}\noindent\color{Ink}\rule{\textwidth}{.5pt}\vspace{10pt}

}]

\thispagestyle{plain}

\section{Introduction}\label{sec:introduction}

Scientific experiments are increasingly selected, executed and interpreted by learned systems that perceive and act through physical interfaces: robotic workcells, flow reactors, microscopes, beamlines and field vehicles. Two research communities build such systems with different aims. Robot learning builds general physical competence through vision-language-action (VLA) policies \citep{zhao2023aloha,pi0}, learned world models \citep{hafner2025dreamerv3,yang2024unisim} and language-model planners \citep{ahn2023saycan,liang2023codeaspolicies}. AI for science builds systems that choose which experiment or observation to make next, from robot scientists and self-driving laboratories (SDLs) to adaptive instruments and language-model agents \citep{king2009,burger2020}. Each community reports a different endpoint: task success for the first, a measured scientific response for the second. This review treats the interface between them as one problem, which we call \emph{embodied AI for science}, and asks what each side must supply for the other to be useful.

\paragraph{Problem.}
We define embodied AI for science as any architecture where a learned or optimized decision procedure couples, through a physical interface, to sensing or acting on an experimental specimen, and where feedback from that interface informs subsequent decisions. Embodied systems encompass diverse physical forms, including robotic workcells, flow reactors, automated microscopes, beamlines, and field rovers. Adaptive acquisition allocates measurements, while intervention deliberately changes specimen or process state; acquisition can also perturb the specimen through dose, contact or sampling. A foundational modeling requirement connects these communities: \emph{the experiment is itself partially observed}. Partial observability, costly interaction, and uncertain rewards characterize both robotics and physical experimentation. In laboratory environments, an observed signal variation can stem from intrinsic specimen dynamics, apparatus drift, or measurement artifacts, so attribution is not automatic. Actions consume finite materials and exhibit physical irreversibility, while response admissibility depends jointly on experimental design and execution fidelity. A rigorous experimental model must therefore capture joint uncertainty over latent scientific phenomena and the physical conditions realized during execution.

\begin{figure*}[t]
\centering
\includegraphics[width=\textwidth,keepaspectratio]{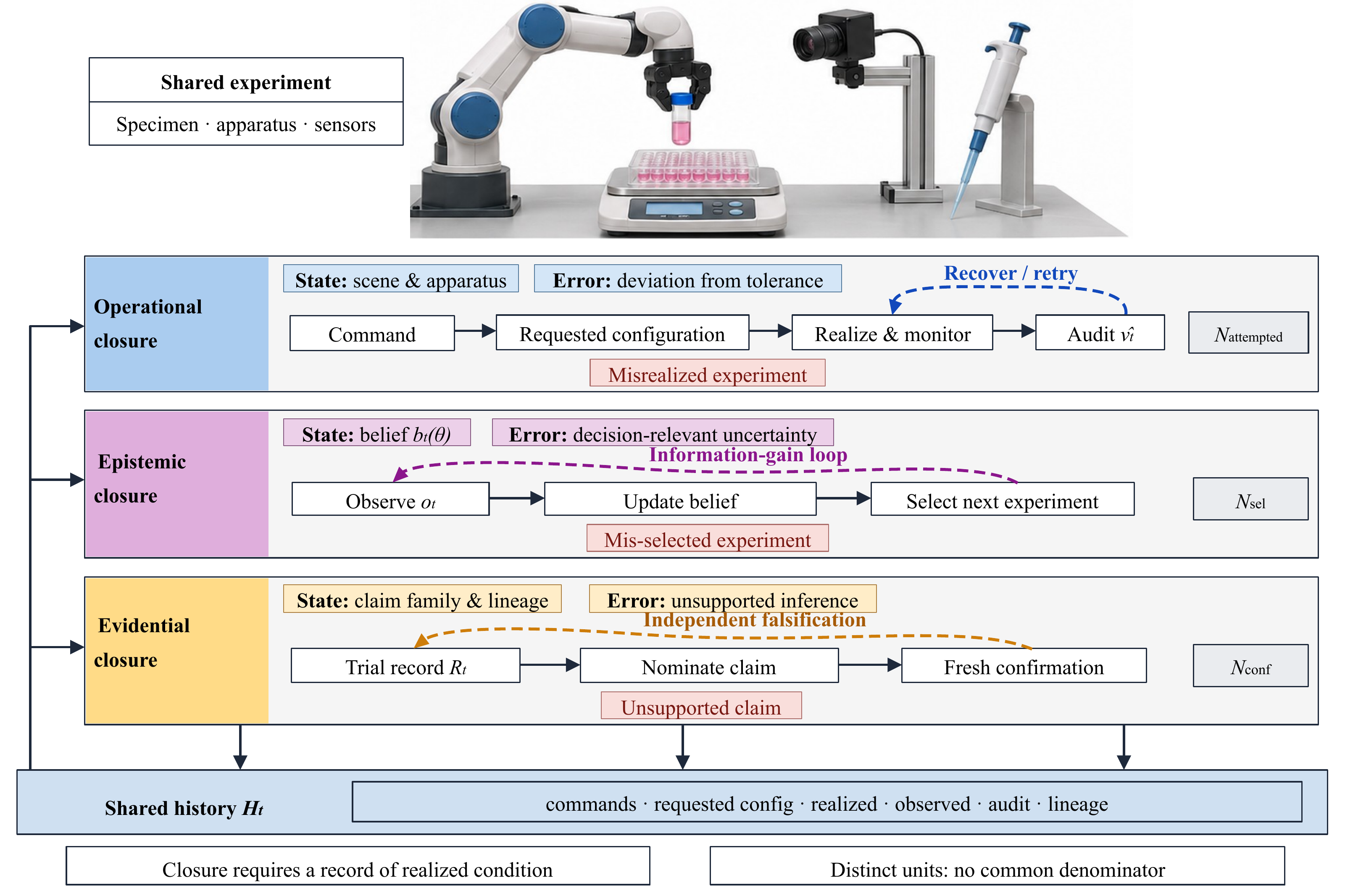}
\caption{Three closures around one shared physical experiment. Each row pairs a closure with its realized state, a three-step loop with an error feedback arc, its characteristic failure and the reporting unit it counts. Operational closure compares commanded with realized conditions and recovers from tolerance violations; epistemic closure updates belief from observations and chooses the next experiment; evidential closure turns trial records into candidate claims that need fresh independent confirmation. All three loops read the shared experiment history at the base. The reporting units $N_{\mathrm{attempted}}$, $N_{\mathrm{sel}}$ and $N_{\mathrm{conf}}$ shown here are distinct, and $N_{\mathrm{valid}}$ (Section~\ref{sec:evaluation}) is a further unit; no common denominator or one-to-one mapping across trial and claim units is implied. The diagram is conceptual and reports no measured data.}
\label{fig:dual-loop}
\end{figure*}

\paragraph{Three closures.}
We organize the literature by the feedback relation a method serves (Figure~\ref{fig:dual-loop}). \emph{Operational closure} connects the intended protocol to the realized experiment through perception, protocol compilation, visuomotor control, failure detection and recovery, with teleoperation as a human fallback. \emph{Epistemic closure} connects observations to a scientific belief or the next experiment through Bayesian optimization (BO), active learning, Bayesian experimental design (BED), sequential control and language-model planners. \emph{Evidential closure} connects the record to a claim through realized conditions, measurement validity, specimen lineage, calibration and independent confirmation. Achieved closure requires the corresponding measurements and decision rules; the feedback relation alone does not establish it. The loops fail differently. An operational failure changes the intervention or acquisition; an epistemic failure misallocates experiments relative to the question; an evidential failure leaves an inference unsupported by its record or follow-up. These failures overlap, but require different observations to resolve. A single scalar reward mixes them: a misrealized experiment and a misselected experiment both lower it, so the reward cannot say which loop to repair. A system tuned on such a reward can improve its throughput while its evidential record stays unchanged, and two systems with the same reward can fail for different reasons.

\paragraph{Evidence integrity as the organizing metric.}
We adopt evidence integrity as the primary metric for comparing physical autonomous workflows. Candidate screening speed, algorithmic ranking, and manipulation task completion reflect operational competence, not scientific validity. While simulation and software agent benchmarks \citep{f2024scienceagentbench,f2024corebench} evaluate planning, code execution, and computational reproduction, physical scientific claims require verified specimen preparation, qualified analytical measurement, and independent confirmatory testing. In addition, scripted platforms reach high evidentiary rigor when physical execution and analytical validation are tightly coupled, even with little adaptive decision-making. Because operational throughput and scientific evidence follow distinct objectives, systems must be evaluated by the depth of evidence their experimental records substantiate, not a single acceleration factor.

\paragraph{Where the frontier stands.}
Generalist VLA policies combine a vision-language trunk with diffusion or flow-matching action experts, while hierarchical designs separate deliberation from execution \citep{pi0,shi2025hirobot}. Reinforcement-learning fine-tuning improves task reliability in tested settings \citep{chen2025conrft}; world models are also evaluated for functional use as data engines and policy evaluators, not only for visual quality \citep{shang2026worldarena}. On the science side, language models enter experiment selection as proposal and prior sources \citep{liu2024llambo}, learned surrogates and amortized design policies move computation offline \citep{muller2023pfns4bo,foster2021dad}. Each shift advances a specific operational or epistemic loop. Connecting these capabilities to reproducible discovery requires explicit accounting of realized physical conditions, as formalized in the three closures below.

\paragraph{Research questions.}
The review is organized around four questions.
\emph{RQ1: What decisions does a physical workflow delegate, and which method fits each?} Property improvement, response mapping, hypothesis discrimination and adaptive observation have different terminal decisions and therefore different selection rules (Sections~\ref{sec:framework} and~\ref{sec:mf-methods}).
\emph{RQ2: What must a model of the experiment represent?} Commands do not determine specimen state or what an instrument observes, so intended, realized and observed conditions, validity and non-random missingness have to be modeled (Section~\ref{sec:formulation}).
\emph{RQ3: What do current embodied capabilities and physical campaigns establish?} Perception, policies, world models and orchestration are assessed against the experimental variable they must affect, and application evidence is graded by whether it reaches execution, valid measurement or confirmation (Sections~\ref{sec:architectures} and~\ref{sec:applications}).
\emph{RQ4: How should such systems be evaluated and what remains open?} Evaluation has to separate actions, valid experiments, campaign progress and confirmed claims (Sections~\ref{sec:evaluation}--\ref{sec:priorities}).

\paragraph{Relation to existing surveys.}
Existing surveys separately cover AI for science, self-driving laboratories, agentic science and embodied intelligence; Section~\ref{sec:scope} and Table~\ref{tab:table1} locate this review among those accounts.

\paragraph{Contributions.}
\begin{enumerate}
\item A \emph{taxonomy} (Section~\ref{sec:framework}) crossing the three closures with the decisions a workflow delegates and the embodiment through which it acts.
\item A \emph{formulation} (Section~\ref{sec:formulation}) that separates commanded, realized and observed conditions, treats diagnosis as a priced action and validity as a variable distinct from evidential status, and specializes the standard observational-equivalence condition to the record distribution of an adaptive campaign, giving a sufficient condition under which two mechanisms cannot be separated by any feasible adaptive experiment within a stated nuisance model.
\item A \emph{comparison} of experiment-selection and planning methods by terminal decision, required observability and failure condition (Section~\ref{sec:mf-methods}), and of embodied-AI components by what each must emit as evidence (Section~\ref{sec:architectures}).
\item A \emph{cross-domain synthesis} grading selected chemical, materials, biological, microscopy and field workflows by how far their records reach (Section~\ref{sec:applications}), and an \emph{evaluation and reporting proposal}, which is the authors' own and not an existing standard, with testable open problems, each stated as a controlled comparison with a falsifying outcome (Sections~\ref{sec:evaluation}--\ref{sec:priorities}).
\end{enumerate}
The review develops an analytical framework, a decision-theoretic formulation, and reporting standards to compare representative physical workflows. Section~\ref{sec:scope} and Appendix~\ref{sec:appendix} describe literature selection and the evidence map. A cohort of 28 workflows (26 physical, two software-boundary) accompanies the synthesis (Appendix~\ref{sec:appendix}), mapping delegated decisions and evidence levels across laboratory implementations. Workflows are analyzed throughout via the record-linked evidence cycle: Record $\rightarrow$ Diagnose $\rightarrow$ Calibrate $\rightarrow$ Confirm.

\section{Scope, boundaries and historical context}\label{sec:scope}

\subsection{What is reviewed and how}\label{sec:selection}

This review synthesizes core developments across robot learning and physical experimentation, focusing on demonstrated capability boundaries within declared resource constraints. The literature corpus spans embodied intelligence, AI for science, self-driving laboratories, adaptive instruments, robot learning, experimental design, world models, and laboratory safety; records and claims are grounded in primary publisher sources, proceedings, preprints, and author-published corrections \citep{alab2023,alabcorr2026} (Appendix~\ref{sec:appendix}). Comparisons are made only within a matched endpoint and resource boundary. Where a source defines a rate with a different denominator, we report that denominator with the value; a conversion would import a definition the source did not use. Unreported capabilities are classified as unrecorded, not negative, because absence from a report does not show that the capability fails.

A study entered the application synthesis (Section~\ref{sec:applications}) if it had a defined scientific objective, a physical interface and feedback into execution, acquisition or experiment selection. Without a question there is no decision to assess; without an interface no physical condition is tested; without feedback the workflow does not adapt. Nonadaptive platforms, computational AI for science and general robotics are used as enabling or boundary evidence, because their results do not establish sample identity, assay validity or confirmation. Inclusion is based on the scientific objective, physical interface and feedback-driven decision; the evidence-map denominator is intentionally broader because it retains boundary cases such as software or design workflows for comparison. Within this cohort, Access is coded with categorical interface tags: X = cross-station-mobile (multi-station sample routing or cross-instrument handover); M = local-manipulation (single-station manipulation, loading, sensing or localized actuation); F = fixed-interface or instrument-local control (fixed-inline, fluidic, microplate, or instrument-local); S = software-boundary (software-only focal component); ? = unresolved. F marks instrument-local control and does not imply sample routing; X is the sample-routing tag. The two S rows are software-led proposal components (BioDiscoveryAgent, Co-Scientist) retained as boundary comparators; physical follow-up they report is tagged separately. RoboChem is a physical, fixed-inline (F) platform and is not a software row. Confirm is coded as typed evidence with the final vocabulary: O = orthogonal physical assay (computational benchmark comparisons are not counted as O); F = repeat/follow-up testing; OF = both; P = planned; N = scoped-not-reported; ? = unresolved. The cohort and its coding are defined in Appendix~\ref{sec:appendix}, which gives the analytical evidence map across the 28 workflows; per-cell provenance is maintained in the adjudication ledger.

\subsection{Enabling capabilities and demonstrated workflows}

We call a benchmark, simulation, computational method or component \emph{enabling} when its physical scientific endpoint was not measured, and a result \emph{demonstrated in a physical science workflow} when a cited experiment reports a scientific observation with a stated boundary. The label applies to the result discussed, not to the platform as a whole.

Generalist manipulation policies, world models and language planners are evaluated primarily on task completion, prediction or executable plans \citep{zhao2023aloha,hafner2025dreamerv3,ahn2023saycan}. Scientific-agent benchmarks likewise assess data-analysis tasks or computational reproduction rather than physical experiments \citep{f2024scienceagentbench,f2024corebench}. Contact representations provide another enabling capability, whose measurement obligations are assessed in Section~\ref{sec:tactile-fm}. Computational selection methods address safe exploration, unknown feasibility and language-guided proposals \citep{safeopt2015,f2025anubis,liu2024llambo}; applying them to an apparatus additionally requires qualified preparation and measurement.

The physical application synthesis spans self-driving laboratories, formulation and scattering, adaptive diffraction, beamlines, spectroscopy and field sampling (Section~\ref{sec:applications}). Field autonomy illustrates the distinction. Perseverance's Enhanced Autonomous Navigation increased average daily driving distance by an order of magnitude, with about 90\% of 32.1 km driven using it \citep{f2026perseverancenav}; this supports rover navigation, not a conclusion about a sample. PIXL recognizes points of interest in real-time compositional data and lengthens integrations there \citep{f2025pixl}; this demonstrates adaptive planetary measurement, not the claim the measurements support. Two-vehicle ocean mapping directs sampling from measured chlorophyll distributions but does not establish unbiased maps or causal control \citep{f2024multiuv}.

The two enabling families---contact-rich manipulation and computational selection---are the frontier this review treats as unresolved. Bridging these capabilities requires two metrological anchors: a contact representation calibrated in newtons against a force--torque reference across sensor wear, and a batch or language-model-guided selector whose proposed designs are evaluated across infeasible, invalid, and valid preparation states. Both comparisons require an independently measured experimental endpoint, although their sensing and selection errors differ. Sections~\ref{sec:tactile-fm} and~\ref{sec:llm-bo} state the specific observations that would close each gap.

\subsection{What constitutes embodiment in AI for science}

Embodied intelligence here denotes a decision system whose workflow senses or acts through a physical interface: a manipulator, fluidic apparatus, microscope or mobile sampler. The criterion is the evidence pathway, not robot shape. Actions change an experimental condition or an acquisition, and observations can change later decisions. Intervention changes the specimen or process; acquisition changes where, when or how it is measured. Acquisition can improve coverage without identifying causality, while intervention tests causal alternatives only when controls and measurements support that inference. A study that reports a causal effect of an intervention must therefore show the control arm and the measurement that separates the alternatives; a coverage gain reported from acquisition alone does not license that claim.

A scripted liquid handler is embodied but not adaptive when a person selects experiments; a Bayesian optimizer is adaptive but not embodied if a person executes each operation. We assess the connected system, including human decisions and actions, so that platform labels do not hide who is responsible for objectives, ambiguity, recovery or confirmation. Contact sensing makes this boundary sharper: a gripper with a force--torque sensor is embodied whether or not a learned representation consumes the signal, and what changes is the record, because a calibrated readout makes the realized condition of a contact step observable instead of inferred from a policy's success flag.

\subsection{Relationship to existing reviews}

Existing reviews cover complementary parts of this territory. Broad AI-for-science reviews address representation, prediction and generation \citep{wang2023}; the comprehensive SDL review covers algorithms, apparatus and chemical practice \citep{tom2024}; agentic-science and embodied-AI surveys cover process architectures and computation grounded in physical interaction \citep{f2025agenticsurvey,f2025embodiedsurvey}. Focused accounts address chemical laboratories and data-driven microscopy \citep{seifrid2022,morgado2024}, while performance metrics, physics-experiment design and Bayesian design provide the comparison and allocation perspectives \citep{volk2024,physics2026,chaloner1995}. We connect these accounts by tracing each workflow's operational, epistemic and evidential closures. Table~\ref{tab:table1} records the overlap.

\begin{table*}[t]
\caption{Representative reviews and the analytical connections developed here.}
\label{tab:table1}
\centering\fontsize{8.5}{10.2}\selectfont
\setlength{\tabcolsep}{5pt}\renewcommand{\arraystretch}{1.12}
\begin{tabularx}{\textwidth}{@{}YYY@{}}
\toprule
\rowcolor{Light}\textbf{Perspective} & \textbf{Main emphasis} & \textbf{Cross-index used here}\\
\midrule
Broad AI-for-science reviews \citep{wang2023} & Representations, prediction and generation & Proposal, physical realization and evidence status.\\[3pt]
\addlinespace[1.5pt]
SDL surveys \citep{tom2024} & Algorithms, apparatus and chemical practice & Scientific/operational feedback, validity and confirmation.\\[3pt]
\addlinespace[1.5pt]
LLM/agentic surveys \citep{f2025agenticsurvey} & Process stages, architectures, progress and metrics & Outputs placed within physical execution and evidence boundaries.\\[3pt]
\addlinespace[1.5pt]
Embodied AI \citep{f2025embodiedsurvey} & Computation aligned with the physical world & Capabilities mapped to scientific endpoints and validity.\\[3pt]
\addlinespace[1.5pt]
Focused perspectives \citep{seifrid2022,morgado2024} & Chemical laboratories; data-driven microscopy & Decision--action--interpretation links across domains.\\[3pt]
\addlinespace[1.5pt]
Performance metrics \citep{volk2024} & Transparent laboratory comparison & Completion, valid data, progress and confirmation distinct.\\[3pt]
\addlinespace[1.5pt]
Physics experiment design \citep{physics2026} & Configuration as optimization & Setup variables and constraints as reported.\\[3pt]
\addlinespace[1.5pt]
Bayesian design \citep{chaloner1995} & Information value and allocation & Informative experiments tied to feasibility and interpretation.\\[3pt]
\addlinespace[1.5pt]
Present review & Evidential relations across physical workflows & Cross-workflow analysis of delegated decisions, realized conditions and claim support.\\[3pt]
\bottomrule
\end{tabularx}
\par\vspace{4pt}\begin{minipage}{\textwidth}\fontsize{8}{9.7}\selectfont SDL, self-driving laboratory; LLM, large language model. The right column connects each perspective to experimental decisions, realized conditions and claim assessment.\end{minipage}
\end{table*}

\subsection{Historical context and physical interfaces}

\begin{figure*}[t]
\centering
\includegraphics[width=\textwidth,keepaspectratio]{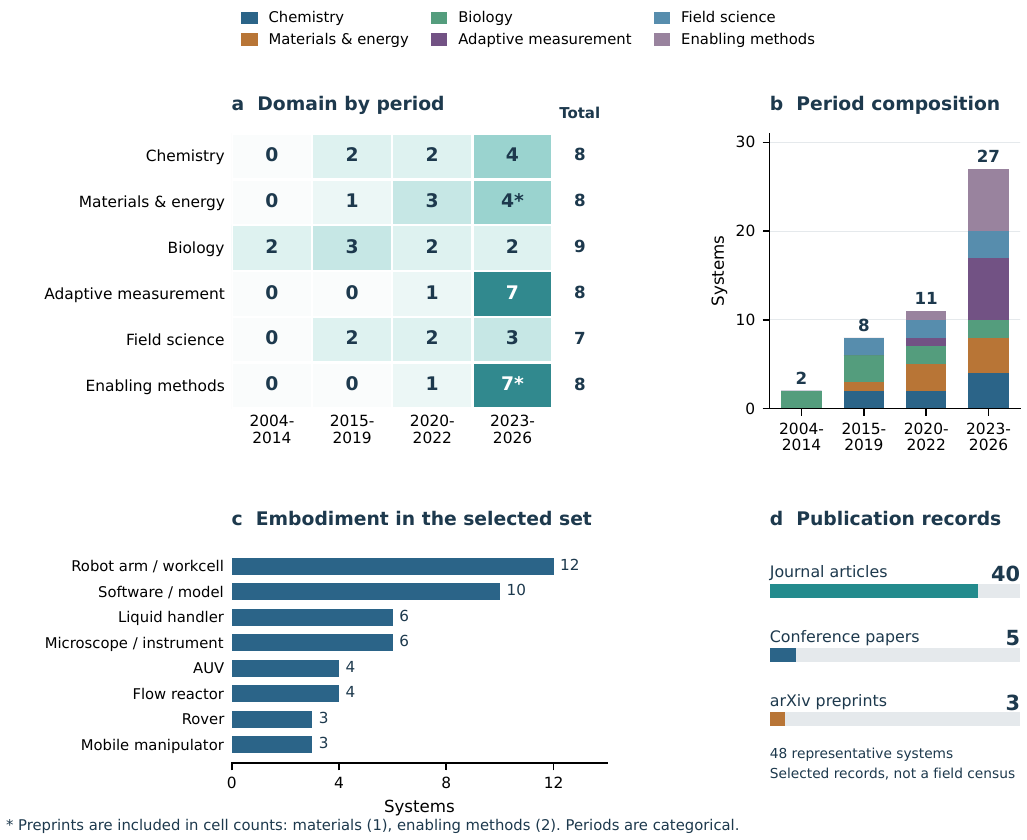}
\caption{Historical profile of the 48 representative systems. (a) Exact domain-by-period counts and domain totals; an asterisk marks cells containing already-counted arXiv preprints (one in materials and two in enabling methods). (b) Domain composition within each publication period. (c) Embodiment distribution under the source dataset's single-category coding. (d) Publication records: 40 journal articles, five conference papers and three arXiv preprints. Co-Scientist uses its published Nature version, whereas $\pi_0$ uses its arXiv record. These are distributions within the selected set; periods are categorical and do not estimate field-wide growth.}
\label{fig:historical-profile}
\end{figure*}

The reviewed systems expose recurring interfaces between decision and physical action, not a single progression. Robot scientists selected discriminating assays and programmable synthesis made procedures reusable \citep{king2004,steiner2019}. Later work joined mobile search, orthogonal analysis and language interfaces \citep{burger2020,dai2024,organa2025}. Over the same period robot learning moved from task-specific policies to action-chunked, diffusion and vision-language-action models trained on pooled multi-robot data \citep{zhao2023aloha,diffusion2023,pi0}, and world models progressed toward broader learned representations of environment dynamics, as in DreamerV3's model-based training across diverse control tasks \citep{hafner2025dreamerv3}. The two lines have developed largely independently. Several of the reviewed systems extend this interface into specific domains. In biology, the Robot Scientist Eve combined automated library screening, hit confirmation and lead generation with structure-activity learning, and used econometric modelling to show that AI-selected compounds outperform standard screening \citep{eve2015}. In materials, ARES closed the loop between carbon-nanotube growth experiments and their analysis, learning conditions that reach targeted growth rates \citep{nikolaev2016}, and the Artificial Chemist used machine-learning experiment selection with autonomous flow chemistry to tune perovskite quantum dots \citep{epps2020}. Bayesian reaction optimization \citep{shields2021} and autonomous small-angle scattering for soft-material formulation optimization \citep{f2025saxs} extend optimization to chemical synthesis and formulation measurement, while a synchrotron X-ray scattering campaign mapped a block-copolymer morphology space without direct human intervention \citep{f2023blockcopolymer}. At the beamline, an online convolutional-network analysis extracted film thickness, density and roughness within a closed reflectometry loop \citep{f2023synchrotronloop}. In field science, an AUV tracked and mapped phytoplankton biomass in three dimensions with an embedded Gaussian-process model \citep{fossum2019}, and a coordinated robot team sampled open-ocean microbial communities in a Lagrangian mode \citep{ocean2021}. On the software side, ART recommends strain designs with probabilistic predictions for successive engineering cycles \citep{art2020}, PDDLStream extends planning to black-box samplers for continuous constraints \citep{pddl2020}, and Genie learns an interactive generative environment as a world model \citep{bruce2024genie}. A 2026 preprint on A-Lab GPSS extends robotic solid-state experimentation to air-sensitive materials with agentic reasoning \citep{fei2026}. Figure~\ref{fig:historical-profile} summarizes the 48-system representative set through a domain-by-period count matrix, period composition, embodiment distribution and publication records, without inferring annual trends or a single chronology. The evidence map in Section~\ref{sec:applications} (Appendix~\ref{sec:appendix}) instead analyzes 28 workflows (26 physical, two software-boundary) at the level of delegated decisions and evidence.

\section{An organizing framework: three closures, six delegated decisions}\label{sec:framework}

We describe workflows by three axes: the \emph{closure} served (operational, epistemic or evidential), the delegated \emph{decision} (Access, Prepare, Select, Acquire, Interpret or Confirm), and \emph{embodiment}. A fourth attribute records the evidence reached through Record, Diagnose, Calibrate and Confirm. Figure~\ref{fig:decision-interfaces} locates the six decision interfaces in these coordinates; the method and application comparisons retain their scientific endpoints.

\subsection{Three closures and why they are separated}\label{sec:closures}

A laboratory loop connects measurement to the next experiment through three closures, each with its own state, error signal and failure mode, as introduced in Figure~\ref{fig:dual-loop}.

\emph{Operational closure} compares what was commanded with what was realized. Its state is the scene and apparatus (object poses, contact, liquid level and instrument telemetry); its error signal is deviation from protocol tolerance; its actions are motions, dispensing and recovery. Visuomotor policies, perception, learned simulators, failure detection and teleoperation operate here (Section~\ref{sec:architectures}). Much robot-learning work targets this loop, for which task benchmarks exist \citep{libero,li2024simpler}.

\emph{Epistemic closure} compares the current scientific belief with the information needed for a decision. Its state is a belief over scientific parameters or hypotheses; its error signal is decision-relevant uncertainty; its actions are the choice of the next experiment or measurement. Optimization, active learning, experimental design, sequential control and language-model planners operate here (Section~\ref{sec:mf-methods}).

\emph{Evidential closure} compares a scientific claim with its record and follow-up. Its state comprises the claim family $\mathfrak C_t$, trial records $R_t$, lineage, realized conditions, calibration and confirmation status; its error signal is an unsupported or under-specified inference. Diagnosis, calibration and confirmation support retaining, narrowing or revising claims (Sections~\ref{sec:formulation}, \ref{sec:evaluation} and~\ref{sec:reproducibility}).

The closures require different remedies: operational failure violates protocol tolerance; epistemic failure fails to resolve decision-relevant uncertainty; evidential failure leaves a claim unsupported by measurement or follow-up. A bare scalar reward merges the three and hides which remedy applies; it becomes usable only when each term is explicit and tied to its own validation target. A manipulation policy therefore needs a scientific interface to support selection, and an acquisition function needs execution and validity records to support a claim.

\subsection{Six delegated decisions}\label{sec:decisions}

A workflow delegates some subset of six decisions. The scientific objective is an input specified by a person or institution; it is recorded as the target of the campaign, not counted as a seventh machine decision.

\begin{itemize}
\item \textbf{Access} brings the specimen, reagent or instrument into the required physical relationship. Its unordered interface tags are X = cross-station-mobile (transport, sample routing or instrument handover); M = local-manipulation (single-station handling, loading, sensing or localized actuation); F = fixed-interface or instrument-local control (inline, fluidic or microplate); S = software-boundary (a non-spatial focal component); and ? = unresolved. X requires inter-station routing; F does not.
\item \textbf{Prepare} establishes the realized specimen condition through dispensing, mixing, heating, culturing or processing. It is where a compiled command becomes a physical condition.
\item \textbf{Select} chooses the next experiment or intervention from the history $H_t$. It is the epistemic decision in the narrow sense.
\item \textbf{Acquire} chooses how, when and where to measure, including adaptive sampling, trigger rules and signal channels. The choice is epistemic; its execution is operational.
\item \textbf{Interpret} maps a signal and its metadata to a statement about the specimen, such as phase identification, segmentation, model update or trial-status assignment.
\item \textbf{Confirm} records reported validation pathways: O = orthogonal physical assay (not computational comparison), F = repeat/follow-up, OF = both, P = planned, N = scoped-not-reported and ? = unresolved. Software-boundary rows describe the focal AI component; sourced external human validation is assessed separately. These pathway codes do not imply the formal state $\sigma_k=\mathrm{orthogonally\_confirmed}$, which requires fresh evidence independent of nomination, prespecified endpoints and documented error dependencies.
\end{itemize}

\begin{figure*}[t]
\centering
\includegraphics[width=\textwidth,keepaspectratio]{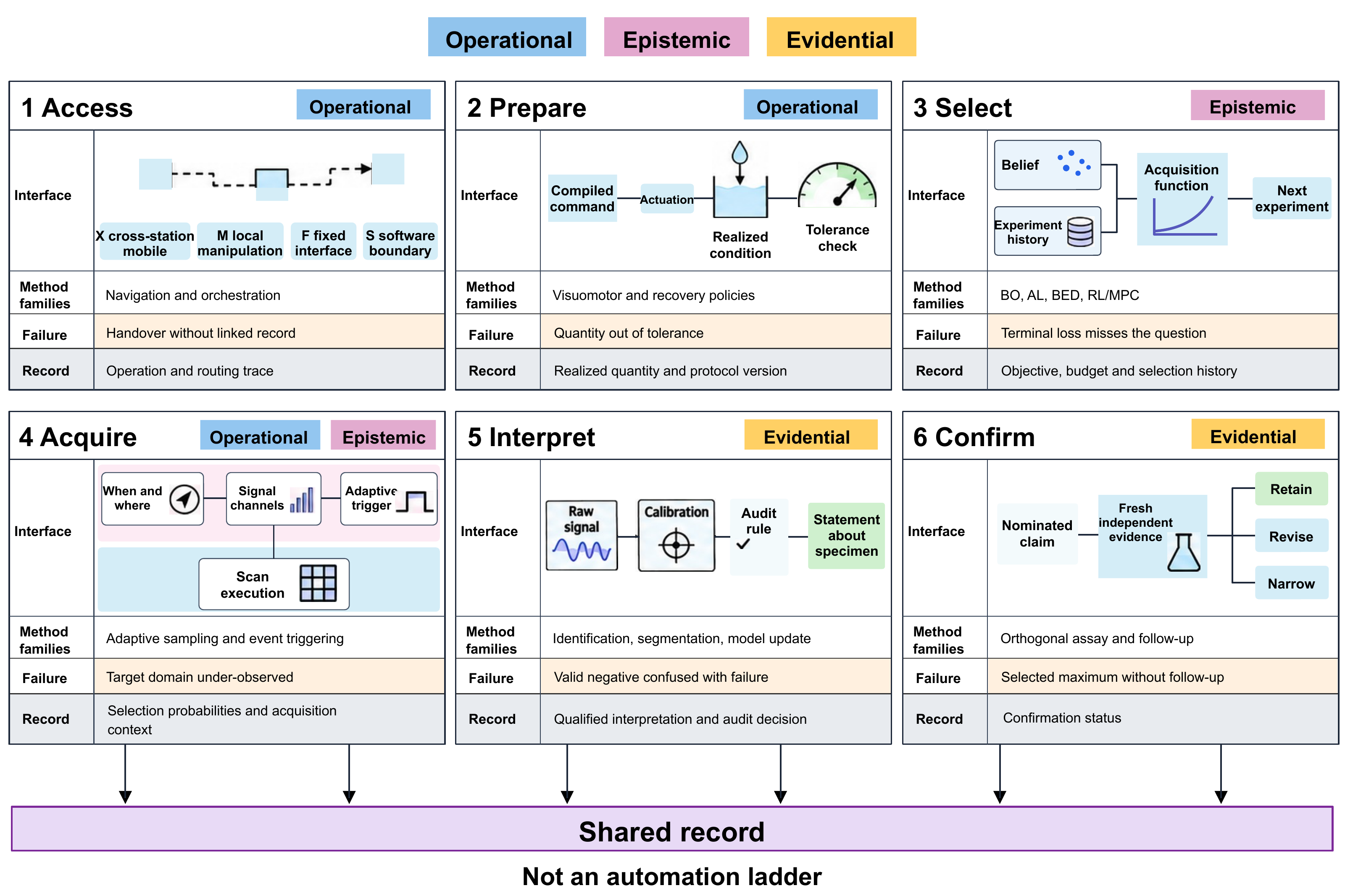}
\caption{Six delegated decisions as complementary scientific interfaces rather than an automation ladder. Each panel pairs a decision with a schematic glyph and the record it must leave; the glyphs are abstract illustrations, not measured data or depictions of particular apparatus. The bottom Shared record bar represents a unified trial ledger across all six decisions; connecting stems are schematic, and the absence of a drawn stem beneath a panel does not imply that decision omits a record. Access and Prepare primarily serve operational closure; Select serves epistemic closure; Acquire spans operational and epistemic closure; Interpret and Confirm serve evidential closure. A workflow may delegate a subset of the six decisions, and the panels are not a sequence.}
\label{fig:decision-interfaces}
\end{figure*}

Figure~\ref{fig:decision-interfaces} makes the six decisions concrete through their physical or informational interface and the evidence each must leave; its panel numbers identify responsibilities, not levels of autonomy.

Access describes physical reach; Prepare describes the operation at a station. Either fixed-interface or local-manipulation Access can accompany automated preparation. Prepare, Select, Acquire and Interpret use R (reported), A (assisted or configuration-dependent), P (planned), N (scoped-not-reported) and ? (unresolved). Confirm uses its separate pathway vocabulary above. These categorical codes describe interfaces and reported support, not levels of automation complexity.

These decisions form a dependency graph, not an automation ladder. A scripted platform can automate Prepare and Acquire while Select and Interpret remain human; a Bayesian optimizer can automate Select while a person performs Prepare. The corresponding record must preserve the routing and receiving-station check for Access; realized quantity, protocol version and exceptions for Prepare; objective, feasible set, budget and selection history for Select; acquisition context, selection probabilities and held-out acquisitions for Acquire; raw signal, calibration, processing version and audit rule for Interpret; and the indexed nomination set, independence criterion and follow-up status for Confirm. The method families are developed in Sections~\ref{sec:mf-methods} and~\ref{sec:architectures}.

\subsection{Embodiment determines which decisions are reachable}\label{sec:embodiments}

The physical interface determines which decisions a workflow can delegate. We distinguish five embodiment classes by what they can physically change and observe.

\emph{Fixed instruments} (microscopes, diffractometers, spectrometers and beamlines) primarily control acquisition; illumination, dose or probe contact can also alter the specimen. Their delegated decisions are Acquire and Interpret, and their epistemic feedback changes imaging rate, diffraction position or scan density \citep{event2022,f2023xrdphase,f2024arpes,fast2023}. \emph{Workcells with arms} add Prepare. Powder handling, dispensing, capping and plate handling are settings for robotic synthesis and bioengineering campaigns, where delivered quantities and sample handling are operational uncertainties \citep{alab2023,sample2024,longley2026robinhood,quijano2024pipetting}. \emph{Mobile manipulators} add Access across stations, enabling instrument reuse while introducing handovers whose state and identity require traceability \citep{burger2020,dai2024,lunt2024}. \emph{Flow and microfluidic platforms} implement continuous reagent-stream control and inline sensing; the realized condition is a trajectory, which makes sequential methods natural \citep{coley2019,robochem2024,alphaflow2023}. \emph{Field vehicles} (rovers and underwater vehicles) act in environments with limited reset and revisit; Access and Acquire dominate while communication and energy constrain Select and Interpret \citep{f2026perseverancenav,f2025pixl,zhang2020,f2024multiuv}.

Embodiment fixes which quantities a command can set and which an instrument can read. A tabletop manipulator can dispense, transfer and grip; a flow reactor sets residence time and concentration along a stream; a fixed microscope has no manipulation component at all. Because the command-to-condition map differs across these interfaces, transfer of a result between them preserves a scientific endpoint only when the endpoint, the state variables, the action semantics and the measurement map are all preserved. If any one changes, the same nominal protocol can yield a different realized experiment. Embodiment also sets error costs: a dose damages a specimen, a dispensing error changes concentration, and an unreachable field location ends a campaign. These costs enter the diagnostic and confirmation allocation (Section~\ref{sec:formulation}).

\subsection{Evidence reached through the record-linked evidence cycle}\label{sec:evidence-levels}

Application studies are read by the furthest evidence stage reached by their record. \emph{Record} captures the intended protocol, compiled configuration $m_t^{\mathrm{cfg}}$, commands, lineage, observations, metadata packet $m_t$ and human interventions. \emph{Diagnose} adjudicates execution and measurement anomalies, separates valid negatives from failures and assigns trial status $c_t$ with an auditable rule. \emph{Calibrate} links the realized condition and measurement to a reference, tolerance or known apparatus state, so the declared quantity has a traceable interpretation. \emph{Confirm} tests a nominated claim in $U_t^{\mathrm{claim}}$ with fresh confirmatory evidence not used to nominate the candidate, according to a prespecified endpoint and test; dependence and shared-preparation errors are recorded. The choice of test may legitimately depend on the nominated claim, and an orthogonal modality alone does not guarantee statistical independence. These stages answer different questions and have distinct accounting units: attempted trials ($N_{\mathrm{attempted}}$), audit-adjudicated valid observations ($N_{\mathrm{valid}}$), selected candidate claims ($N_{\mathrm{sel}}\equiv|U_T^{\mathrm{claim}}|$), and independently confirmed findings ($N_{\mathrm{conf}}\equiv|\{k\in U_T^{\mathrm{claim}}:\sigma_k=\mathrm{orthogonally\_confirmed}\}|$). Operational throughput measures execution efficiency, whereas valid measurement and orthogonal confirmation require metrological adjudication and independent experimental verification; each field in the trial record must explicitly identify its corresponding evidence stage and denominator.

The stages form a causal chain, and each link has a specific object. Record fixes the completed trial record $\bar R_t$: the command, the requested configuration and the observation it produced. Diagnose maps that record to the audit decision $\hat v_t$ and trial status $c_t$ through the prespecified rule $G_{\mathrm{audit}}$, so a record can enter the valid pool only if its status is valid. Calibrate ties the recorded measurement to a physical reference and, where telemetry or an independent measurement constrains the realized condition $x_t$, to that condition; this is what licenses reading $y_t$ as the response to the intended intervention, not to an unmeasured delivery. Confirm then tests a nominated claim against fresh evidence and sets its state $\sigma_k$. Each link can break independently: a valid record can rest on an uncalibrated readout, and a calibrated readout can support a claim that was never independently tested. The chain is closed only when every link is recorded, which is why the evidence stage is reported per claim, not per campaign.\par

Methods and evidence stages meet in the evidence map of Appendix~\ref{sec:appendix}, where the six decisions are columns. The map records reported delegation. Claim-level evidence is assessed through the measurements, calibration references and follow-up associated with each claim. The same cycle organizes the failure and confirmation workflow in Section~\ref{sec:evaluation}.

\subsection{Relating the framework to autonomy profiles}\label{sec:vs-autonomy}

Autonomy levels summarize responsibility or workflow coverage \citep{f2024communitysurvey,f2026autonomyfacilities}. Our profile additionally records each decision's closure, interface, evidence and human responsibility. Platforms with similar workflow coverage can differ in realized-condition records, validity adjudication, calibration or confirmation; scalar autonomy summaries do not resolve those differences.

\section{Coupling scientific inference to physical action}\label{sec:formulation}

\subsection{Three closures and a shared experimental history}

The three closures of Section~\ref{sec:framework} share a campaign history $H_t$: proposed experiments, commands, recorded realized conditions, sensor observations, audit decisions and human interventions before trial $t$. Each closure reads from this history and writes to it. Selection reads $H_t$ and writes the next command. Execution writes the realized condition, which enters the record. The record then updates two quantities: the scientific belief and the estimate of execution reliability. Claim assessment reads the record and writes the claim state $\sigma_k$. The loop closes only if the record stores the realized condition; otherwise the belief is updated from what was commanded rather than from what occurred (Figure~\ref{fig:dual-loop}).

\begin{figure*}[t]
\centering
\includegraphics[width=\textwidth,keepaspectratio]{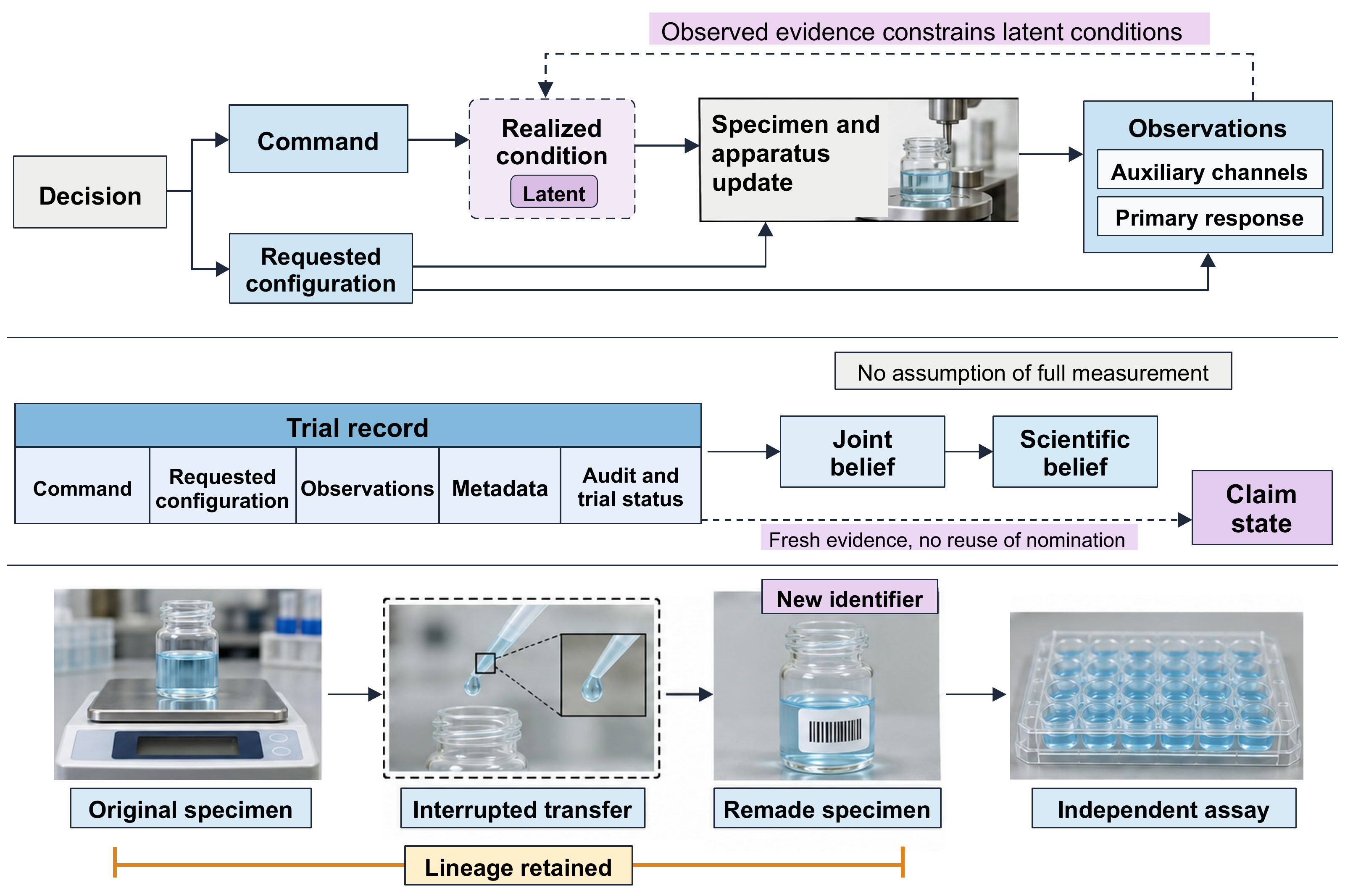}
\caption{Causal execution, observed records and specimen lineage. A decision compiles to a command and requested measurement configuration. The command acts through a latent realized condition and an updated specimen--apparatus state before observations are generated; measurement configuration may affect both state and sensing. A dashed inference link indicates that observed evidence constrains, rather than equals, the latent condition. The trial record stores commands, requested configuration, observations, metadata, audit decisions and trial status; a noisy condition estimate can be part of metadata. An interrupted transfer stays in lineage and a remade specimen receives a new identifier. Record state is distinct from claim state. Apparatus imagery is synthetic.}
\label{fig:lineage}
\end{figure*}

An experiment specification states the question, intended intervention or acquisition, specimen identity, measurement method, preconditions and interpretation criterion. Its record adds timing, delivered quantities, instrument settings, sample transformations and exceptions. Response models depend on realized conditions, making the distinction between intent and recorded execution necessary \citep{steiner2019,rauschen2024,canty2024}. Figure~\ref{fig:lineage} connects these layers to specimen and apparatus dynamics: the command acts through a latent realized condition, the trial record stores what was commanded, requested, observed and audited, and a remade specimen receives a new identifier so that lineage survives interruption.

Operational observations must constrain the variable being interpreted: a balance measures delivery, whereas a camera view of a robot near a vessel need not. An observation informs a decision only through the variable it measures, so a correlated but different measurement moves the belief about the wrong quantity. Epistemic updates depend on the task---response-surface estimation, mechanism discrimination or acquisition allocation.

The formulation fixes three interfaces: state $(s_t,\kappa_t)$, observation $(o_t,m_t)$ and decision $a_t$. Each downstream quantity is defined only through them. The belief depends on the observation interface, and the feasible set depends on the decision interface, so a change in either changes what the optimizer can learn or choose. Transition and measurement models must be estimated, and practical policies replace the exact filter with approximate inference and planning; that approximation is a property of the method and must be reported with it.

\subsection{A decision-theoretic formulation}\label{sec:decision-formulation}

At trial $t$, $s_t$ is the latent specimen state (composition, structure, concentration, viability and treatment history), and $\kappa_t$ is the apparatus and execution state (configuration, calibration, wear and environment). A high-level action compiles as $(u_t,m_t^{\mathrm{cfg}})=\Gamma(a_t,H_t)$ into a command and requested measurement configuration. The realized condition $x_t$ is latent until measured. The observation $o_t$ contains acquired sensor channels, including the primary response $y_t$ when available. The metadata packet $m_t$ records executed configuration, timing, lineage, calibration version, telemetry and any noisy estimate of $x_t$. Thus $m_t^{\mathrm{cfg}}$ describes the request and $m_t$ the reported execution; shared fields in $o_t$ and $m_t$ retain their dependence.

Validity is the audit decision $\hat v_t=\mathbf{1}\{\omega_t\geq\tau_{\mathrm{val}}\}$, where $\omega_t=g_{\mathrm{audit}}(o_t,m_t,b_t)\in[0,1]$ and the rule and threshold are prespecified; the adjudication map $G_{\mathrm{audit}}$ of Section~\ref{sec:partial-obs} is a distinct prespecified map from the recorded fields to the pair $(\hat v_t,c_t)$. Here $o_t=(o_t^{\mathrm{aux}},y_t^{\mathrm{rec}})$ is the recorded observation of Section~\ref{sec:partial-obs}, the same input that $G_{\mathrm{audit}}$ uses. The scientific belief $b_t$ of Eq.~\ref{eq:scientific-marginal} summarizes the pretrial history $H_t$; together with the observed record, it supplies the rule's inputs without an additional latent variable. The score represents a probability only when calibrated. Controlled fault-injection benchmarks supply a protocol-relative reference label $v_t^\star\in\{0,1\}$ for evaluating audit error (Section~\ref{sec:evaluation}). The completed record $\bar R_t=(a_t,u_t,m_t^{\mathrm{cfg}},o_t,m_t,\hat v_t,c_t)$ stores command and requested configuration explicitly, despite their deterministic compilation, so it is self-contained. The trial status is
\begin{multline}\label{eq:trial-status}
 c_t\in\{\mathrm{valid\_positive},\mathrm{valid\_negative},\mathrm{inconclusive},\\
\mathrm{execution\_failure},\mathrm{censored}\}.
\end{multline}
Let $\mathcal V=\{\mathrm{valid\_positive},\mathrm{valid\_negative},\mathrm{censored}\}$. The protocol enforces $\hat v_t=1\iff c_t\in\mathcal V$ and the response-coherence constraints of Section~\ref{sec:partial-obs}; an accepted record therefore carries an exact or interval response. A protocol without this biconditional must branch on $(\hat v_t,c_t)$, not on $c_t$ alone. Define the interpretability indicator $A_t=\mathbb 1\{\hat v_t=1\}$; $\mathcal H_t^{\mathrm{val}}$ in Eq.~\ref{eq:scientific-marginal} indexes accepted records. Trial status interprets one execution, whereas claim assessment combines records and follow-up. Quarantine, recalibration, replication and termination are subsequent decisions, retained in history. The audit can falsely accept or reject: the filter in Eq.~\ref{eq:joint_belief} enforces the deterministic rule but contains no $p(\hat v_t\mid v_t^\star)$ factor and does not correct these errors. Estimating them requires external reference adjudications; Section~\ref{sec:evaluation} operationalizes this through controlled injections.

Let $\theta$ denote the scientific parameters, $\phi$ the execution and transition parameters and $\psi$ the measurement and metadata parameters, and let $q_t$ be the joint belief over $(s_t,\kappa_t,\theta,\phi,\psi)$. Its scientific marginal is
\begin{equation}\label{eq:scientific-marginal}
\begin{aligned}
\mathcal H_t^{\mathrm{val}}&=\{R_\tau:A_\tau=1,\ \tau<t\},\\
b_t(\theta)&=\iiiint q_t(s_t,\kappa_t,\theta,\phi,\psi\mid
H_t)\,ds_t\,d\kappa_t\,d\phi\,d\psi,
\end{aligned}
\end{equation}
where $q_t$ is the full-history filter of Section~\ref{sec:partial-obs}. Every attempted record contributes its available channels and metadata, including exact readings rejected by the audit. The reporting subset $\mathcal H_t^{\mathrm{val}}$ imposes no second conditioning step. Although acceptance is a measurable function of the record and pretrial belief, conditioning on acceptance still selects a population. Distinguish the full-record parameter posterior $p(\theta\mid H_t)$, the selected outcome distribution in Eq.~\ref{eq:selection}, and the accepted-record likelihood $p_\Omega(R_t\mid e,H_t)/P_\Omega(A_t=1\mid e,H_t)$ on accepted records. The latter uses the coarsened record law, with $\Omega=(\theta,\phi,\psi)$ indexing both factors, and requires a prior to yield a posterior. For $n$ accepted records under an adaptive policy, the joint normalizer differs from a product of per-record acceptance probabilities:
\[
\begin{aligned}
&p_\Omega(R_{1:n}\mid A_{1:n}=1,\text{policy},H_0)\\
&\quad=\frac{\prod_{t=1}^n p_\Omega(A_t=1\mid R_t,e_t,H_t)}
{P_\Omega(A_{1:n}=1\mid\text{policy},H_0)}\\
&\qquad\times\prod_{t=1}^n p_\Omega(R_t\mid e_t,H_t).
\end{aligned}
\]
The denominator is the expectation of the numerator's acceptance-indicator product under the policy. For a stochastic policy, the joint law also contains $p(e_t\mid H_t)$ factors; these are parameter-free only when decisions use observed history. A policy depending on unobserved quantities requires its selection factors in numerator and denominator. Accepted-only inference combines this likelihood with the sampling-event-conditioned prior. Full-record inference instead uses Eq.~\ref{eq:joint_belief}, marginalizing nuisance parameters, not assuming them known. Execution, calibration and lineage remain conditioning context from $H_t$; the audit rule and acceptance rate are reported separately.

Claim assessment has its own state,
\begin{equation}\label{eq:claim-state}
\begin{aligned}
 U_t^{\mathrm{claim}}&=\{1,\ldots,n_t\},\\
 \mathfrak C_t&=\{(g_k,\mathcal E_k,\sigma_k)\}_{k\in U_t^{\mathrm{claim}}},\\
 \sigma_k&\in\{\mathrm{open},\mathrm{pending},\mathrm{supported},\\[-2pt]
 &\qquad\mathrm{orthogonally\_confirmed},\mathrm{unsupported}\},
\end{aligned}
\end{equation}
where $U_t^{\mathrm{claim}}=\operatorname{dom}(\mathfrak C_t)$ indexes $n_t$ claim objects. Each $g_k$ specifies an estimand or hypothesis, target population, applicability domain and acceptable error; $\mathcal E_k$ links supporting and conflicting records, selection history and shared errors. A claim is \emph{open} at nomination, \emph{pending} during required follow-up, \emph{supported} when it meets its prespecified criterion, \emph{orthogonally\_confirmed} when it additionally survives fresh confirmatory observations, and \emph{unsupported} when it fails the criterion. Confirmation requires a declared test and documented error dependencies, with no reuse of nomination observations. An independent preparation can establish replicability under the same assay, whereas remeasurement of the same preparation retains preparation-level error; an orthogonal modality alone need not remove shared errors. Prespecified criteria $\mathcal S$ define a deterministic claim-update map $\mathcal G$, so that $\mathfrak C_{t+1}=\mathcal G(\mathfrak C_t,R_t;\mathcal S)$ updates the claim family and its domain as claims are nominated. Unresolved checks remain pending and request diagnosis, recalibration, new preparation or a narrower claim; conflicting evidence can revise earlier support. Each check and decision enters the Record $\rightarrow$ Diagnose $\rightarrow$ Calibrate $\rightarrow$ Confirm cycle.

A feasible decision $a_t\in\mathcal A(H_t,B_t)$ can select an experiment, request diagnosis, choose an acquisition, or initiate recovery, repair or escalation. The feasible set encodes instrument capability, material, time, sample consumption and hard safety constraints. Modeled risk penalties below do not relax mandatory interlocks or exposure limits. A dispense can both prepare a sample and diagnose delivery if its mass is measured.

Define the pretrial decision context $\mathcal R_t=(H_t,\mathfrak C_t,B_t)$, distinct from the completed record $\bar R_t$, with remaining budget $B_t$. Suppress time as $\mathcal R=(H,\mathfrak C,B)$ and write its continuation as $\mathcal R'=(H',\mathfrak C',B')$. For terminal decision $\delta\in\mathcal D(\mathfrak C)$ and utility $u_{\mathrm{sci}}(\theta,\delta;\mathfrak C)$, campaign value includes stopping to issue a claim or continuing optimally after a priced action:
\begin{equation}\label{eq:utility}
\begin{aligned}
 V(q,\mathcal R)&=\max\Big\{\ \max_{\delta\in\mathcal D(\mathfrak C)}
       \mathbb E_{\theta\sim b(q)}\big[u_{\mathrm{sci}}(\theta,\delta;\mathfrak C)\big],\\
 &\hspace{4mm}\max_{a\in\mathcal A(H,B)}
       \mathbb E\big[V(q',\mathcal R')-\mathcal K(a)\mid q,\mathcal R,a\big]\Big\},\\
 Q(q,\mathcal R,a)&=\mathbb E\big[V(q',\mathcal R')-\mathcal K(a)\mid q,\mathcal R,a\big],\\
 a^*&=\arg\max_{a\in\mathcal A(H,B)}Q(q,\mathcal R,a),
\end{aligned}
\end{equation}
where $q$ is the full filter belief and $b(q)$ its scientific marginal. Planning uses $(q,\mathcal R)$: the marginal alone discards apparatus state and nuisance parameters, while $\mathcal R$ carries history, feasibility, claims and budget. Stop when the terminal branch is at least $\max_a Q(q,\mathcal R,a)$.
Here $\mathcal K(a)=\lambda C(a)+\rho R_{\mathrm{risk}}(a)+\chi\,\mathrm{Irr}(a)$ prices material, instrument time and human attention, modeled risk, and irreversible change, with nonnegative campaign weights. The expectation integrates transitions, observations, audit and claim updates. Since $V$ optimizes full continuation, $Q(q,\mathcal R,a)-V(q,\mathcal R)\leq0$ measures shortfall from the optimum. By contrast, the myopic stopping-score change $\mathbb E_o[S_{\log}(q^o)]-S_{\log}(q)$, where $S_{\log}(q)=\max_\delta\mathbb E_{\theta\sim b(q)}[u_{\log}(\theta,\delta)]$, gives an information-gain benchmark with cost reported separately. Other utilities encode prediction, improvement or claim assessment \citep{chaloner1995}. A finite horizon ensures finite planning; alternatively, a finite budget $B$ and minimum consumption $\varepsilon>0$ per non-stopping action bound their number by $\lceil B/\varepsilon\rceil$, provided budget cannot be replenished. Unbounded horizons require conditions such as discounting or proper stopping to ensure a well-defined value.

A diagnostic $a^{\mathrm{diag}}$ changes the available decisions and may itself resolve a claim. Write its updated belief and context as $(q^{+},\mathcal R^{+})$. Its branch charges the diagnostic cost once before optimal continuation, including stopping. We write $\mathrm{VOI}$ for the diagnostic's value, defined as that branch minus the best alternative that does not begin with the diagnostic:
\begin{equation}\label{eq:voi}
\begin{aligned}
 \mathrm{VOI}(a^{\mathrm{diag}}\!\mid q,\mathcal R)
 &=\mathbb E_{o,\mathcal R^{+}\mid q,\mathcal R,a^{\mathrm{diag}}}
   \big[V(q^{+},\mathcal R^{+})-\mathcal K(a^{\mathrm{diag}})\big]\\
 &\quad-\max\Big\{\ \max_{\delta\in\mathcal D(\mathfrak C)}
   \mathbb E_{\theta\sim b(q)}[u_{\mathrm{sci}}(\theta,\delta;\mathfrak C)],\\
 &\hspace{19mm}\max_{a\in\mathcal A(H,B)\setminus\{a^{\mathrm{diag}}\}}
   Q(q,\mathcal R,a)\Big\}.
\end{aligned}
\end{equation}
The budget $B$ restricts feasibility in resource units; $\mathcal K(a)$ prices preferences in utility units. Utility cost is subtracted once in each $Q$ branch. If a monetary budget is the sole cost accounting, omit its corresponding component from $\mathcal K$ consistently across all actions. In Eq.~\ref{eq:voi}, $Q(q,\mathcal R,a^{\mathrm{diag}})$ is compared with the best stopping or non-diagnostic first branch.

This definition separates the value of beginning with diagnosis from the value of its alternatives. It also credits a diagnostic's direct claim-level contribution: a confirmatory assay can update $\mathfrak C$ and reach the stopping branch, rather than being valued solely through the next experiment. Both branches use the same scientific endpoint; their difference can have either sign.

Non-negativity holds for a free diagnostic that leaves physical and apparatus states, budget and feasible continuations unchanged, changes claim context only through belief, and produces an outcome the decision maker may ignore. Fixed budget alone is insufficient if feasibility depends on belief or claims. Under coherent updating, $\mathbb E[q^{+}\mid q]=q$; expected utility for each fixed decision or continuation policy is affine in belief, so their maximum $V$ is convex. Convexity in $q$ alone is the relevant property only because the first condition holds: the context $\mathcal R$ is then unchanged except through $q$, so $V$ is evaluated at a fixed context. Jensen's inequality gives $\mathbb E[V(q^{+},\mathcal R^{+})]\geq V(q,\mathcal R)$, and $V(q,\mathcal R)$ in turn weakly exceeds the baseline of Eq.~\ref{eq:voi}, which omits the diagnostic branch. Sample consumption, apparatus degradation, changed feasibility or an incoherent update invalidates these conditions. Physical diagnosis therefore requires the explicit cost-benefit comparison, potentially over several later trials. A blank tests contamination, a balance tests delivery and an orthogonal assay tests identity; retrying delivery also intervenes on the specimen and its future options.

The coupled generative model uses $\phi$ for execution parameters, $\theta$ for scientific response and $\psi$ for measurement parameters. The realized condition $x_t$ is the mediating variable between the command and everything that follows: the response to a specimen depends on the condition actually realized, not on the condition that was requested. With disturbance $d_t$ and measurement noise $\eta_t$,
\begin{equation}\label{eq:generative}
\begin{aligned}
 x_t&\sim T_\phi(\cdot\mid s_t,\kappa_t,u_t,d_t),\\
 z_{t+1}&\sim T_z(\cdot\mid z_t,x_t,a_t,m_t^{\mathrm{cfg}},d_t;\phi),\\
 (o_t,m_t)&\sim M(\cdot\mid z_{t+1},x_t,a_t,m_t^{\mathrm{cfg}};\theta,\psi).
\end{aligned}
\end{equation}
The execution kernel uses command, specimen and apparatus state to generate $x_t$, constrained by telemetry or independent measurement. The joint transition $T_z$ propagates realized conditions and treatment history; shared disturbances may correlate specimen and apparatus updates. Requested configuration $m_t^{\mathrm{cfg}}$ enters both transition and observation: destructive measurement can change state, while channel choice sets detection limits, noise and metadata. Its transition dependence may be omitted when explicitly nonperturbing. The joint observation map $M$ uses scientific parameters $\theta$ and measurement parameters $\psi$, with $\eta_t$ integrated out; Section~\ref{sec:partial-obs} derives the coarsened record law. Retain its generic $\theta$ dependence unless a detector-artifact regime with a justified mixture weight supports an independent component. The audit gate alone supplies no such justification.

Disturbance $d_t$ covers gradual drift, such as catalyst fouling or thermal baseline wander, and abrupt regime shifts, such as clogging, phase separation or sensor delamination. A jump-diffusion or switching-regime $T_\kappa$ can represent both, including reversals and adversarially induced faults. A deliberately injected fault distribution must be stated separately from routine operating variability. Detector drift changes the apparatus state and likelihood; a preparation fault can change both apparatus and specimen state.

Consider a concrete numerical instantiation of delivery-error and execution-monitoring failure modes: a fixed $200\,\mu\mathrm{L}$ command yields either $200$ or $188\,\mu\mathrm{L}$ with equal probability. These volumes and probabilities are hypothetical illustrative values, not measurements reported by any published study. For this example, let the response be $y_t=1$ at $200\,\mu\mathrm{L}$ and $y_t=0$ at $188\,\mu\mathrm{L}$. A model conditioned on the command alone correctly estimates $p(y_t=1\mid u_t)=1/2$ for that executor, but it cannot separate the realized-volume response from delivery reliability or transfer it to an executor with a different delivery distribution: the response is a function of $x_t$, and $x_t$ is not a function of $u_t$ alone. Measuring $x_t$, for example by weighing the dispense, resolves this execution--response ambiguity, because the measurement of $x_t$ is what allows the command-to-response map to be factorized into delivery reliability and response. Published viscous-liquid pipetting experiments demonstrate execution-dependent delivery: aspiration and dispense rates affect transferred volume, and multi-objective Bayesian optimization of these rates achieved transfers within 5\% percentage transfer error for viscosities up to 1275 cP on the tested automated pipettes \citep{quijano2024pipetting}. The result is specific to the tested pipettes and viscosities. Adaptive imaging has the related complication that acquisition policy changes which events enter the dataset.

A no-product result can have $\hat v_t=1$ and $c_t=\mathrm{valid\_negative}$ when preparation and detection are qualified. Its negative sign is stored with the response. A response below a known detection limit can likewise be valid: the declared status $\mathrm{censored}$ is reserved for a valid record whose response was observed only as an interval $[L_t,U_t]$ set by that limit, and it contributes an interval likelihood (Section~\ref{sec:partial-obs}); it is not a label for follow-up that has not finished. The boundary is defined by the protocol, not by a universal limit-of-detection rule: a record is coded \emph{valid negative} when the declared interpretation rule certifies a negative result for the stated target under qualified preparation and measurement, whatever the sign or scale of the readout; it is coded \emph{censored} when the protocol declares that the target response is observable only as an interval. The two codes are not interchangeable, and a value below a nominal detection threshold is not automatically either one. A failed transfer may make the intended response undefined, giving $c_t=\mathrm{execution\_failure}$ with $\hat v_t=0$, and an ambiguous or saturated detector can leave only some target quantities interpretable, giving $c_t=\mathrm{inconclusive}$. Validity is therefore target- and protocol-specific. All trials remain in $H_t$: valid negatives update scientific response, failed transfers inform execution reliability, and uncertain cases trigger further checks. Audit decisions are revised with provenance when later evidence changes the interpretation.

Identifiability depends on the feasible interventions and observations. Use the complete observable record $\bar R_t$, with derived fields modeled jointly. Fix a nuisance model $\mu$ for execution, measurement and latent-state priors. Two scientific mechanisms are indistinguishable within a feasible policy class if, at every reachable history,
\begin{equation}\label{eq:equiv}
\begin{aligned}
 &p_\mu(R_t\mid H_t,a_t,\theta)\\
 &\quad=p_\mu(R_t\mid H_t,a_t,\theta')
 \quad\forall a_t\in\mathcal A(H_t,B_t).
\end{aligned}
\end{equation}
The sequential record distribution factors into these conditional predictive distributions and the policy's action probabilities. Equality at every reachable history therefore gives equal full-record distributions under every feasible adaptive policy. Replication within this class cannot separate the mechanisms. A new measurement, intervention or identifying assumption is needed; otherwise the claim remains unresolved. The criterion holds for the stated nuisance model. With unknown nuisance mechanisms, joint equivalence between $(\theta,\mu)$ and $(\theta',\mu')$ must also be examined: a changed response, short delivery and detector drift can compensate for one another. Weighing, blanks and reference specimens constrain different parts of this equivalence. Structural identifiability concerns exact equality; finite-sample discrimination additionally depends on signal separation, budget and measurement error.

\paragraph{Selection mechanisms and structured missingness.}
Trial validity and acceptance can depend on design, apparatus state and response. Dropping rejected trials selects the observed sample, while assigning every failure a low response conflates activity with executability. The metadata distinguish valid negative responses, censoring, undefined outcomes after failed preparation and uninterpretable measurements. These attributes are separate from the trial status $c_t$ and the claim state $\sigma_k$. Section~\ref{sec:partial-obs} develops the selection identity and states its estimand.

Four questions follow for any method. Which variables does it condition on---commands, realized conditions, observations or the full history? Which uncertainties does it represent? Which actions, including diagnosis and recovery, lie in its decision space? How do failed and invalid trials enter its model? These questions organize the methods comparison in Section~\ref{sec:mf-methods}, architecture interfaces in Section~\ref{sec:architectures} and evaluation criteria later in the review.

\subsection{Partial observability and the experimental data-generating process}\label{sec:partial-obs}

The variables needed to control an apparatus need not be those needed to interpret its result. Let
\[
 z_t=(s_t,\kappa_t),\qquad z_{t+1}=(s_{t+1},\kappa_{t+1}),
\]
with parameters $\Omega=(\theta,\phi,\psi)$ and compilation $(u_t,m_t^{\mathrm{cfg}})=\Gamma(a_t,H_t)$ as defined above. To distinguish raw response from recorded content, write
\[
 \bar R_t=(a_t,u_t,m_t^{\mathrm{cfg}},o_t,m_t,\hat v_t,c_t),\qquad o_t=(o_t^{\mathrm{aux}},y_t^{\mathrm{rec}}),
\]
where $o_t^{\mathrm{aux}}$ contains auxiliary channels and $m_t$ may contain a noisy estimate $\hat x_t$ of the realized condition $x_t$. The primary channel records an exact value $y_t$, censoring interval $[L_t,U_t]$, or absence $\varnothing$; $J_t$ below distinguishes missing from undefined responses. Let $Y_t$ denote the underlying response when defined, latent whenever unrecorded. The prespecified map $(\hat v_t,c_t)=G_{\mathrm{audit}}(o_t^{\mathrm{aux}},m_t,y_t^{\mathrm{rec}},b_t)$ uses the pretrial belief of Eq.~\ref{eq:scientific-marginal}; its deterministic output adds no independent measurement. For $\mathcal V=\{\mathrm{valid\_positive},\mathrm{valid\_negative},\mathrm{censored}\}$, require
\begin{equation}\label{eq:gate-status}
 \hat v_t=1 \iff c_t\in\mathcal V .
\end{equation}
Under Eq.~\ref{eq:gate-status}, $A_t=\mathbb 1\{\hat v_t=1\}=\mathbb 1\{c_t\in\mathcal V\}$; rejected records remain in history. Channel availability and coarsening depend jointly on requested configuration, detection limits and noise. Record $J_t\in\{\mathrm{exact},\mathrm{censored},\mathrm{missing},\mathrm{undefined}\}$ in $m_t=(J_t,m_t^{-J})$. Missing means an absent but defined response; undefined means the intended response was not generated. Require coherence between $G_{\mathrm{audit}}$, $J_t$ and recorded content:
(i) $A_t=1\Rightarrow J_t\in\{\mathrm{exact},\mathrm{censored}\}$;
(ii) $A_t=1\Rightarrow (J_t=\mathrm{censored}\Leftrightarrow c_t=\mathrm{censored})$ and $(J_t=\mathrm{exact}\Leftrightarrow c_t\in\{\mathrm{valid\_positive},\mathrm{valid\_negative}\})$;
(iii) $J_t=\mathrm{undefined}\Rightarrow c_t=\mathrm{execution\_failure}$;
(iv) $J_t=\mathrm{missing}\Rightarrow c_t=\mathrm{inconclusive}$.
These protocol constraints ensure accepted records have exact or interval content. Let $\mathcal W_t=(o_t^{\mathrm{aux}},m_t^{-J},J_t,Y_t)$ for a defined response, omitting $Y_t$ when undefined. Its joint law $P_\Omega(\cdot\mid z_{t+1},x_t,a_t,m_t^{\mathrm{cfg}})$ retains dependence among channels, metadata, response and availability; abbreviate this conditioning tuple, including $\Omega$, by $\cdot$. Coarsening gives $E_t=(o_t^{\mathrm{aux}},m_t,y_t^{\mathrm{rec}})=\mathsf C(\mathcal W_t)$ and the flattened record $R_t=(a_t,u_t,m_t^{\mathrm{cfg}},E_t,\hat v_t,c_t)$, without duplicate fields. Its measure is the pushforward
\[
\begin{aligned}
P_\Omega^{\mathsf C}(A\mid z_{t+1},x_t,a_t,m_t^{\mathrm{cfg}})
{}&=P_\Omega\bigl(\mathsf C(\mathcal W_t)\in A\mid z_{t+1},\\
&\qquad x_t,a_t,m_t^{\mathrm{cfg}}\bigr),
\end{aligned}
\]
for measurable record sets $A$. Choose a dominating measure $\nu$ on the coarsened record space: Lebesgue measure for exact continuous channels and for variable reported bounds $(L_t,U_t)$; counting/atomic measure for discrete statuses, availability indicators, the missing atom $y_t^{\mathrm{rec}}=\varnothing$, the undefined indicator, and protocol-fixed censoring intervals. The metadata remainder $m_t^{-J}$ includes any recorded bounds $(L_t,U_t)$ of an interval-censored reading, so the observed record carries them and their law is part of the joint law. If the bounds are random, the censored branch is a joint density in $(Y_t,L_t,U_t)$ integrated over $Y_t\in[L_t,U_t]$; if the protocol fixes them, it is a probability mass on the censoring event. The record likelihood is the Radon--Nikodym derivative
\begin{equation}\label{eq:record-likelihood}
\ell_0(E_t\mid\cdot)
=\frac{dP_\Omega^{\mathsf C}(\cdot\mid\cdot)}{d\nu}(E_t),
\end{equation}
when this derivative exists. For $J_t\neq\mathrm{undefined}$, a chain-rule representation of the full-data law is
\[
\begin{aligned}
&p_\Omega(o^{\mathrm{aux}},m^{-J},J,Y\mid z_{t+1},x_t,a_t,m_t^{\mathrm{cfg}})\\
&\quad=p(o^{\mathrm{aux}}\mid\cdot)\,p(m^{-J}\mid o^{\mathrm{aux}},\cdot)\\
&\qquad\times p(J\mid o^{\mathrm{aux}},m^{-J},\cdot)\,
p(Y\mid J,o^{\mathrm{aux}},m^{-J},\cdot),
\end{aligned}
\]
representing an exact chain-rule factorization. For $J_t=\mathrm{undefined}$ no response is generated, so the response factor is absent and the law is $p_\Omega(o^{\mathrm{aux}},m^{-J},J=\mathrm{undefined}\mid z_{t+1},x_t,a_t,m_t^{\mathrm{cfg}})$. The corresponding record-density cases, with shared fields conditioned jointly, are
\begin{equation}\label{eq:response-branches}
\ell_0(E_t\mid\cdot)=\begin{cases}
 \pi_\Omega(\mathrm{exact},y_t), & J_t=\mathrm{exact},\\[3pt]
 \displaystyle\int_{L_t}^{U_t}\pi_\Omega(\mathrm{censored},y)\,dy, & J_t=\mathrm{censored},\\[7pt]
 \displaystyle\int_{\mathcal Y}\pi_\Omega(\mathrm{missing},y)\,dy, & J_t=\mathrm{missing},\\[7pt]
 \pi_\Omega(\mathrm{undefined}), & J_t=\mathrm{undefined},
\end{cases}
\end{equation}
where the two branch densities are
\begin{align*}
\pi_\Omega(j,y)&=p_\Omega\bigl(o_t^{\mathrm{aux}},m_t^{-J},J_t{=}j,Y_t{=}y\mid z_{t+1},x_t,a_t,m_t^{\mathrm{cfg}}\bigr),\\
\pi_\Omega(j)&=p_\Omega\bigl(o_t^{\mathrm{aux}},m_t^{-J},J_t{=}j\mid z_{t+1},x_t,a_t,m_t^{\mathrm{cfg}}\bigr).
\end{align*}
Here $\mathcal Y$ is the full response support. The exact case is a density with respect to the exact-channel component of $\nu$; the censored case integrates over the observed interval; the missing case integrates the latent response; and the undefined case retains the probability of response non-generation. Each branch is a joint density or mass that retains the probability of the observed $J_t$ together with the remaining fields; the four branches partition the coarsened record space, so the record law of Eq.~\ref{eq:record-likelihood} has total mass one on that space, and no separate $p_\Omega(J_t\mid\cdot)$ factor is applied. If the observation law is not dominated by the chosen $\nu$, retain the pushforward probability measure rather than asserting a density.

The deterministic audit/status rule is applied to the \textbf{recorded} fields: $(\hat v_t,c_t)=G_{\mathrm{audit}}(o_t^{\mathrm{aux}},m_t,y_t^{\mathrm{rec}},b_t)$. Under the record dominating measure $\nu_R=\nu\otimes\#$ (counting measure for $(\hat v_t,c_t)$), the complete likelihood is, with $\cdot$ denoting the conditioning tuple as above,
\begin{equation}\label{eq:complete-record-likelihood}
\begin{aligned}
\ell(R_t\mid\cdot)
={}&\ell_0(E_t\mid\cdot)\\
&\times\mathbb 1\{G_{\mathrm{audit}}(o_t^{\mathrm{aux}},m_t,y_t^{\mathrm{rec}},b_t)=(\hat v_t,c_t)\}.
\end{aligned}
\end{equation}
The joint filtering update is
\begin{equation}\label{eq:joint_belief}
\begin{split}
q_{t+1}(z_{t+1},\Omega\mid H_{t+1})\propto{}&
 \iint p_\phi(z_{t+1},x_t\mid z_t,a_t,u_t,m_t^{\mathrm{cfg}})\\
&\quad\times q_t(z_t,\Omega)\,
 \ell(R_t\mid\cdot)\,dx_t\,dz_t,
\end{split}
\end{equation}
All likelihood factors depending on $x_t$ remain inside the integral. The kernel $p_\phi$ follows Eq.~\ref{eq:generative}, marginalizing unobserved i.i.d. disturbances and conditioning on the compiled pair $(u_t,m_t^{\mathrm{cfg}})=\Gamma(a_t,H_t)$. Persistent regimes belong in $\kappa_t$. The deterministic indicator enforces record consistency; it cannot select a $\theta$-free artifact law from the audit outcome. Such a component requires a justified regime indicator and mixture weight in $P_\Omega$. Updating $H_{t+1}=H_t\cup\{R_t\}$ and integrating $(z_t,x_t)$ leaves a posterior over $(z_{t+1},\Omega)$, with all parameters updated jointly and a normalizer over that pair.

The belief used for scientific inference is the full-record posterior of this filter; the gate is not a second conditioning step on top of it:
\begin{equation}\label{eq:joint_scientific_marginal}
 b_t(\theta)=\iiiint q_t(s_t,\kappa_t,\theta,\phi,\psi\mid
 H_t)\,ds_t\,d\kappa_t\,d\phi\,d\psi,
\end{equation}
as in Eq.~\ref{eq:scientific-marginal}. The reporting subset $\mathcal H_t^{\mathrm{val}}$ identifies adjudicated-interpretable records but excludes no available channel from this full-record likelihood, including exact readings rejected by the audit.

For accepted-only analysis, the conditional record likelihood is $p_\Omega(R_t\mid A_t=1,e,H_t)=p_\Omega(A_t=1\mid R_t,e,H_t)p_\Omega(R_t\mid e,H_t)/P_\Omega(A_t=1\mid e,H_t)$. Its numerator retains the exact or interval likelihood of Eq.~\ref{eq:response-branches}; the acceptance indicator equals one on an accepted record. Missing and undefined responses are excluded by the protocol constraints. Section~\ref{sec:formulation} gives the joint accepted-campaign likelihood and its sampling-event-conditioned prior. These normalizers belong to selected-sample inference, not to the full-record posterior. Audit-defined acceptance also differs from reference validity when the audit errs; controlled injections or external reference adjudications are needed to estimate that discrepancy (Section~\ref{sec:evaluation}). Selection statements below condition on a fixed design and history; the full history retains irreversible transformations and human interventions for diagnosis and claim assessment.

Measurements are generally non-injective: many states produce the same observation. Transparent and reflective labware, meniscus curvature, gripper occlusion and colorimetric interference can map distinct states to similar images or signals. Tactile and force sensing add contact channels that vision lacks. Cross-sensor touch representations transfer across several visuo-tactile sensors \citep{higuera2025sparsh,feng2025anytouch}, and force-aware policies treat six-axis force as an input \citep{yu2025forcevla}. These results identify sensing capabilities; the required contact variable and sample-identity condition remain properties of the scientific operation. A vision-language module can localize and inspect laboratory ware \citep{zhou2025lira}; localization, identity and assay validity enter the record as separate observations and decisions.

Diagnosis chooses the check that resolves the relevant uncertainty at acceptable cost (Eq.~\ref{eq:voi}). Remeasurement can probe detector error but cannot resolve doubtful preparation; conversely, repeating a synthesis wastes material when calibration suffices. A retry can alter the specimen, so recovery updates both the transition kernel and claim lineage. Record the check, its reference and its resulting decision.

Let $e$ denote an intended trial design, $r_t\equiv A_t$ its adjudicated-interpretability indicator, and $\widetilde y$ the response defined under that experiment. At a fixed design and history, the selected conditional distribution satisfies
\begin{equation}\label{eq:selection}
\begin{aligned}
p(\widetilde y\mid e,r_t=1,H_t)
={}&\frac{p(r_t=1\mid\widetilde y,e,H_t)}{P(r_t=1\mid e,H_t)}\,
p(\widetilde y\mid e,H_t),\\
P(r_t=1\mid e,H_t)={}&\int p(r_t=1\mid y,e,H_t)p(y\mid e,H_t)\,dy.
\end{aligned}
\end{equation}
If acceptance increases with $\widetilde y$ at fixed $e$ and $H_t$, the accepted conditional mean is at least the attempted-design mean, and strictly larger when the response varies and acceptance is not constant over its support. A campaign mean additionally depends on the policy's design distribution. Preparation failures with undefined responses require an explicit target population before cross-design comparison. Equation~\ref{eq:selection} normalizes the accepted-subpopulation distribution; Eq.~\ref{eq:joint_belief} instead conditions on all attempted records. Selection normalization is required for the former target, not as an extra correction to the latter. Preserve $\hat v_t$, $c_t$ and $\sigma_k$ separately, with coarsening handled by Eq.~\ref{eq:response-branches}.

Non-random missingness changes the estimand learned by a response surface. A model trained only on accepted trials describes the selected executable regime. Imputing every failure as a low response estimates a mixture of scientific response and executability. The direction and size of the resulting bias depend on the response, composition, apparatus state, policy-induced design distribution and the definition of the target population. An independently audited sample, randomized checks or explicit execution and response models can identify parts of this dependence. Report which outcomes were modeled, discarded or left unresolved, and preserve the evidence stage reached by each record.

The formulation marks the limit imposed by observability. If two candidate mechanisms are observationally equivalent under the feasible actions and measurements (Eq.~\ref{eq:equiv}), replication within that design class leaves the ambiguity intact. An intervention or measurement outside the equivalence class, an identifying assumption with stated scope or a narrower unresolved claim is then required. Diagnosis and measurement design therefore belong in the method comparison as decision variables, alongside apparatus sensing.

These distinctions suggest concrete tests. A representation comparison should withhold preparation batches, not only images. A diagnostic policy should be assessed on matched campaigns with controlled delivery faults or instrument drift, under the same total cost and sensor access. A missing-data model should be tested against an independent audit subset selected without the optimizer's acceptance rule. Such comparisons evaluate whether state estimation improves scientific decisions while retaining execution, calibration and confirmation records.

\subsection{Scientific objectives and the meaning of discovery}

A property-optimization campaign asks which feasible design performs well under a defined assay. The endpoint may be yield, conductivity, stability or another measured response. Its scientific value depends on reproducible improvement and on the connection between the assay and the intended use. Optimizing a proxy should therefore be followed by tests of the application-level property when those differ, as illustrated by electrolyte optimization followed by cell testing \citep{dave2022}.

A hypothesis-directed campaign asks which observations separate candidate explanations. It may choose a condition with modest expected performance if the alternatives predict different responses there. Biological robot scientists and closed-loop model development provide concrete precedents \citep{king2004,king2009,coutant2019}. Mechanistic interpretation remains conditional on the interventions, controls and alternatives tested. AutoSciLab illustrates the distinction between distilling interpretable latent variables and optimizing a single scalar objective \citep{f2025autosclab}.

An adaptive-observation campaign chooses where, when or how to measure. Its objective may be identifying a transition, capturing a transient event or characterizing a moving environmental feature \citep{kusne2020,event2022,zhang2020}. Because policy changes the sampled data, performance includes coverage and uncertainty alongside reconstruction accuracy. These objectives can coexist within one platform, and a change of objective changes the sampling rule, so the change must be recorded in the trial history, not inferred from the output.

\subsection{Autonomy as a profile}

We describe autonomy through four questions: who chooses the scientific objective, who chooses the next experiment or acquisition, who handles physical execution and recovery, and who determines whether evidence supports the claim? These questions identify responsibilities. A platform can automate transport and selection while depending on expert product identification; another can choose measurement locations under a fixed mission objective.

Other schemes assign levels to laboratory workflows or large-scale facilities \citep{f2024communitysurvey,f2026autonomyfacilities}; those levels summarize workflow coverage, while the four questions identify responsibility for each scientific decision. A dexterous manipulator can expand the accessible experiment space; improved inference can reduce measurement cost. Each source changes a different term in the cost of a result: a dexterous manipulator changes which experiments are reachable, while better inference changes how many measurements each result costs. A comparison that credits one with the other's gain attributes a capability change to the wrong component, so each requires a baseline that holds the other fixed.

The profile includes human work before, during and after the campaign. Apparatus construction, protocol development, real-time intervention, curation and manual confirmation affect deployment cost or claim assessment. Reporting only unattended operating time can conceal the work required to obtain and assess the result.

\section{Epistemic layer: experiment selection and scientific reasoning}\label{sec:mf-methods}

\subsection{Terminal decision criteria governing method selection}

The epistemic layer compares Select and Acquire methods by their terminal decisions: recommend a design, predict a response over a domain, discriminate hypotheses, or choose a state-dependent action sequence. These decisions have different losses; comparisons therefore fix the scientific endpoint and resource boundary. Each method must also support the evidence cycle: traceable records, fault diagnosis, calibration and claim-specific follow-up (Section~\ref{sec:framework}).

Embodiment constrains the experiments a selector can deliver. Programmable synthesis supplies an execution interface \citep{steiner2019,mehr2020}; robot scientists and adaptive microscopy use its observations differently \citep{king2009,event2022}. Matched comparisons specify five attributes: scientific objective; action type (preparation, intervention, acquisition or diagnosis); feedback timescale (within-operation, between-trial or between-campaign); state information (commands, telemetry, specimen history or latent estimates); and evidential endpoint. These distinguish execution correction, experiment allocation and transfer, and whether selection conditions on commanded or realized conditions.

\subsection{Four terminal losses}

Let $\zeta$ denote a nominal campaign design, $X(\zeta)$ its realized physical condition, $g$ the declared endpoint, $\zeta^\star$ a feasible nominal design maximizing the platform-conditioned response, and $\hat\zeta_T$ the design recommended after independent verification. Let
\[
\begin{aligned}
G_{\mathrm{resp}}(\zeta)&=\mathbb E\big[g(X)\mid\operatorname{do}(\zeta),\hat v_t=1,\,H_t,\\
&\hspace{22mm}\text{qualified measurement}\big]
\end{aligned}
\]
be the response functional for the declared platform and validity rule, with $\hat v_t$ the audit acceptance of a generic trial $t$ run at design $\zeta$, as defined in Section~\ref{sec:formulation}. Let $\xi$ be a target input, $f$ the target response, $\ell$ a prediction loss, $\hat f_T$ the predictor, $P_{\mathrm{target}}$ the separately declared target distribution, $\mathcal L(\theta,\delta_T)$ the posterior loss for terminal decision $\delta_T$, and $\ell_{\mathrm{ctrl}}$ a fast-loop control cost. A policy $\pi^{\mathrm{ctrl}}$ satisfies the campaign resource constraint over horizon $T$:
\begin{equation}\label{eq:mf-terminal}
\begin{aligned}
L_{\mathrm{BO}}&=\mathbb E_{\mathrm{verification},\,\epsilon}
 \left[G_{\mathrm{resp}}(\zeta^\star)-G_{\mathrm{resp}}(\hat\zeta_T)\right],\\
L_{\mathrm{AL}}&=\mathbb E_{\xi\sim P_{\mathrm{target}}}
 \left[\ell(\hat f_T(\xi),f(\xi))\right],\\
L_{\mathrm{BED}}&=\mathbb E_{\theta\sim p(\theta\mid H_T)}
 \left[\mathcal L(\theta,\delta_T)\right],\\
L_{\mathrm{RL/MPC}}&=\mathbb E_{\pi^{\mathrm{ctrl}}}
 \left[\sum_{t=0}^{T}\sum_{k=0}^{K_t^{\mathrm{ctrl}}-1}
 \ell_{\mathrm{ctrl}}(x_{t,k}^{\mathrm{ctrl}},u_{t,k}^{\mathrm{ctrl}},w_{t,k}^{\mathrm{ctrl}};d_t)\right].
\end{aligned}
\end{equation}
The BO expectation covers independent verification and outcome noise under the declared qualified-measurement rule. The AL, BED and control expectations are over the variables shown. At campaign level, $z_t=(s_t,\kappa_t)$ is the scientific--apparatus state, $a_t$ is the high-level decision and $d_t$ is the disturbance or regime variable from Eq.~\ref{eq:generative}. At the fast control scale, $x_{t,k}^{\mathrm{ctrl}}$ is the local state, $u_{t,k}^{\mathrm{ctrl}}$ the low-level command and $w_{t,k}^{\mathrm{ctrl}}$ the fast process disturbance, and $K_t^{\mathrm{ctrl}}$ is the number of control steps within trial $t$. A campaign action compiles to a control sequence,
\[
\{u_{t,k}^{\mathrm{ctrl}}\}_{k=0}^{K_t^{\mathrm{ctrl}}-1}
 =\Gamma_{\mathrm{ctrl}}(a_t,s_t,H_t),
\]
\[
x_{t,k+1}^{\mathrm{ctrl}}
 =f_{\mathrm{ctrl}}(x_{t,k}^{\mathrm{ctrl}},u_{t,k}^{\mathrm{ctrl}},w_{t,k}^{\mathrm{ctrl}}).
\]
Thus the low-level state and command do not reuse the campaign symbols, and $d_t$ remains the campaign disturbance rather than a nominal design. Simple regret, predictive loss, posterior decision loss and trajectory cost are distinct estimands. Each of them is also conditional on which records were adjudicated valid, because the response functional $G_{\mathrm{resp}}$ in Eq.~\ref{eq:mf-terminal} is defined under a validity rule; so a study must report attempted, valid and confirmed units separately before any of these losses can be compared across methods (Section~\ref{sec:evaluation}). Table~\ref{tab:mf-decisions} connects each terminal decision to the assumptions required by a method family and to evidence from cited studies.

\paragraph{Bayesian optimization.}
BO recommends a feasible design maximizing a measured response or Pareto trade-off. It requires qualified preparation and measurement, a bounded design space, and noise and feasibility models. Independent verification distinguishes an optimum from a noise-favored winner. Multi-fidelity BO has no automatic advantage over single-fidelity search: its value depends on approximate-fidelity cost and informativeness \citep{f2025mfbobest}. Anubis handles unknown feasibility and failed experiments explicitly \citep{f2025anubis}. Proxy misalignment, validity errors and unmodeled feasibility change the response functional represented by the fitted surface.

Noisy expected hypervolume improvement targets batch multiobjective trade-offs \citep{qnehvi}. Olympus compares selectors on shared noisy-response problems, including mixed-parameter and multiobjective extensions, while Atlas connects planning to laboratory execution \citep{f2021olympus,f2023olympusenhanced,f2025atlas}. These comparisons screen selectors; prospective campaigns must additionally test physical feasibility and whether the assay measures the application property.

\paragraph{Active learning.}
Active learning minimizes predictive or representation loss over a declared target distribution specified independently of acquisition. Materials mapping, biological optimization and interpretable latent models illustrate coverage, response-guided sampling and low-dimensional discovery representations \citep{kusne2020,metis2022,f2025autosclab}. Each chooses where to measure next, but their terminal claims differ; acquisition score alone is not scientific utility.

High variance can reflect irreducible noise or regions outside the target distribution, not reducible uncertainty. BO can find a good design without learning a global response map; active learning can improve prediction without improving the best response. Accordingly, verify a BO recommendation independently and assess an active-learning model on observations from the declared target distribution.

\paragraph{Bayesian experimental design.}
Hypothesis-directed selection asks which explicit alternatives should govern the terminal decision. With $h$ the hypothesis variable, $e$ a feasible intervention or measurement design, $Y$ the resulting observation and $H_t$ the recorded history, the information-gain criterion is
\begin{equation}\label{eq:mf-information}
\mathrm{IG}(e)=I(h;Y\mid e,H_t),
\end{equation}
where $I(\cdot;\cdot\mid\cdot)$ is conditional mutual information \citep{chaloner1995}. Information gain is an intermediate objective. A measurement can distinguish two models yet leave the decision unchanged if both imply the same intervention; a less informative measurement can be decisive because it changes which claim or experiment is selected. The terminal loss $\mathcal L(\theta,\delta_T)$ in Eq.~\ref{eq:mf-terminal} makes this distinction explicit. BED is appropriate when alternatives predict distinct observable distributions under feasible interventions and measurements; robot-scientist and biological model-development campaigns provide experimentally connected examples \citep{king2004,coutant2019}. Observational equivalence under the feasible design class leaves the mechanism unresolved (Section~\ref{sec:partial-obs}); posterior concentration among misspecified alternatives therefore requires an external validation target. Report the alternatives considered and the decision changed by the selected observation.

\paragraph{Reinforcement learning and model-predictive control.}
RL and MPC choose action sequences when intermediate specimen or apparatus states change later options. They require a sufficient state or belief, endpoint-aligned rewards or dynamics, and adequate interaction and reset budgets. AlphaFlow demonstrates sequential nanoparticle-processing decisions with spectroscopic feedback \citep{alphaflow2023}. Proxy rewards can favor behavior unrelated to the final claim, and model-based safety guarantees depend on the model \citep{safeopt2015,cbf2017}. Under partial observability, model shift or reward mismatch, control return can diverge from scientific loss.

\begin{figure*}[t]
\centering
\includegraphics[width=\textwidth,keepaspectratio]{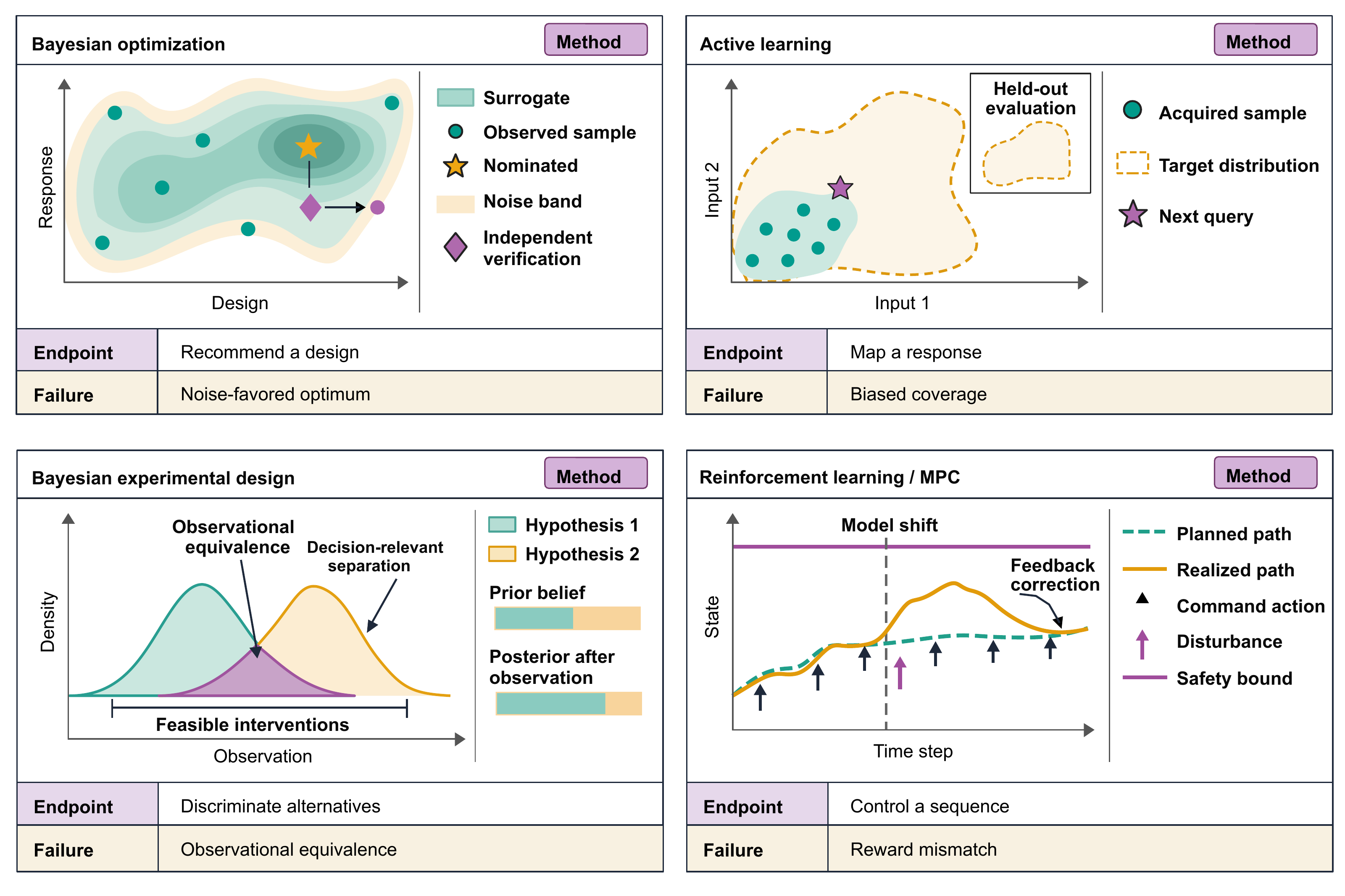}
\caption{Four terminal decisions determine the epistemic method comparison. Bayesian optimization recommends a design that requires independent verification; active learning supplies a response map assessed on a declared held-out distribution; Bayesian experimental design distinguishes decision-relevant alternatives; reinforcement learning and model-predictive control choose state-dependent action sequences. Their characteristic failures are different, so method choice and terminal loss must be tied to the scientific endpoint. Curves, maps and trajectories are conceptual illustrations with no empirical performance values.}
\label{fig:method-map}
\end{figure*}

Figure~\ref{fig:method-map} contrasts the four terminal decisions and their characteristic failures; the following constraints apply across these choices.

\subsection{Selection under feasibility, fidelity and cost}

\paragraph{Known constraints, safe exploration, unknown feasibility.}
SafeOpt assumes a known safe seed and a constraint model for exploration above a safety threshold \citep{safeopt2015}; control-barrier-function guarantees likewise depend on the model \citep{cbf2017}. Laboratory feasibility is often learned only after resource-consuming preparation attempts. Anubis models unknown feasibility; Dual-GP oversight constrains search through real-time raw-data assessment and permits human intervention for unanticipated results \citep{f2025anubis,f2025oversight}. Comparisons must distinguish known constraints, information learned from failures and safety under model error.

\paragraph{Batch selection and learned constraints.}
Plates, parallel reactors and multi-arm cells make a batch the selection unit. Local penalization discourages nearby points using an estimated objective Lipschitz constant \citep{gonzalez2016batch}. Gelbart et al. weight acquisition by surrogate feasibility probability; constrained predictive entropy search extends information-based selection \citep{gelbart2014unknown,hernandez2015pesc}. Applying these methods requires reliable feasibility labels. A degraded or missing measurement after failed preparation is not necessarily an infeasible design (Section~\ref{sec:formulation}); conflating them teaches the selector the wrong boundary.

\paragraph{Multi-fidelity value is a bias--cost--correlation trade-off.}
An approximate fidelity helps when it is cheaper than the terminal assay and remains informative about it once its bias is modeled. If it is cheap but weakly related to the endpoint, it consumes budget without moving the posterior about $G_{\mathrm{resp}}$; if its bias varies with design beyond the correlation model, it can mislead selection. The systematic MFBO evaluation supports this conditional reading: approximate-fidelity informativeness and cost determine whether MFBO is worthwhile \citep{f2025mfbobest}. Early prediction inside a closed-loop campaign is the laboratory version of the same choice. Attia et al. predicted battery cycle life from the first few cycles and used BO to screen 224 six-step fast-charging protocols in 16 days, with subsequent validation \citep{attia2020}. The relevant comparison quantifies selection bias introduced when the inexpensive signal, rather than the terminal assay, drives most choices.

\paragraph{Asynchronous Execution, Batch Concurrency, and Delayed Assays.}
Batch selection acquires multiple designs together; asynchronous selection must act while assays remain pending. Noisy expected hypervolume improvement handles parallel noisy multiobjective selection \citep{qnehvi}, and Olympus and Atlas supply comparison and execution interfaces \citep{f2021olympus,f2025atlas}. Setup, cleaning, plate capacity and assay delay determine whether ten parallel designs cost less than ten sequential trials. Report costs per attempt, valid observation and elapsed campaign time. Pending outcomes enter acquisition through predictive distributions, not realized values. Delays change information, dependence and cost without changing a fixed terminal estimand; outcome-dependent completion, clock cutoffs or selective inclusion of pending results can instead change the observed target population and require explicit modeling.

\paragraph{Resets, execution parameters and belief-sequential control.}
Scientific campaigns do not restart from identical states. A failed trial can consume a rare sample, dirty an apparatus or leave a specimen changed, so the reset assumption is a modeling choice. When preparation is expensive, irreversible or state-dependent, selection should value keeping options alive and act on the belief state of Eq.~\ref{eq:joint_belief}, not on a fresh-start approximation. BO remains useful when intermediate observations leave later feasible actions unchanged, because a recipe can then be represented as one structured design. Execution parameters can themselves be optimized: above 100~cP, air-displacement pipettes fail to perform accurate and precise transfers, and multi-objective Bayesian optimization of aspiration and dispense rates reached transfers within 5\% percentage transfer error for viscosities up to 1275 cP on the tested pipettes \citep{quijano2024pipetting}, a result that depends on transfer volume. Such tuning changes which nominal designs are realizable.

\paragraph{Diagnosis is a selection decision.}
Blanks, delivery measurements, references and orthogonal assays diagnose contamination, execution, drift and identity, respectively. Equation~\ref{eq:voi} values diagnosis by its cost-adjusted continuation relative to the best alternative first action, including stopping. It credits both improved decisions and direct claim resolution. Charge attempts, diagnosis and resets to the same resource boundary, and state whether missing outcomes represent infeasibility, censoring or unresolved evidence. Otherwise a method can appear efficient by excluding quality control or invalid trials from its costs.

\begin{table*}[t]
\caption{Decision frameworks: terminal decision, required assumptions, observed evidence and use rule. Each row links a method family to the assumptions required for its scientific decision and to evidence from cited studies.}
\label{tab:mf-decisions}
\centering\fontsize{8.2}{9.8}\selectfont
\setlength{\tabcolsep}{4.5pt}\renewcommand{\arraystretch}{1.12}
\begin{tabularx}{\textwidth}{@{}YYYYY@{}}
\toprule
\rowcolor{Light}\textbf{Framework} & \textbf{Terminal decision} & \textbf{Assumption required} & \textbf{Evidence and failure observed/reported} & \textbf{Decision rule: use when / fails when}\\
\midrule
BO & Best feasible response or Pareto trade-off; regret on a verified recommendation & Response is defined under qualified preparation/measurement; feasible region and observation noise are modeled & Multi-fidelity evaluation finds value dependent on fidelity cost and informativeness \citep{f2025mfbobest}; noisy maxima and unknown feasibility require independent verification and explicit failure handling \citep{f2025anubis} & Use when endpoint and feasible designs are measurable; proxy, validity or feasibility errors change the response functional\\[2pt]
\addlinespace[1pt]
Active learning & Predictive or decision-relevant loss over a declared target distribution & Target population is specified independently of acquisition; labels represent that target & Materials mapping, biological optimization and interpretable latent models illustrate coverage and representation learning \citep{kusne2020,metis2022,f2025autosclab}; high variance can select irreducible noise or irrelevant regions & Use for prediction or coverage over a declared domain; target or label selection makes the fitted model conditional on the acquired subset\\[2pt]
\addlinespace[1pt]
Bayesian experimental design & Choose among explicit hypotheses by expected information or terminal posterior decision loss & Candidate mechanisms are represented and make distinct predictions under feasible interventions/measurements & Robot-scientist and biological model-development examples connect hypotheses to experiments \citep{king2004,coutant2019}; identical observable distributions remain unresolved \citep{chaloner1995} & Use for discrimination among identifiable alternatives; equivalence or omitted mechanisms leave posterior concentration insufficient for claim support\\[2pt]
\addlinespace[1pt]
RL / MPC & Minimize control cost across a state-dependent trajectory while preserving the scientific endpoint & State/belief is sufficient for control; reward and dynamics track the endpoint; interaction/reset budget is available & AlphaFlow demonstrates sequential multistep decisions \citep{alphaflow2023}; safety guarantees remain conditional on the model \citep{safeopt2015,cbf2017} & Use when intermediate state and action order change later options; reward mismatch, partial observability or model shift can separate return from scientific loss\\[2pt]
\addlinespace[1pt]
LLM / agent planners & Propose hypotheses, plans, code or tool calls; a numerical and physical procedure assesses the scientific endpoint & Tool/API access, valid protocol constraints and external statistical/execution checks are supplied & Reported benchmarks score task, code or reproduction outcomes; physical claim assessment requires an independently specified preparation or measurement \citep{f2024scienceagentbench,f2024corebench} & Use for literature-grounded proposals and workflow coordination; task success does not substitute for uncertainty calibration or claim confirmation\\
\bottomrule
\end{tabularx}
\par\vspace{4pt}\begin{minipage}{\textwidth}\fontsize{8}{9.6}\selectfont Each row connects a terminal decision to the assumptions required by a method family and to evidence from the cited studies. Frameworks can be combined, for example BO between trials and RL within a trial.\end{minipage}
\end{table*}

\subsection{Language-model planners and numerical posteriors}\label{sec:llm-planners}

LLM and agent systems occupy two positions relative to numerical selection. As workflow structurers, they turn a stated research goal into hypotheses, protocols, code and tool calls before the design space and objective are formalized. SayCan grounds language plans in learned affordances and Code as Policies emits executable control code \citep{ahn2023saycan,liang2023codeaspolicies}. Scientific instantiations distinguish hypothesis generation and design assistance \citep{gottweis2026coscientist,f2025crisprgpt} from multistep physical execution and tool-mediated laboratory work \citep{f2025aichemsit,organa2025}. These systems expand objective specification and execution coordination; the posterior remains tied to the recorded observations and likelihood.

Language models supply literature-grounded priors and proposals. BioDiscoveryAgent combines perturbation reasoning with numerical selection and reports improved prediction on evaluated datasets \citep{f2024biodiscoveryagent}. That is proposal performance, not uncertainty calibration, which requires held-out observations and a recorded prior and update rule. When proposals and evaluations reuse model-derived evidence, represent their dependence. Record the provenance of variables, constraints and candidates, including proposals originating outside the fitted model.

Two further routes place learned models inside the numerical selector. LLAMBO frames BO in natural language so that a language model proposes and evaluates candidates conditioned on past evaluations, with results reported mainly on hyperparameter-tuning problems \citep{liu2024llambo}; PFNs4BO trains a network by in-context learning on a sampleable prior, allowing user hints about optimum location and learned acquisition functions \citep{muller2023pfns4bo}. In both cases the prior is encoded in weights or a prompt, not a stated covariance, so a campaign should record the prior and test it against a conventional Gaussian-process surrogate on its own data. Amortized design moves part of online design computation into offline training; language-model proposals provide an additional source of candidate structure. Deep Adaptive Design trains a design policy offline so each experiment is chosen in a single forward pass, and RL-DAD extends this to implicit models whose likelihood cannot be computed but which can be simulated \citep{foster2021dad,lim2022policybed}. These methods suit settings with a trustworthy simulator and a cheap online decision, and they inherit simulator misspecification: if execution failure is absent from the simulator, the learned policy has no basis for diagnosing it (Eq.~\ref{eq:voi}).

Agent benchmarks locate the current boundary. ScienceAgentBench's expert-checked 102 tasks yield 32.4\% independent success for the best agent, 34.3\% with expert knowledge, and 42.2\% with a stronger model at more than ten times the cost \citep{f2024scienceagentbench}. CORE-Bench reports 21\% accuracy on its hardest computational-reproducibility level \citep{f2024corebench}. These endpoints concern successful data analysis or computational reproduction, not a physical-claim probability. A task passes a scripted check; a scientific claim requires independent preparation or measurement. Solved tasks, valid experiments and confirmed claims therefore remain separate record units.

\subsection{Language-guided proposals and constrained batch execution}\label{sec:llm-bo}

Language-guided proposals and constrained optimization meet at the execution interface. Literature-based candidates (Section~\ref{sec:llm-planners}) must be evaluated against feasibility boundaries learned through batch and constrained BO \citep{gonzalez2016batch,qnehvi,gelbart2014unknown,hernandez2015pesc,f2025anubis,f2025oversight}. An execution-aware model predicts whether a candidate will yield a valid observation, marginalizing execution uncertainty and updating with telemetry. Controlled comparisons distinguish design infeasibility from execution failure, charge every attempt to the same budget, and retain the provenance of proposal priors and learned constraints.

\subsection{Interfaces beyond the optimizer}

Numerical and linguistic layers issue commands that an apparatus must realize. Section~\ref{sec:architectures} evaluates perception, policies, world models and recovery against realized conditions, independent measurements and held-out interventions, alongside plan fluency or visual plausibility. Structured dynamical modeling and label-free prediction illustrate the same boundary from the measurement side: a reconstructed signal is a model output, while an independent observation enters the record \citep{sindy2016,noise2void,ounkomol}.

\section{The operational layer: embodied architectures for scientific experiments}\label{sec:architectures}

\begin{figure*}[t]
\centering
\includegraphics[width=.95\textwidth,keepaspectratio]{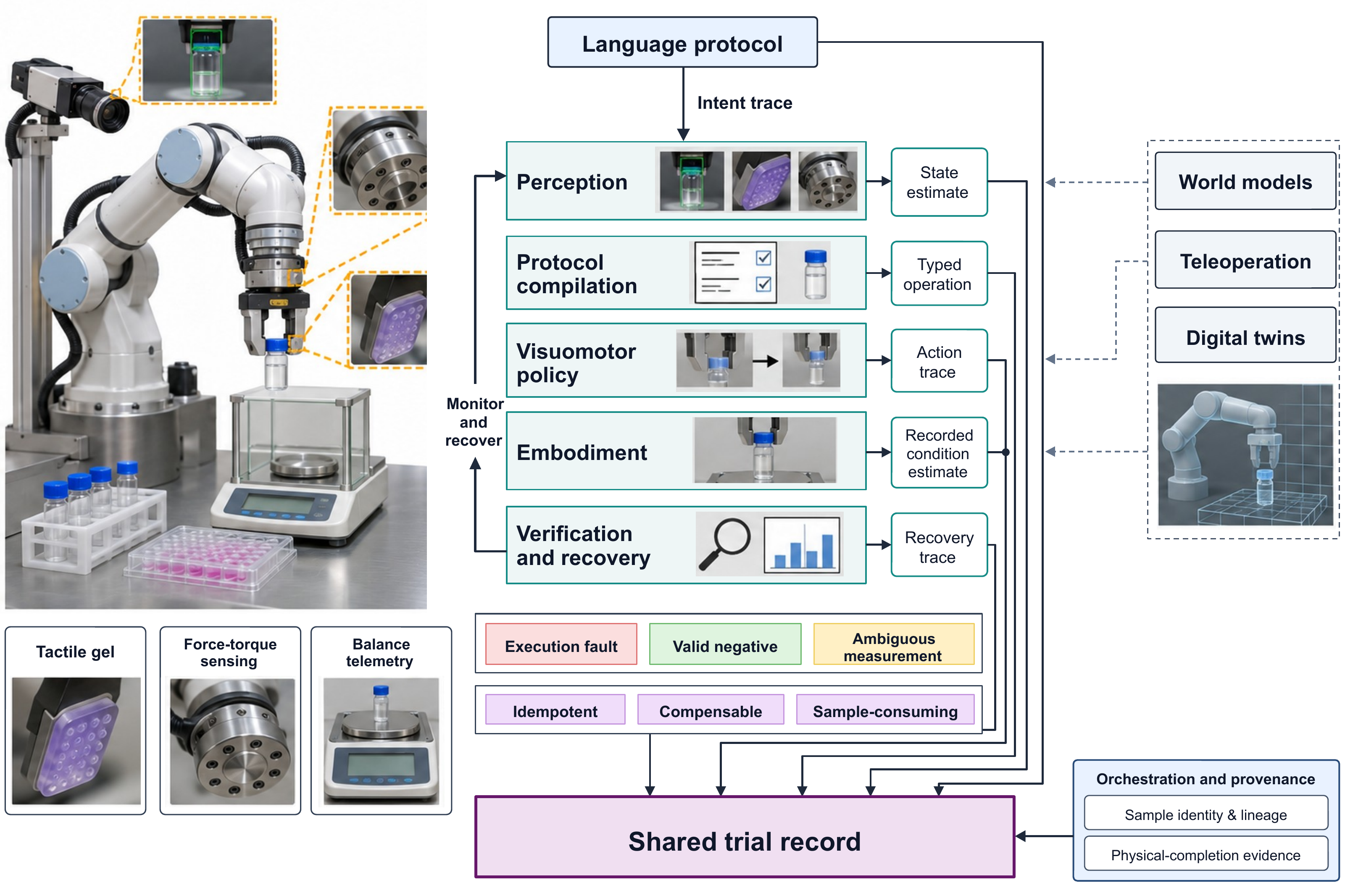}
\caption{Operational stack for a physical workcell. Language protocol is an input, not a stack layer: it emits the intent trace. Five operational layers---perception, protocol compilation, visuomotor policy, embodiment, and verification and recovery---emit, respectively, a state estimate, typed operation, action trace, recorded condition estimate and recovery trace into the shared trial record, from which evidence obligations are assessed. World models, teleoperation and digital twins support these layers and do not replace physical measurement. Sensing modalities include cameras, tactile gel, force--torque sensing and balance telemetry. The apparatus is a synthetic illustrative scene.}
\label{fig:stack}
\end{figure*}

The operational layer executes selected experiments and supplies telemetry constraining $x_t$ in $R_t$ (Section~\ref{sec:formulation}). Observations update $b_{t+1}$ and $\mathfrak C_{t+1}$, while resource, risk and irreversible-change costs enter Eq.~\ref{eq:utility}. Generalist policies, world models and failure monitors provide components that laboratory platforms bind to scientific operations \citep{zhao2023aloha,longley2026robinhood,dai2024}. We compare controlled variables and supporting measurements: correct motion can still deliver the wrong volume through rheology, calibration or pressure dynamics. Figure~\ref{fig:stack} and Table~\ref{tab:embodied-obligations} summarize the interfaces and evidence required.

\subsection{Perception and physical state estimation}\label{sec:perception}

Perception in a laboratory must estimate the quantity an operation depends on---meniscus position, fill state, precipitate presence, vessel pose, tip contact---rather than scene semantics alone. LIRA couples vision-language models to localization, inspection and reasoning for self-driving-lab workflows, adding closed-loop error detection where workflows otherwise run open-loop \citep{zhou2025lira}. Its laboratory setting makes the distinction concrete: localization supplies an enabling state estimate, while an experimental measurement additionally requires a qualified physical readout.

Contact sensing provides the direct observation of the realized condition for contact-dominated steps, extending perception beyond visual geometry. Vessel pose alone may leave immersion depth or contact force unresolved; the step requires a validated readout of its relevant physical variable, whether optical, tactile or force-based. Section~\ref{sec:tactile-fm} examines the criteria under which emerging tactile and force foundation models can serve as calibrated laboratory instruments, not task-specific policy embeddings; a model that has not been calibrated against a reference cannot yet be cited as a measurement.

Transparent or reflective labware and gripper or lid occlusion obscure transfer completion. Pose estimates identify location, not contents or preparation history. Tip immersion, cap torque and septum-piercing force require contact measurements, and glass imposes tighter safety tolerances than many manipulation benchmarks.

The consequence is an evaluation rule. Perception should be scored against the protocol-dependent quantity, with an independent measurement as reference: a balance for delivered mass, a force--torque sensor for contact, an orthogonal assay for identity. Error under the relevant conditions---occlusion, reflection, transparent labware, wet surfaces---and calibration drift over an extended campaign belong in the record, because cumulative thermal and mechanical drift over multi-day operation can misattribute an execution deviation to the specimen. Localization and inspection are enabling capabilities; they become scientific evidence when their error is stated on the variable the protocol tolerates.

\subsection{Tactile and force foundation models as laboratory instruments}\label{sec:tactile-fm}

The question a foundation model for contact has to answer in a laboratory is which physical quantity its readout supports and which decisions it informs. Tactile and force--torque sensing serve two complementary roles. As execution-governance signals they reveal slip, collision, seal integrity, vial seating and other contact events that determine whether a protocol step remains within tolerance. As primary scientific channels they can measure contact mechanics, rheology, modulus maps and wetting or adhesion responses. A representation used only for policy control need not itself be a force value, but a claim about a measured force or displacement requires traceability to a calibrated physical reference. Chemical or structural identity generally requires orthogonal spectroscopic or diffraction assays because contact signals do not uniquely identify composition or phase. Three capabilities therefore matter: representations that transfer across sensor units and wear, a readout calibrated to the quantity required by the named operation, and a drift record that makes the calibration interval a property of the record rather than an assumption.

The available evidence establishes transferable representations and parts of the control-oriented readout. Self-supervised touch representations pre-trained on large tactile image collections transfer across vision-based tactile sensors and downstream tasks, and unified static--dynamic representations span four visuo-tactile sensors \citep{higuera2025sparsh,feng2025anytouch}. ForceVLA treats six-axis force as a first-class modality through a force-aware mixture of experts \citep{yu2025forcevla}. These results are reported on contact-rich manipulation benchmarks, where the endpoint is task success. They support tactile sensing for execution governance; a scientific contact-mechanics claim still requires calibrated measurements.

A contact model should therefore be evaluated in two tracks. For execution governance, the test reports slip, collision, seal-integrity or seating detection against event labels and measures the resulting intervention rate, sample loss and protocol-valid rate. For contact as a scientific measurement, the test reports force, displacement or material-response error against an independent reference across sensor units, wear and temperature. Representation capacity and calibrated physical readout are separate evaluation targets. The drift record should state when the physical readout exceeds the protocol tolerance; without that record, a learned representation may support control while the scientific measurement remains unqualified.

\subsection{Language planning and protocol compilation}\label{sec:protocol-compilation}

Language is the practical interface for intent, but a scientific protocol must reach hardware as a typed, checkable artifact before motion can mean the intended experiment. Several complementary mechanisms have emerged. Affordance-grounded planning scores proposed steps against estimated executability; runtime checks must still assess whether the selected step meets physical preconditions \citep{ahn2023saycan}. Code generation makes the plan's assumptions explicit and executable: generated policy code composes perception APIs and parameterized control primitives, including feedback loops and logic \citep{liang2023codeaspolicies}. The Embodied Agent Interface formalizes where such systems fail---goal interpretation, subgoal decomposition, action sequencing and transition modeling---and reports fine-grained errors, not final success only \citep{li2024eai}.

Laboratory systems bind these interfaces to domain tools. Coscientist and ORGANA connect language planning to hardware and tool calls \citep{boiko2023,organa2025}; ChemAgents reports executed multistep synthesis-and-characterization tasks, including one in a different laboratory \citep{f2025aichemsit}. Earlier work already treats protocol as a first-class object through digitized synthesis and chemical programming languages \citep{mehr2020,rauschen2024}. The application synthesis distinguishes physical execution from software-led design assistance (Section~\ref{sec:applications}).

Language-mediated laboratory control needs an explicit reliability boundary. Empirical evaluations of autonomous chemistry agents, ORGANA and LLM agents for atomic-force microscopy report task- and benchmark-specific failures in planning, tool use and instrument operation \citep{boiko2023,organa2025,aila2025}. An observed error rate describes one task, tool set and escalation policy, so it transfers only to that configuration, and a low rate leaves structural failure modes unchecked. A model can hallucinate a parameter absent from the protocol, express uncalibrated confidence in an uncertain interpretation, or emit a syntax-valid tool call whose physical preconditions or safety envelope is violated. Parameter provenance, confidence calibration, typed preconditions and runtime interlocks are therefore separate checks, and successful parsing alone is not evidence of a safe or scientifically valid operation.

The mechanism of the gap is underspecification. A natural-language instruction such as ``add the reagent and heat gently'' leaves open identity, quantity, units, order, preconditions, acceptance thresholds and safety envelopes; neither affordance filtering nor fluent code generation resolves them, because the resolution is a domain decision, not a linguistic one. A protocol compiler---the function $\Gamma$ of Section~\ref{sec:formulation} in its operational role---must therefore emit the parsed protocol together with resolved parameters, rejected alternatives, preconditions, safety authorization and compilation errors, and link that version to the realized effects it produced. Compilation errors and human disambiguations are evidence about the feasible experiment space, not noise to be hidden; a reader who can inspect the compiled protocol and its execution trace can distinguish a valid assay from a fluent plan. The compiled configuration $m_t^{\mathrm{cfg}}$ and the metadata packet $m_t$ recorded afterwards are separate objects, and the compiler's output belongs to the first while timing, lineage, calibration version and telemetry belong to the second; a comparison across runs is only meaningful when both are present and the realized condition $x_t$ is measured independently of the command that produced it.

\subsection{Visuomotor policies: tokenized, diffusion and flow action generation}\label{sec:visuomotor}

Many generalist policies share a skeleton of vision and language encoders feeding a transformer trunk, while the action decoder determines important precision and latency trade-offs. ACT introduced action chunking for fine bimanual tasks such as threading cable ties and slotting a battery: predicting a sequence of future actions reduces compounding error from per-step prediction \citep{zhao2023aloha}. Diffusion Policy models the action distribution with a denoising process, capturing multimodal continuous trajectories \citep{diffusion2023}; $\pi_0$ formulates the action expert as a flow-matching model for general robot control \citep{pi0}. Autoregressive VLAs such as OpenVLA emit discrete action tokens and reuse language-model machinery \citep{openvla2025}. FAST identifies per-dimension, per-timestep action binning as a bottleneck and compresses action sequences in frequency space \citep{pertsch2025fast}. OpenVLA-OFT combines parallel decoding, action chunking and continuous actions, reporting LIBERO average success from 76.5\% to 97.1\% with 26-fold higher action throughput \citep{kim2025openvlaoft}.

These choices trade precision, multimodality and latency. Discrete tokenization lets a policy reuse the transformer's sequence machinery and benefits from web-scale pretraining, but binning continuous actions loses resolution exactly where laboratory tolerances live---microliters, millimetres of immersion, newton-metres of torque---and the quantization resolution can become a limiting factor even when the token distribution represents multiple modes. The relevant error is an error-propagation question: for a commanded variable $u$, to first order a realized-condition deviation satisfies $\delta x\approx(\partial \bar x_u/\partial u)\delta u+\delta x_{\mathrm{exec}}$, where $\bar x_u$ is the mean realized condition under $T_{\phi}$ and $\delta x_{\mathrm{exec}}$ is execution noise, so action quantization, execution noise and state-estimation error should be compared with the protocol tolerance, not judged from token accuracy alone. Continuous experts (diffusion or flow heads) represent those distributions natively, at the cost of iterative sampling and a heavier forward pass. Chunking amortizes that pass over a horizon, but feedback cannot change actions already issued before the next observation and replan. Receding-horizon execution and temporal ensembling can reduce this delay; the executed chunk length, not merely the predicted horizon, determines the response time available for a safety or contact event. For laboratory contact steps, force and slip evolve faster than vision-language inference: a slip that begins during a chunk is executed before the model can react. Contact-critical phases therefore need a fast inner controller, a shorter chunk or force feedback that bypasses the language trunk, and the chunk length used in each phase belongs in the record.

Viscous-liquid transfer is a documented case in which the realized condition departs from the command. Air-displacement pipettes fail to perform accurate and precise transfers above 100~cP \citep{quijano2024pipetting}. Multi-objective Bayesian optimization of aspiration and dispense rates, closed by automated gravimetric feedback, reached 5\% percentage transfer error for liquids up to 1275~cP on the tested pipettes; the result depends on transfer volume and is not a general tolerance (Section~\ref{sec:formulation}). RobInHood integrates liquid and solid handling, capping, heating, stirring, filtration and imaging in a fume hood \citep{longley2026robinhood}; mobile chemists connect workstations into campaigns \citep{burger2020,dai2024,lunt2024}. Evaluating generalist policies on these operations requires independent measurements of delivered quantity, contact or alignment error, contamination or specimen integrity against protocol tolerances, together with validity, latency, interventions and reset or sample-loss cost.

Two developments extend a policy's laboratory role. First, fine-tuning combines demonstrations, deployment interaction and human correction. ConRFT combines offline behavior cloning and Q-learning with online consistency-policy fine-tuning and human interventions \citep{chen2025conrft}; Online RL addresses deployment-state shift by collecting policy-induced states, at additional interaction cost. Laboratory use must charge irreversible sample consumption and exploration risk, and validate learned rewards against physical endpoints. Second, hierarchical designs separate deliberation from execution: Hi Robot resolves prompts and feedback above a low-level action policy \citep{shi2025hirobot}. A laboratory upper layer additionally needs typed steps that a verifier can reject.

Demonstration collection supplies a complementary route. A demonstration records motion, not delivered quantity. Laboratory demonstrations therefore need a synchronized independent measurement (mass, volume, torque) stored with each trajectory, or the resulting dataset cannot support the evidence rules of this section.

\subsection{World models and learned simulators}\label{sec:world-models}

Action conditioning lets a world model compare candidate interventions before they consume material, provided predictions are checked against measured outcomes. DreamerV3 learns dynamics and improves behavior through imagined rollouts across diverse control tasks \citep{hafner2025dreamerv3}; UniSim demonstrates deployment of policies trained inside a learned simulator to real robots \citep{yang2024unisim}. Joint policy--model learning extends this interface: WorldVLA reports mutual improvement when the world model predicts future images conditioned on actions and the action model predicts actions from observations \citep{cen2025worldvla}. These examples motivate action-aware prediction without establishing that visual rollouts recover a laboratory's scientific variables.

Action conditioning compares interventions within a model's supported regime; physical validity still requires intervention coverage and measured outcomes. Equation~\ref{eq:generative} separates execution $T_\phi$, joint specimen--apparatus transition $T_z$ and observation $M$, which video prediction entangles in pixels. Appearance alone need not identify delivered material, phase or assay response. One testable use is an execution model $T_\phi$ predicting realized conditions from commands, apparatus state and calibrated telemetry.

The evaluation rule follows. A world model earns scientific standing by rollout error on measured variables---delivered mass, temperature trajectory, measured response---under held-out interventions, with horizon-dependent uncertainty and paired agreement between simulated and physical rollouts at the same intervention. Paired sim-to-real evaluation of manipulation policies shows how to control the comparison, by identifying control and visual disparities between real and simulated environments without assuming a digital twin \citep{li2024simpler}. Structured dynamical identification remains the comparator when the goal is an interpretable mechanism rather than a predictive engine \citep{sindy2016}. Pixel-level fidelity is a development metric; the scientific metric is whether the model predicts what the apparatus and specimen will do under an intervention it has not seen. Benchmarks for world models are moving in this direction. WorldArena evaluates embodied world models on perceptual quality and on functional roles as data engine, policy evaluator and action planner, because perceptual fidelity alone overlooks downstream utility \citep{shang2026worldarena}. A policy evaluated inside a learned simulator is, however, evaluated against the model's physics. For a laboratory task, the proxy is acceptable only after its ranking of policies has been checked against measured endpoints on a held-out set of real runs.

\subsection{Failure detection, verification and recovery}\label{sec:failure-recovery}

A laboratory failure monitor must arbitrate among three interpretations that a task-level detector cannot distinguish: an execution fault, a valid negative result, and an ambiguous measurement. All three can produce the same task-level signal, a low or missing response, so a success flag alone cannot tell them apart; only an independent physical reference can. AHA trains a vision-language model to detect and explain manipulation failures using procedurally perturbed demonstrations \citep{duan2025aha}. Foresight monitors long-horizon manipulation using latents from an action-conditioned world model, trained only with final success labels, and calibrates detection thresholds with functional conformal prediction \citep{zhang2026foresight}. These results establish detection capabilities; laboratory recovery must additionally be assessed for specimen integrity and validity. Laboratory-specific inspection and safety work addresses wet-workflow faults and monitoring \citep{zhou2025lira,f2026chemisteye,f2026labrobfail}. A monitor should therefore be scored on whether it separates the three interpretations, with its ambiguity threshold checked against an independent measurement, not the monitor's own confidence (Section~\ref{sec:evaluation}).

The mechanism that separates laboratory recovery from software retry is state change. Repeating a dispense alters sample amount, concentration and contamination risk; remaking a sample creates a new physical object with a new history; re-imaging may be non-destructive only when exposure, photobleaching and specimen viability permit it. Idempotence is a property of the operation, not of the robot, and it must be declared per operation: idempotent (re-observe), compensable (undo or discard-and-remake) or sample-consuming (irreversible). Because a motion can succeed while the transfer fails, and a low signal can be a true negative, a classifier without a physical reference---a balance reading, an orthogonal assay, a pressure trace---cannot decide which entry of the categorical status $c_t$ is warranted. The record must therefore carry the detector's decision together with its physical reference and its calibration: false-accept and miss rates against declared reference criteria, human escalations, recovery actions, recovery time, sample loss, and a new lineage record when a sample is remade. Calibrated detection with declared retry semantics is what turns ``the robot noticed something'' into an interpretable entry of $c_t$.

\subsection{Teleoperation, cross-embodiment and sim-to-real transfer}\label{sec:transfer}

Transfer of motor competence and transfer of a scientific endpoint require different evidence: the endpoint, its interpretation and the measurement map must survive the change, which requires an equivalence check on a held-out unit (Section~\ref{sec:evaluation}). CrossFormer trains one policy on 900k trajectories from 20 embodiments without manual alignment \citep{doshi2025crossformer}; Open X-Embodiment and DROID provide heterogeneous demonstration data for policy learning \citep{oxe,droid}. These results establish motor-learning resources and capabilities, not endpoint equivalence. Teleoperation remains complementary: ALOHA's low-cost bimanual leader--follower platform supplies demonstrations and a human fallback for contact-rich operations \citep{zhao2023aloha}.

The mechanism of the endpoint gap is that the same motion produces different realized conditions on different apparatus. Geometry, calibration, tool compliance and fluid path alter what a command delivers; two platforms can agree on the motion while disagreeing on the intervention. Scientific transfer must therefore preserve intervention and observation semantics---what was added, what was measured, against what acceptance criterion---not motor commands. The evidence that closes this gap is endpoint agreement within a prespecified equivalence margin on a named held-out unit: a batch, an apparatus, a protocol version or a laboratory, together with the adaptation data, human labor and failed attempts the transfer consumed. Paired sim-to-real evaluation provides the template by comparing the same policy in matched simulated and physical settings while controlling visual and control disparities \citep{li2024simpler}; the laboratory analogue pairs the transfer with an independent endpoint measurement. Failure to detect a difference is not evidence of equivalence---the margin and the unit must be stated in advance.

\subsection{Orchestration and provenance}\label{sec:orchestration}

Orchestration is where per-step capabilities become a campaign, and its scientific function is to keep software state and apparatus state from silently diverging. ChemOS 2.0 provides an orchestration architecture for chemical self-driving laboratories; AlabOS contributes a reconfigurable workflow model with resource reservation; PyLabRobot offers a hardware-agnostic interface to liquid handlers \citep{chemos2,f2024alabos,pylab2023}. Chemical programming languages and reproducibility abstractions make the protocol itself executable and inspectable \citep{steiner2019,rauschen2024,canty2024}. Software coordination requires validation against physical-completion telemetry: communication latency, hardware timeouts, and mechanical deviations can cause physical execution to diverge from software acknowledgments.

The consequence is a provenance requirement that runs through every layer above. Each operation must join a stable sample identifier and parent--child lineage to an operation identifier, the protocol and device versions, calibration state, the physical-completion evidence (not merely the software acknowledgement), realized conditions, raw data and the analysis version. The acceptance test is a single-sample trace: a reader can follow one specimen through every operation and tell at each step whether a command was acknowledged, physically verified or left unresolved. Without that join, the layers can each perform well while the campaign still cannot establish which workflow produced the reported value.

\subsection{Component obligations}

Table~\ref{tab:embodied-obligations} condenses the section. Each row pairs a capability family with the experimental variable it must affect, the evidence a scientific workflow requires, and the observation that would close the remaining boundary. The final column specifies the measurement needed to assess each component in a scientific workflow.

\begin{table*}[t]
\caption{Operational-layer capabilities, experimental variables and evaluation requirements. Columns distinguish reported component results from the physical measurements needed to evaluate their use in scientific experiments.}
\label{tab:embodied-obligations}
\centering\fontsize{7.4}{8.5}\selectfont
\setlength{\tabcolsep}{3pt}\renewcommand{\arraystretch}{1.02}
\begin{tabularx}{\textwidth}{@{}>{\raggedright\arraybackslash}p{2.1cm}YYY@{}}
\toprule
\rowcolor{Light}\textbf{Component} & \textbf{Established in cited work} & \textbf{Scientific obligation and checkable evidence} & \textbf{Boundary and the observation that closes it}\\
\midrule
Perception and state estimation & VLM localization and inspection in labs; cross-sensor touch representations; force as a policy modality \citep{zhou2025lira,higuera2025sparsh,yu2025forcevla} & Estimate the protocol-dependent quantity (meniscus, delivered mass, contact, torque); report error against an independent reference under occlusion, reflection and transparent labware; track calibration drift & Pair each percept with an independent sensor reading and report error on the operation's dependent variable, not pose accuracy\\[2pt]
\addlinespace[1pt]
Language planning and protocol compilation & Affordance-grounded plans; executable policy code; laboratory tool-bound planning \citep{ahn2023saycan,liang2023codeaspolicies,boiko2023} & Emit a typed protocol with units, identity, preconditions, acceptance criteria, safety authorization and rejected alternatives; link the compiled version to realized effects & Treat compilation as a verified transformation: record unresolved parameters and rejected alternatives, and check acceptance criteria against measurements rather than fluent explanation\\[2pt]
\addlinespace[1pt]
Visuomotor policies & Action chunking for fine bimanual tasks; flow action experts; compressed action tokenization \citep{zhao2023aloha,pi0,pertsch2025fast} & Report delivered quantity, contact tolerance, valid-observation rate, latency, interventions and reset cost, compared with protocol tolerance & Score policies on realized $x_t$ and specimen integrity; state the action-representation and latency choices (tokens vs.\ continuous experts, chunk length) that produced the result\\[2pt]
\addlinespace[1pt]
World models and learned simulators & Model-based control; simulator-trained robot policies; joint policy--world-model training \citep{hafner2025dreamerv3,yang2024unisim,cen2025worldvla} & Predict measured scientific variables under held-out interventions with horizon-dependent uncertainty; report paired simulated--physical agreement at the same intervention & Evaluate rollouts on measured quantities rather than pixels, and separate execution, specimen and measurement dynamics; an execution model for $T_\phi$ is the near-term fit\\[2pt]
\addlinespace[1pt]
Failure detection and recovery & VLM failure detection and explanation; world-model-latent monitors; laboratory closed-loop inspection \citep{duan2025aha,zhang2026foresight,zhou2025lira} & Separate execution fault, ambiguous measurement and valid negative; report false-accept and miss rates against declared reference criteria, recovery actions, sample loss and new lineage records & Calibrate detectors against physical references and declare retry semantics per operation: idempotent, compensable or sample-consuming\\[2pt]
\addlinespace[1pt]
Transfer and teleoperation & Cross-embodiment policy training; teleoperation; paired sim-to-real evaluation \citep{doshi2025crossformer,zhao2023aloha,li2024simpler} & Preserve intervention and observation semantics across apparatus; report endpoint agreement within a prespecified equivalence margin on a named held-out unit, with adaptation data, labor and failed attempts & Test the scientific endpoint, not the motion, on the held-out apparatus or laboratory, with the margin fixed in advance\\[2pt]
\addlinespace[1pt]
Orchestration and provenance & Workflow orchestration; hardware-agnostic liquid handling; executable chemical protocols \citep{chemos2,pylab2023,rauschen2024} & Join sample identity, operation identifier, physical-completion evidence, protocol and device versions, calibration and raw data; support an end-to-end single-sample trace & Log physical completion and realized conditions at each step so that software state and apparatus state cannot silently diverge\\
\bottomrule
\end{tabularx}
\par\vspace{4pt}\begin{minipage}{\textwidth}\fontsize{8}{9.6}\selectfont Columns distinguish reported component results from the physical measurements needed to evaluate their use in scientific experiments.\end{minipage}
\end{table*}

\section{Evidence across scientific applications}\label{sec:applications}

For each domain, this section records which of the six decisions the cited workflows delegate, through which embodiment, and how far the record advances through the evidence cycle (Record, Diagnose, Calibrate, Confirm; Section~\ref{sec:evidence-levels}). Execution can outrun measurement validity: an apparatus can complete its operations while the identity or quantity of interest remains unresolved. Selection policies shape the data they later explain, so acquisition and optimization results need independent held-out evidence. A reported decision implementation is workflow automation, not independent claim confirmation. Chemistry and materials provide examples linking preparation, selection and characterization; biology, microscopy and field science highlight hypothesis testing, acquisition and access. Throughout, the labels ``design-only'', ``execution'', ``valid measurement'' and ``confirmation'' map respectively to the figure's Select/Interpret without physical execution, Prepare/Acquire/Access, Interpret with Acquire, and Confirm (reported, not claim-level).

\subsection{Chemistry and molecular experimentation}

Chemistry studies establish that an embodied decision loop is testable when a physical interface connects a selected reaction to a measurable analytical endpoint. Modular and flow platforms made procedures executable across defined spaces; process sensors monitor whether an operation reached its intended condition, while an optimizer selects the next recipe \citep{steiner2019,coley2019}. Since a nominal recipe does not reveal the realized trajectory, studies combining process and experiment-selection feedback must preserve both.

RoboChem connects liquid handling, pumping and a tunable photoreactor to inline NMR and Bayesian optimization of photocatalytic reactions \citep{robochem2024}. Researchers still define the search space, prepare stocks and isolate products; isolation checks the automated analytical endpoint but does not establish independent chemical novelty. Its autonomy depends on configuration: the instrument closes reaction optimization, while retained human operations bound the claim.

Mobile robots solve station access, not the same control problem. Burger et al. connect existing stations to researcher-defined photocatalyst search; Dai et al. add LC--MS and NMR and use heuristic follow-up to assess reaction hits and reproducibility \citep{burger2020,dai2024}. Orthogonal analysis determines which findings merit further work, but structural novelty still needs expert scrutiny. Granda et al. likewise retain chemist-led follow-up in robotic reactivity exploration \citep{granda2018}. These comparisons imply a concrete reporting requirement: name which sample handovers and analytical decisions are automated, and add discriminating characterization when the primary endpoint cannot separate products.

Language agents differ in whether they execute experiments. Coscientist, ChemCrow and ORGANA connect language-mediated planning to tools, experiments or visual feedback \citep{boiko2023,bran2024,organa2025}. ChemAgents physically executed six multistep synthesis-and-characterization tasks and a seventh photocatalytic task in a new laboratory, establishing execution within prescribed tasks rather than independent product confirmation or cross-laboratory transfer \citep{f2025aichemsit}. BioDiscoveryAgent and CRISPR-GPT instead support perturbation or gene-editing design and planning, with the agent described as an AI co-pilot in genome engineering \citep{f2024biodiscoveryagent,f2025crisprgpt}.

Co-Scientist generates hypotheses for downstream testing. Its published validation tested drug-repurposing candidates and synergistic combinations for acute myeloid leukemia using \textit{in vitro} cell-line assays, conducted with expert scientists who guided the work and prioritized the wet-laboratory experiments \citep{gottweis2026coscientist}. The authors describe these experiments as an initial biological validation step; they are evidence for the tested hypotheses, not autonomous experimental execution by the software or demonstrated clinical efficacy. By contrast, BioDiscoveryAgent's benchmark scores selected perturbations against existing datasets (see the biological applications below). Proposal quality, execution and physical validation are therefore separate responsibilities; tool use or a successful software run alone does not confirm a scientific result.

\subsection{Materials synthesis and energy systems}

Materials studies separate computational candidates, executed synthesis and experimentally confirmed identity. GNoME expands a predicted-crystal set; Cheetham and Seshadri find little evidence that the computational candidates jointly meet novelty, credibility and utility; their critique addresses that computational set, not the A-Lab campaign \citep{merchant2023,f2024aidriving}. In the amended account, A-Lab synthesized 36 compounds from 57 targets in 17 days with robotic synthesis and diffraction-guided refinement \citep{alab2023}. The 2026 correction finds the original conclusion correct for 36 of 40 reported successes, marks four compounds inconclusive after post-publication reanalysis, and narrows novelty to materials new to the prediction platform \citep{alabcorr2026}. Leeman et al. questioned automated Rietveld reliability and noted neglected compositional disorder \citep{leeman2024}. The record supports automated campaign execution and diffraction-guided synthesis; phase confirmations and novelty claims are calibrated by the published correction, which leaves four compounds inconclusive. Because automated synthesis policies require discriminating characterization to resolve ambiguous diffraction, future campaigns require independent orthogonal assays alongside separated accounting for attempted syntheses, reported candidates, and confirmed phases.

Materials platforms also show why processing and measurement belong in the endpoint. Lunt et al. integrated crystallization, sample preparation and diffraction, closing recipe-selection and measurement-readiness problems \citep{lunt2024}. MacLeod et al. coupled thin-film preparation and characterization to selection, making processing history part of the material state \citep{macleod2020}. Ada Pareto mapped palladium-film conductivity against processing temperature and transferred three drop-cast recipes to ultrasonic spray coating on glass at 191, 200 and 226\,$^\circ$C; it did not report deposition on the named polymer substrates \citep{adapareto2022}. This demonstrates process transfer between specific deposition modalities on glass, while substrate portability across flexible polymers represents a separate experimental objective requiring explicit interface characterization.

Energy studies distinguish fast proxies from use-level performance. Attia predicted cycle life from early-cycle data and used Bayesian optimization to screen 224 six-step fast-charging protocols in 16 days, followed by validation \citep{attia2020}; Dave's robotic conductivity search was followed by pouch-cell testing, so conductivity alone is not battery performance \citep{dave2022}. AlphaFlow addresses flow processing, staged synthesis and sequential actions, where intermediate state shapes the next operation \citep{alphaflow2023}. Rainbow used four robotic modules for 3,648 perovskite-nanocrystal syntheses, optimizing quantum yield and linewidth at six target emission energies and scaling synthesis 30-fold \citep{rainbow2025}. The campaign reports optimization and scale-up at the specified optical endpoints. The studies use different materials, apparatus and endpoints, so each next study should state whether it reports a design, executed material, measured property or confirmed claim.

\subsection{Biological experimentation and regenerative medicine}

Biological systems show that embodied loops support bounded hypothesis testing or design optimization, with conclusions set by the assay and supplied search space. Robot scientists tested functional-genomic hypotheses; Coutant et al. coupled model development to laboratory automation to revise a yeast model; BacterAI selected culture conditions and used robotic measurements to derive interpretable microbial-growth rules \citep{king2004,king2009,coutant2019,bacterai}. These systems connect decisions to physical perturbation and observation, making results inspectable against tested organisms, media and assays. Starting without a response dataset does not remove structure: available interventions still define the search. BacterAI differs from explicit model-based systems in deriving rules from selected measurements.

Other platforms optimize specified objectives, not general mechanisms. BioAutomata couples design--build--test--learn to a constrained lycopene pathway, while METIS optimizes cell-free expression, circuits and carbon fixation through manual or liquid-handler experiments \citep{bioautomata2019,metis2022}. They demonstrate selection interfaces, but assay improvement is not clinical efficacy and optimized designs do not establish causal mechanisms. Follow-up should test the intended application or explanation, not merely repeat the optimization score.

Recent agents clarify design versus execution. BioDiscoveryAgent designs perturbation experiments and reports an average 21\% improvement in predicting relevant genetic perturbations across six datasets---including the authors' unpublished CAR-T dataset, which limits training-data contamination---relative to task-specific Bayesian-optimization baselines (hit ratio at experimental round 5, averaged over 10 runs; source version arXiv:2405.17631v3, revised 9 March 2025); this reflects retrospective predictive design performance on curated benchmark datasets, not an embodied physical campaign \citep{f2024biodiscoveryagent}. CRISPR-GPT assists with experiment planning, guide-RNA design and analysis; Co-Scientist generates hypotheses for downstream verification \citep{f2025crisprgpt,gottweis2026coscientist}. SAMPLE instead physically assembled 1,352 designed GH1 sequences, expressed them cell-free and measured stability by T50. Follow-up in \textit{E. coli} lysate used different expression and assay conditions and a same-modality thermostability measurement \citep{sample2024}. This contrast separates design, physical execution and confirmation: a proposal is not a biological response, and a result in one expression context does not automatically transfer to another. In regenerative medicine, passage and prior handling shape cell state \citep{kanda2022}; independently prepared cultures, batches or laboratories—not repeated readings of one well—test transfer.

\subsection{Microscopy and adaptive scientific measurement}

Adaptive measurement changes the evidence before it changes its interpretation: a policy chooses which events, locations or spectral features enter the dataset. Accuracy measured only on acquired points therefore cannot validate the policy, because unvisited regions provide no errors for that reconstruction to reveal. Event-driven microscopy changes imaging rates when learned recognition detects biological dynamics, while Xu et al. train beam alignment in simulation and deploy it on a scanning transmission electron microscope; the first selects when to observe, the second positions the instrument \citep{event2022,align2022}. FAST combines a neural network, route optimization and hardware control to sample sparsely; in dark-field X-ray microscopy of a WSe$_2$ film, under 25\% of a full scan sufficed \citep{fast2023}. This establishes efficient imaging in that test, not that the route policy preserves rare or unanticipated structures.

The beamline studies show distinct steps of this problem. Adaptive XRD coupled phase identification to a diffractometer and used early measurements to steer later positions \citep{f2023xrdphase}. The XRD result supports phase identification in tested settings, but identification confidence over model-selected positions does not establish coverage of alternatives outside those positions. ARPES used Gaussian-process regression to search momentum and sample-surface positions for high intensity or sharp features \citep{f2024arpes}. These records establish instrument steering and bounded campaigns, not a shared endpoint or policy-independent validation.

The inferential gap is structural. A detector that triggers only on familiar phenotypes can exclude unfamiliar events; spatial selection can miss rare regions; a phase model or ARPES acquisition function directs measurements toward its own candidate features. Even perfect reconstruction of the acquired subset cannot recover evidence never sampled. Denoising and label-free prediction add a second boundary: model-derived pixels are not acquired measurements, so errors in missed cells or boundaries must be checked against independent annotations or complementary measurements \citep{noise2void,ounkomol}. AILA and AFMBench distinguish domain question answering from AFM operation, showing why tool knowledge cannot substitute for complete workflow performance \citep{aila2025}. The consequence is to report policy and coverage together and retain exploratory, calibration and held-out acquisitions; Table~\ref{tab:table8} states the corresponding tests. They assess what the selector overlooks, not merely whether its chosen points are reconstructed accurately.

\begin{table*}[t]
\caption{Measurement decisions and validation requirements in embodied AI for science}
\label{tab:table8}
\centering\fontsize{8.5}{10.2}\selectfont
\setlength{\tabcolsep}{5pt}\renewcommand{\arraystretch}{1.12}
\begin{tabularx}{\textwidth}{@{}YYY@{}}
\toprule
\rowcolor{Light}\textbf{Measurement decision} & \textbf{Scientific quantity at risk} & \textbf{Validation required}\\
\midrule
Select event-containing regions \citep{event2022} & Counts, morphology, composition & Propagate selection and boundary errors to the endpoint; test new batches and imaging conditions\\[3pt]
\addlinespace[1.5pt]
Trigger acquisition \citep{event2022} & Event frequency, duration, sampled population & Compare independent windows; assess missed or unfamiliar events\\[3pt]
\addlinespace[1.5pt]
Select spatial measurements \citep{fast2023,kusne2020} & Response maps, transitions, rare regions & Use held-out acquisitions; account for measurement and motion costs\\[3pt]
\addlinespace[1.5pt]
Choose a signal channel \citep{channel} & Relationships between physical signals & Check calibration and held-out measurements; separate association from mechanism\\[3pt]
\addlinespace[1.5pt]
Denoise or predict image content \citep{noise2void,ounkomol} & Weak structures and quantitative signals & Preserve acquired data; verify model-derived content independently\\[3pt]
\addlinespace[1.5pt]
Accept a scientific identity \citep{dai2024,alabcorr2026} & Product or phase assignment & Use complementary characterization when the primary measurement cannot distinguish alternatives\\[3pt]
\bottomrule
\end{tabularx}
\par\vspace{4pt}\begin{minipage}{\textwidth}\fontsize{8}{9.7}\selectfont The validation column proposes tests for the scientific quantity affected by each measurement decision.\end{minipage}
\end{table*}

\subsection{Field science and planetary exploration}

Field systems close an acquisition-and-control loop because communication delay and limited energy prevent continuous human selection of every action. Perseverance's ENav uses two-stage path selection, conservative collision checks, slip-robust bounding boxes and FPGA-accelerated vision on a single-core CPU; it increased average daily driving distance by an order of magnitude, and about 90\% of 32.1 km of driving used ENav \citep{f2026perseverancenav}. PIXL recognizes points of interest from ternary composition diagrams or machine-learning thresholds and lengthens integrations there; its authors describe this as the first autonomous decision-making based on real-time compositional analysis on another planet \citep{f2025pixl}. These results establish onboard navigation and adaptive measurement, not geological explanation: the mission supplies preferences or decision rules, and longer integrations at selected points do not test unselected locations.

Ocean vehicles show the same trade-off in a changing environment. Zhang et al. tracked and sampled a deep chlorophyll maximum; two autonomous vehicles later measured real-time chlorophyll-a distributions, adapted their movement to build three-dimensional maps of phytoplankton patches, and used those maps to direct targeted sampling \citep{zhang2020,f2024multiuv}. This closes the loop from measurement to target allocation, but selection-conditioned maps are not unbiased estimates of the whole field, and targeted sampling is not a controlled intervention. Because communication, energy and environmental change constrain repeat visits, a field campaign cannot assume that missed locations can be measured later or that an uncontrolled change identifies a cause. Reports therefore establish operational autonomy and targeted observation; causal or population-wide claims require independent coverage, repeated observations or an intervention that separates competing explanations. AEGIS likewise selects ChemCam targets under scientist-defined preferences, not mission objectives or geological explanations \citep{francis2017}.

\subsection{Cross-domain synthesis}\label{sec:cross-domain}

The record-linked evidence cycle runs Record, Diagnose, Calibrate and Confirm as analytic functions with feedback. Record captures the declared intervention, realized condition, measurement and lineage. Diagnose separates execution faults, valid negatives and ambiguous measurements. Calibrate establishes the relation between the recorded signal and the physical quantity or tolerance relevant to the claim. Confirm subjects a nominated finding to an independent preparation, orthogonal assay, held-out condition or separate laboratory. These functions answer different questions and form a claim-specific dependency; Table~\ref{tab:table3} connect each workflow's delegated decisions to its measured endpoint and follow-up evidence. Software-led proposal components such as BioDiscoveryAgent and Co-Scientist provide Record- and Select-level evidence for their AI component; human-executed follow-up is credited only when sourced and assay-established. FAST and PIXL provide physically executed adaptive measurements. Ada Pareto's transfer of three recipes from drop-casting to spray-coating on glass and SAMPLE's \textit{E. coli} lysate follow-up illustrate the Confirm stage \citep{adapareto2022,sample2024}.

The named evidence atlas in Appendix~\ref{sec:appendix} records reported delegation and validation pathways for 26 physical workflows and two software-boundary components; these codes are not composite autonomy or claim-confirmation scores.

Five failure modes explain why execution outruns evidence. First, insufficient selectivity leaves identity unresolved: four of 40 A-Lab reported successes remained inconclusive after reanalysis, and Leeman et al. questioned Rietveld reliability and neglected compositional disorder \citep{alab2023,alabcorr2026,leeman2024}. The failure is measurement ambiguity, not synthesis; complementary characterization is needed before claiming phase or novelty. Second, policy-dependent sampling makes observed data a product of the selector. Event-triggered microscopy, FAST, adaptive XRD and PIXL choose when, where or how long to measure; unvisited events and regions cannot be validated by accuracy on acquired points. Independent coverage and held-out scans test that gap (Table~\ref{tab:table8}). Third, supplied search spaces recur because optimization requires an objective and candidate set: Mobile Chemist uses researcher-defined options, RoboChem researchers define the search space and prepare stocks, and Co-Scientist addresses a stated objective. This establishes conclusions within the declared search space; expanding that space requires additional interventions and validation.

Fourth, human steps supply what the automated endpoint cannot: Granda's chemist follows up, RoboChem researchers prepare stocks and isolate products, and PIXL's mission team interprets measurements. These decisions are part of the workflow and must count in autonomy claims. Fifth, campaigns often stop at operation or proxy measurement. Dai et al. check hit reproducibility, Dave tests electrolytes in pouch cells, Ada Pareto transfers recipes, and SAMPLE retests in \textit{E. coli} lysate; each tests a specific boundary, not general validity. Without a claim-testing follow-up, completion establishes a measured outcome, not a scientific interpretation that survives another test.

A pooled comparison would require incompatible endpoints—yield, conductivity, phase identity, imaging efficiency, stability or field coverage—with different populations, budgets and human work. Scan, sample and confirmed-claim counts are not interchangeable, and elapsed time can omit preparation, recovery and measurement costs. The review therefore compares which decisions each system closes, which endpoint it measures and which evidence function remains unresolved within each workflow.

\begin{table*}[t]
\caption{Cross-domain comparison by physical feedback, scientific endpoint and validation boundary.}
\label{tab:table3}
\centering\fontsize{8.1}{9.8}\selectfont
\setlength{\tabcolsep}{4pt}\renewcommand{\arraystretch}{1.13}
\begin{tabularx}{\textwidth}{@{}YYYY@{}}
\toprule
\rowcolor{Light}\textbf{Workflow family} & \textbf{Physical feedback} & \textbf{Scientific endpoint} & \textbf{Validation boundary}\\
\midrule
Mobile and exploratory chemistry \citep{burger2020,dai2024,granda2018} & Mobile station access; reaction and formulation selection & Photocatalytic performance or reaction hits & Supplied search space; Dai and Granda add sourced follow-up of selected outcomes.\\[3pt]
\addlinespace[1pt]
Flow and sequential synthesis \citep{robochem2024,alphaflow2023} & Inline flow measurement; Bayesian or sequential action selection & Analytical reaction response or nanoparticle spectrum & Preserve realized trajectories; RoboChem retains stock preparation and isolation by researchers.\\[3pt]
\addlinespace[1pt]
Language-mediated laboratories \citep{organa2025,f2025aichemsit} & Robot/tool execution with visual feedback or multiagent planning & Completion of prescribed experimental tasks & Human disambiguation and task scope remain explicit; task completion does not establish independent confirmation.\\[3pt]
\addlinespace[1pt]
Inorganic and diffraction workflows \citep{alab2023,alabcorr2026,lunt2024} & Powder/furnace synthesis or integrated sample-to-PXRD handling & Phase assignment or measurement readiness & A-Lab uses the corrected record; operational integration and resolved material identity are distinct endpoints.\\[3pt]
\addlinespace[1pt]
Thin films and optical materials \citep{macleod2020,adapareto2022,rainbow2025} & Automated preparation, characterization and objective-driven selection & Film properties or nanocrystal optical response & Ada Pareto transfers three recipes between deposition modalities on glass; other substrates remain untested.\\[3pt]
\addlinespace[1pt]
Battery and electrolyte search \citep{attia2020,dave2022} & Physical cycling or robotic formulation; predictive/Bayesian selection & Cycle-life or conductivity endpoint & Validate the use-level target: conductivity search is followed by separate pouch-cell testing in Dave et al.\\[3pt]
\addlinespace[1pt]
Biological models and culture \citep{king2004,king2009,kanda2022} & Robotic perturbation, culture and assay feedback & Gene-function hypotheses or culture conditions & Available interventions, supplied models and preparation history bound interpretation.\\[3pt]
\addlinespace[1pt]
Biological design loops \citep{bioautomata2019,metis2022,sample2024} & Foundry or liquid-handler selection; sequence/function updating & Pathway, network or protein assay response & Integration depends on the application; SAMPLE adds follow-up in an alternative expression/assay context.\\[3pt]
\addlinespace[1pt]
Event and spatial acquisition \citep{event2022,fast2023,f2023xrdphase} & Adaptive rates, routes and diffraction positions & Sampled events, images or phase assignments & Test held-out acquisitions and missed regions; accuracy on the selected subset does not establish coverage.\\[3pt]
\addlinespace[1pt]
Instrument control \citep{align2022,aila2025} & Electron-beam alignment or AFM tool coordination & Alignment or instrument-workflow performance & Simulation deployment and domain question answering require separate physical endpoint tests.\\[3pt]
\addlinespace[1pt]
Planetary acquisition \citep{f2025pixl,francis2017} & Onboard target selection and adaptive integrations & Composition readings or selected targets & Mission preferences and interpretation remain supplied; targeted readings do not establish unvisited coverage.\\[3pt]
\addlinespace[1pt]
Ocean sampling \citep{zhang2020,f2024multiuv} & Underwater tracking, mapping and targeted sampling & Moving-layer or phytoplankton distributions & Adaptive observation is distinct from a controlled ecological intervention or an unbiased field census.\\[3pt]
\addlinespace[1pt]
\bottomrule
\end{tabularx}
\par\vspace{3pt}\begin{minipage}{\textwidth}\fontsize{7.7}{9.3}\selectfont Rows group related questions rather than pool performance. Named studies retain their individual endpoints and responsibilities in the text. PXRD, powder X-ray diffraction; AFM, atomic force microscopy.\end{minipage}
\end{table*}

\subsection{Access, allocation and interpretation}

The evidence map therefore reports delegated capability, not scientific validity: an R code records reported implementation, an O code records a reported orthogonal assay, and neither establishes claim-level confirmation. In particular, the two software-boundary components are boundary comparators, not independent physical executions, and the remaining rows are not a pooled performance ranking because their endpoints, populations and resource boundaries differ. The complete machine-readable coding is provided in the supplementary files.

\section{Data, benchmarks and evaluation}
\label{sec:evaluation}

Evaluation keeps component performance distinct from campaign-level evidence. Robot, agent, optimizer and laboratory benchmarks assess different operations; a physical claim additionally requires a test of its scientific endpoint. Because $x_t$ is partially observed, a score computed on the outcomes that survive selection and execution is conditioned on those filters. Such a score can be high while the claim it is meant to support remains unsupported, since the filters decide which outcomes enter it. Each endpoint therefore needs its own unit, denominator and measurement reference.

\subsection{What the existing datasets and benchmarks establish}

The datasets and benchmarks sort into seven families (Table~\ref{tab:table5}); a study should cite the family whose question it is actually posing.

\begin{table*}[t]
\caption{Datasets and benchmarks for embodied scientific systems, grouped by the question they answer}
\label{tab:table5}
\centering\fontsize{8.5}{10.2}\selectfont
\setlength{\tabcolsep}{4pt}\renewcommand{\arraystretch}{1.12}
\begin{tabularx}{\textwidth}{@{}YYYY@{}}
\toprule
\rowcolor{Light}\textbf{Dataset or benchmark (examples)} & \textbf{Contribution} & \textbf{What it measures} & \textbf{Scientific question left open}\\
\midrule
Robot training datasets: Open X-Embodiment and DROID \citep{oxe,droid} & Demonstration data for policy and representation training & Training coverage, not scientific fidelity & Whether a policy action produces a valid intervention or measurement\\[3pt]
\addlinespace[1.5pt]
Robot benchmarks: LIBERO and SIMPLER \citep{libero,li2024simpler} & Task suites for knowledge transfer and paired sim--real policy evaluation & Task success and transfer & Whether success implies quantitative delivery, containment, sterility or sample integrity\\[3pt]
\addlinespace[1.5pt]
Policy and world-model evaluation: RoboArena and WorldArena \citep{atreya2025roboarena,shang2026worldarena} & Distributed pairwise real-world policy comparison; perceptual and functional evaluation of world models & Relative policy ranking and proxy agreement with real runs & Whether the ranking holds for a measured physical endpoint rather than a task label\\[3pt]
\addlinespace[1.5pt]
Language-agent benchmark for biology: LAB-Bench \citep{laurent2024labbench} & Multiple-choice questions on literature, figures, databases and sequences & Practical research knowledge and reasoning & Whether a correct answer yields a correct physical intervention\\[3pt]
\addlinespace[1.5pt]
Optimizer benchmarks: Olympus and its extensions \citep{f2021olympus,f2023olympusenhanced} & Controlled problems for comparing selectors & Selector behavior & Whether a selected experiment is reachable, feasible and informative for the target quantity\\[3pt]
\addlinespace[1.5pt]
Agent benchmarks: ScienceAgentBench and CORE-Bench \citep{f2024scienceagentbench,f2024corebench} & Scientific data-analysis and computational-reproduction tasks & Task, code or reproduction outcomes & Whether the artifact is true of, or useful for, a physical system\\[3pt]
\addlinespace[1.5pt]
Laboratory simulation and failure analysis: LabVLA, LabRobFail (preprints) \citep{labvla2026,f2026labrobfail} & Laboratory workflow simulation; robotic failure analysis & Simulated workflow completion; failure detection & Whether modeled workflow or detected failure modes determine real-laboratory validity and outcome\\[3pt]
\bottomrule
\end{tabularx}
\par\vspace{4pt}\begin{minipage}{\textwidth}\fontsize{8}{9.7}\selectfont Rows are grouped by evaluation target. LabVLA and LabRobFail are preprints at the time of writing.\end{minipage}
\end{table*}

Open X-Embodiment supports RT-X training across embodiments, while DROID records real-world manipulation \citep{oxe,droid}. Their scores evaluate policy learning, not experimental validity. Robot benchmarks (Table~\ref{tab:table5}) test planning, generalization and transfer through suite-specific success predicates; those predicates do not establish delivered quantities, sterility or specimen integrity. Nor do elapsed times necessarily expose reset costs, instrument contention or human recovery. A high benchmark score can therefore coexist with infeasible laboratory deployment.

Olympus compares optimization algorithms on shared response data, including mixed-parameter and multiobjective problems \citep{f2021olympus,f2023olympusenhanced}. Atlas connects planning to execution; it is an interface, not a benchmark suite \citep{f2025atlas}. Controlled problems isolate selection, not physical reachability or measurement reliability. Agent task, code and reproduction scores (Section~\ref{sec:mf-methods}) likewise cannot be pooled into a scientific success rate: confirmation requires independently repeated or changed evidence generation.

LabVLA generates demonstrations and evaluates policies in simulated laboratory workflows; it cannot validate mechanisms absent from its simulator \citep{labvla2026}. LabRobFail evaluates robotic failure analysis in chemical laboratories, but detection accuracy on supplied faults does not establish improved campaign reliability \citep{f2026labrobfail}. Report monitor performance and downstream scientific endpoints separately.

Three problems cut across the families. \textit{Splits}: train and test can share physical history, so a random split within one batch or apparatus leaks preparation, calibration and drift; withhold a whole batch, apparatus configuration, protocol family or scientific family, and report adaptation cost. \textit{Leakage}: training exposure can turn apparent generalization into retrieval; use private or post-cutoff tasks and disclose training cutoff and retrieval sources, as BioDiscoveryAgent does with an unpublished dataset \citep{f2024biodiscoveryagent}. \textit{Distribution shift}: report paired agreement on the deployment distribution, not an average; SIMPLER is the clearest manipulation-policy instance, and laboratory simulators require paired agreement on the scientific response, not video resemblance \citep{li2024simpler}. With explicit metric units and denominators, these controls make the seven families interpretable.

RoboArena expands evaluation diversity through double-blind policy comparisons on evaluator-chosen tasks \citep{atreya2025roboarena}; LAB-Bench tests biology-research knowledge through multiple-choice questions \citep{laurent2024labbench}. These measure comparative task performance and reasoning, respectively. Figure~\ref{fig:selector-executor} adds the physical endpoint and measurement reference needed before either can support a campaign claim.

\subsection{Begin with the scientific endpoint}

Fix the scientific decision before evaluating a campaign: reproducible improvement, hypothesis discrimination and coverage against independent observations are distinct endpoints. Pre-register the endpoint, alternatives, measurement process, resources and confirmation rule; these define attempts, valid observations and follow-up. Report dataset prediction, physical selection and complete campaigns separately \citep{volk2024}. Figure~\ref{fig:selector-executor} distinguishes execution, audit, confirmation and claim accounting. State which physical assumptions remain untested.

Robot and laboratory benchmarks diagnose planning, control and transfer under specified conditions; their outcome is generated by the task and simulator or fixture, not by untested mechanisms such as liquid loss, contamination or material change \citep{libero,li2024simpler,labvla2026}. A specialized instrument can support a direct scientific test; a general policy additionally needs measurements establishing its experimental effect.

\subsection{Choose the unit of success before the study}

Fix the unit of success before collection: action tolerance, attempted experiment, valid observation, campaign progress, nominated finding or independently confirmed claim. Changing units afterwards can turn completion into apparent validity or a campaign maximum into apparent discovery. Table~\ref{tab:metrics} joins the reporting units emphasized by Volk and Abolhasani \citep{volk2024} with the corresponding metric definitions. Using the audit decision $\hat v_t$, the valid-observation rate is

\begin{equation}\label{eq:validity}
r_{\mathrm{valid}}=\frac{N_{\mathrm{valid}}}{N_{\mathrm{attempted}}}=\frac{\sum_t\mathbb{1}\{\hat v_t=1\}}{N_{\mathrm{attempted}}}.
\end{equation}

$N_{\mathrm{attempted}}$ includes every trial passing the start criterion, including interrupted and unsuccessful trials. $N_{\mathrm{valid}}$ counts $\hat v_t=1$, including valid negatives and interval-censored responses. Trial status $c_t$ follows Eq.~\ref{eq:trial-status}; claim state $\sigma_k$ follows Eq.~\ref{eq:claim-state}, with unfinished follow-up recorded as pending, not censored. Report validity alongside endpoint quality, calibration and recovery burden. Confirmed findings per budget and per elapsed time have different denominators from $r_{\mathrm{valid}}$ and must be named separately.

\subsection{Metric definitions and privileged labels}

Let $B$ be total declared budget, $U_T^{\mathrm{claim}}$ the indices nominated by campaign end, and $U_T^{\mathrm{claimed}}\subseteq U_T^{\mathrm{claim}}$ those presented as confirmed. Claim objects in $\mathfrak C_T$ carry states $\sigma_k$ and prespecified weights $w_k\geq0$. Confirmed-claim yield per budget (ECYG) counts orthogonally confirmed claims; claim-set overstatement rate (COR) measures the unconfirmed share of those presented as confirmed; claim-weighted utility yield (CSU) weights the yield:

\begin{equation}\label{eq:ecyg}
\mathrm{ECYG}(B)=\tfrac{\lvert\{k\in U_T^{\mathrm{claim}}:\sigma_k=\OC\}\rvert}{B}.
\end{equation}

The overclaim rate is defined only when $\lvert U_T^{\mathrm{claimed}}\rvert>0$.

\begin{equation}\label{eq:cor}
\mathrm{COR}=\frac{\lvert\{k\in U_T^{\mathrm{claimed}}:\sigma_k\neq\OC\}\rvert}{\lvert U_T^{\mathrm{claimed}}\rvert}.
\end{equation}

\begin{equation}\label{eq:csu}
\mathrm{CSU}(B)=\tfrac{\sum_{k\in U_T^{\mathrm{claim}}}w_k\,\mathbb{1}\{\sigma_k=\OC\}}{B}.
\end{equation}

For controlled injection benchmarks, let $\mathcal T_{\mathrm{audit}}$ contain $N_{\mathrm{audited}}$ trials carrying both $\hat v_t$ and reference label $v_t^{\star}$. The latter records compliance with prespecified, protocol-relative criteria: tolerance, detection-limit handling, contamination checks and reference measurements. An injected fault invalidates a response only if it compromises the target under those criteria. Report the reference standard, label-assignment procedure and uncertainty. Audit agreement is

\begin{equation}\label{eq:da}
\mathrm{DA}=\frac{1}{N_{\mathrm{audited}}}\sum_{t\in\mathcal T_{\mathrm{audit}}}\mathbb{1}\{\hat v_t=v_t^{\star}\}.
\end{equation}

\begin{equation}\label{eq:irr}
\mathrm{IRR}=\frac{\sum_{t\in\mathcal T_{\mathrm{audit}}}\mathbb{1}\{\hat v_t=0,\ v_t^{\star}=0\}}{\sum_{t\in\mathcal T_{\mathrm{audit}}}\mathbb{1}\{v_t^{\star}=0\}}.
\end{equation}

The audit decision accuracy (DA) measures agreement between $\hat v_t$ and the reference label, and the invalid-record rejection rate (IRR) measures the share of reference-invalid records that the audit rejects; both require $v_t^{\star}$ and nonzero denominators. Campaigns without reference labels report $\hat v_t$, $c_t$, audit rules and raw evidence. Report COR with coverage $|U_T^{\mathrm{claimed}}|/|U_T^{\mathrm{claim}}|$; undefined ratios are not zero. For comparable ECYG/CSU, nominated indices must refer to a prespecified catalog with frozen granularity, endpoints and reference criteria. Nomination can enlarge the active index set, but post-hoc claim splitting, merging or weight rescaling is disallowed. Fix nonnegative weights on a common scale, for example $\sum_k w_k=|U_T^{\mathrm{claim}}|$; unit weights make CSU equal ECYG. These ratios require a positive scalar budget $B$ in stated units; otherwise report the resource vector and tradeoffs, or declare its scalar conversion. Comparisons retain the same resource boundary, catalog and weights. Table~\ref{tab:metrics} summarizes inputs and truth requirements.

\begin{table*}[t]
\caption{Units, metrics and evidence requirements for evaluating an embodied scientific campaign.}
\label{tab:metrics}
\centering\fontsize{8.1}{9.8}\selectfont
\setlength{\tabcolsep}{4pt}\renewcommand{\arraystretch}{1.13}
\begin{tabularx}{\textwidth}{@{}YYYY@{}}
\toprule
\rowcolor{Light}\textbf{Unit of work} & \textbf{Quantity to report} & \textbf{Required record / denominator} & \textbf{Interpretation}\\
\midrule
Action & Tolerance error, completion, collision and containment & Operation specification and independent physical reference & Completion establishes execution, not observation validity.\\[3pt]
\addlinespace[1pt]
Attempt / valid observation & $r_{\mathrm{valid}}=N_{\mathrm{valid}}/N_{\mathrm{attempted}}$ (Eq.~\ref{eq:validity}) & Every started trial; $\hat v_t$, $c_t$, missing and inconclusive outcomes & Valid positives, negatives and censored responses enter the valid pool; no privileged truth is required.\\[3pt]
\addlinespace[1pt]
Measurement & Calibration error, noise, detection limit and raw-data availability & Reference material, calibration version and processing version & Repeated readings measure repeatability; independent preparations assess preparation variability.\\[3pt]
\addlinespace[1pt]
Recovery & Interventions, recovery time, sample loss and remakes & Fault, corrective action, human work and parent--child lineage & Report whether recovery preserved specimen integrity and protocol validity.\\[3pt]
\addlinespace[1pt]
Campaign / confirmed yield & ECYG and CSU (Eqs.~\ref{eq:ecyg}, \ref{eq:csu}); endpoint versus attempts, time and resources & Nominated catalog, $\sigma_k$, fixed weights $w_k$ and total budget $B$ & Confirmed yield uses all declared costs; CSU weights are fixed before comparison.\\[3pt]
\addlinespace[1pt]
Claims presented as confirmed & COR (Eq.~\ref{eq:cor}) and reporting coverage & $U_T^{\mathrm{claimed}}\subseteq U_T^{\mathrm{claim}}$ and their adjudicated states & Unresolved claims remain non-confirmed; report the nominated population separately.\\[3pt]
\addlinespace[1pt]
Transfer & Held-out endpoint agreement and adaptation cost & Named specimen, apparatus, protocol, domain or laboratory change & Fix the equivalence margin in advance; charge adaptation data, labor and calibration.\\[3pt]
\addlinespace[1pt]
Controlled audit benchmark & DA and IRR (Eqs.~\ref{eq:da}, \ref{eq:irr}) & Audit decision $\hat v_t$ and protocol-relative reference $v_t^\star$ & Privileged reference labels are required; these are not unlabeled online campaign metrics.\\[3pt]
\addlinespace[1pt]
\bottomrule
\end{tabularx}
\par\vspace{3pt}\begin{minipage}{\textwidth}\fontsize{7.7}{9.3}\selectfont Ratios are undefined when their denominators are zero. ECYG and CSU require a positive scalar budget in stated units or a declared resource conversion. Claim granularity, endpoints, confirmation criteria and weights are frozen before comparison.\end{minipage}
\end{table*}

\subsection{Attribute gains to the correct component}

\begin{figure*}[t]
\centering
\includegraphics[width=\textwidth,keepaspectratio]{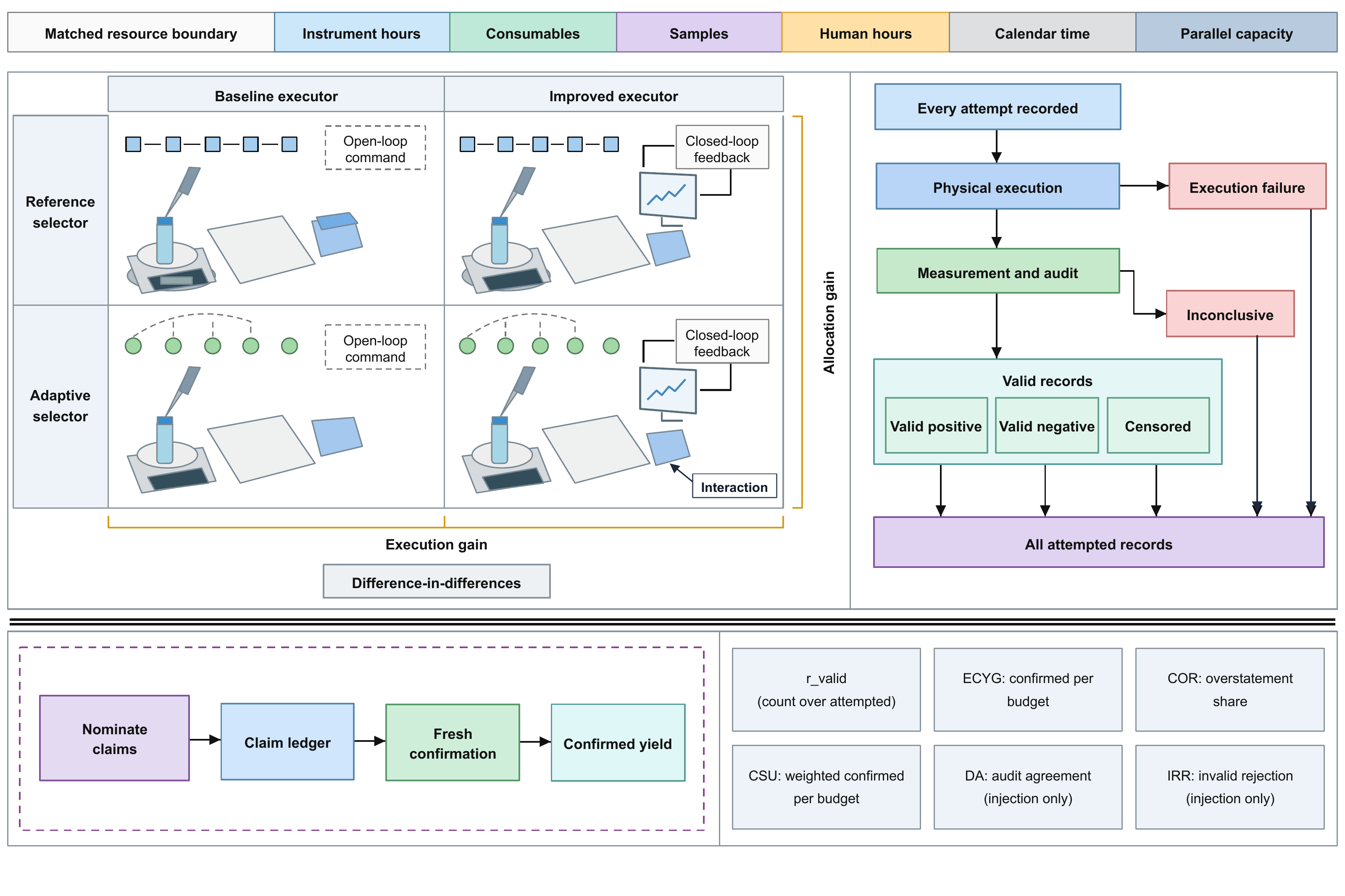}
\caption{Evaluation design for embodied scientific campaigns. A strict two-by-two comparison crosses a reference or adaptive selector with a baseline or improved executor under matched endpoint, information, resource boundary and acceptance criteria. Every attempt enters the physical execution gate and remains in $N_{\mathrm{attempted}}$. The metrological and audit gate assigns $\hat v_t$ and $c_t$: valid positive, valid negative and censored records all count in $N_{\mathrm{valid}}$, whereas execution failures and inconclusive records remain in $N_{\mathrm{attempted}}$ but not $N_{\mathrm{valid}}$. Nominated claims form a separate unit with counts $N_{\mathrm{sel}}$ and $N_{\mathrm{conf}}$; confirmation uses fresh evidence not used for nomination under prespecified criteria. $r_{\mathrm{valid}}$, ECYG, COR and CSU use their declared denominators; DA and IRR require a privileged reference label and apply only to controlled injection benchmarks. The apparatus is a synthetic illustrative scene.}
\label{fig:selector-executor}
\end{figure*}

End-to-end gains confound experiment choice and execution. The two-by-two design in Figure~\ref{fig:selector-executor} crosses reference/adaptive selectors with baseline/improved executors under the same endpoint, initial information, audit rule and resource boundary. Within-selector contrasts estimate execution gain; within-executor contrasts estimate allocation gain; the combined comparison estimates interaction. Same-robot and common-schedule tests isolate selection and control, respectively, but complete campaigns remain necessary because execution changes feasibility and affordability. Retain every attempt and report validity, nomination, confirmation and uncertainty with their respective denominators.

Use a matched resource vector, not experiment count alone: instrument hours, consumables, samples, human hours, calendar time and parallel capacity. Count setup, cleaning, calibration, analysis and failure recovery inside a declared boundary; report capital and development labor separately. Campaign timing illustrates the distinction: a campaign can contain short operating rounds and long interruptions, and the two durations answer different questions. operating duration measures active procedure time, while calendar duration measures elapsed access and downtime. Multi-fidelity evaluation, unknown-feasibility optimization and active oversight motivate recording fidelity cost, infeasible trials and raw-data exceptions alongside the campaign endpoint \citep{f2025mfbobest,f2025anubis,f2025oversight}.

\subsection{Acceleration factors, autonomy labels and transfer axes}

An acceleration factor is interpretable only when numerator, denominator and counted work are fixed. Parallel instruments, easier assays, proxy endpoints, omitted setup or excluded human labor can raise a ratio while confirmed findings per resource fall. Report the attempt/valid/claim/campaign denominators of Table~\ref{tab:metrics} with operating and calendar time, resource vector and human work; show unmatched instrument hours or parallel capacity separately from any speedup. Autonomy taxonomies describe delegated responsibility, not evidential strength: L0--L5 and facility taxonomies identify who or what selected, executed or interpreted an operation, but not whether a finding is valid or confirmed \citep{f2024communitysurvey,f2026autonomyfacilities}. Record autonomy as a profile alongside Table~\ref{tab:metrics}.

Separate five transfer axes: specimen, apparatus, protocol, scientific domain and laboratory. Withhold the unit before training to avoid shared-history leakage, and report adaptation data, calibration, expert labor and endpoint agreement. Batch transfer holds apparatus and assay fixed; apparatus transfer uses matched preparations and references; protocol transfer documents changed operations; domain transfer specifies the relation expected to persist; laboratory transfer records local calibration and operator work. A bundled multi-axis test needs separate effects to make failure interpretable.

An illustrative design template for a laboratory-transfer test prespecifies the endpoint, an equivalence margin $\Delta>0$, and a decision rule---declare transfer supported when the two-sided $(1-\alpha)$ confidence interval for the between-platform difference lies within $(-\Delta,\Delta)$---and fix replicate counts and target power prospectively before data collection; these choices are campaign-specific, not defaults implied by this review.

\subsection{Confirmation and failure attribution}

Confirmation challenges a nominated finding with fresh evidence under a prespecified endpoint, criterion, margin and failure rule. Independent preparations, orthogonal assays and held-out conditions or laboratories address different errors; shared dependencies must be recorded. The selected maximum benefits from favorable noise, so reusing nomination data is not confirmation. The 28-workflow evidence map (Appendix~\ref{sec:appendix}) records reported O/F/OF/P/N/? pathways, separately from formal claim-level assessment $\sigma_k$.

SAMPLE re-tested designed enzymes in \textit{E. coli} lysate under different expression and assay conditions \citep{sample2024}, and RoboChem reports inline NMR yields alongside its automated analysis \citep{robochem2024}. The two checks differ in what changes between measurements, so a single validation flag would conflate them. The A-Lab correction shows why reassessment can narrow an accepted outcome while confirming the remainder: four of 40 reported successes were inconclusive after reanalysis, and the novelty wording narrowed to materials new to the prediction platform rather than necessarily new to science \citep{alab2023,alabcorr2026}.

\begin{figure*}[t]
\centering
\includegraphics[width=\textwidth,keepaspectratio]{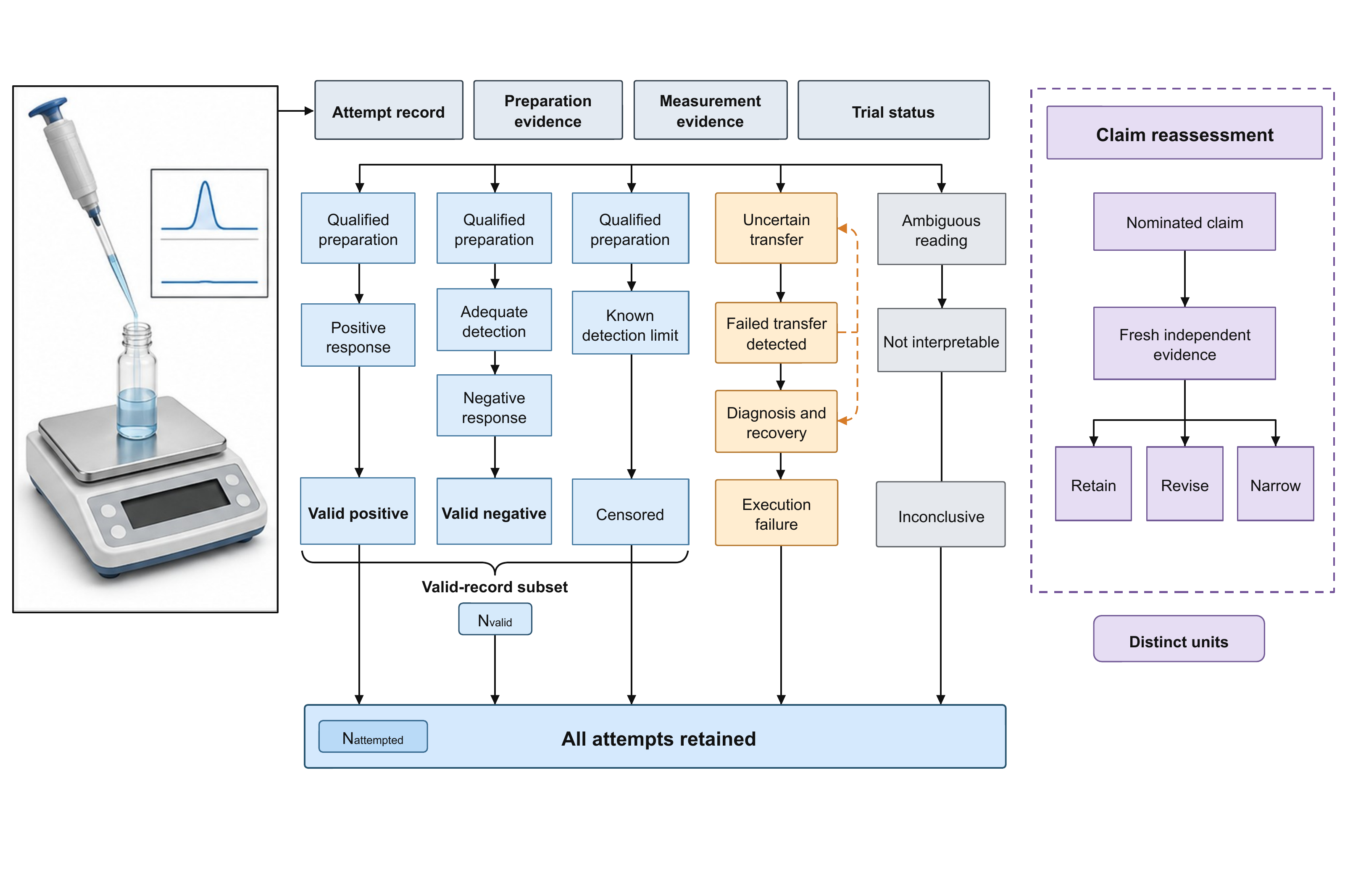}
\caption{Failure attribution and claim reassessment. A low response with qualified preparation and adequate detection can be a valid negative; valid positive, valid negative and censored records remain in the valid pool and count in $N_{\mathrm{valid}}$. Uncertain transfer produces an execution failure that remains in $N_{\mathrm{attempted}}$ and returns through diagnosis and recovery, while an unclear signal is inconclusive. These trial statuses $c_t$ are distinct from claim state $\sigma_k$. Independent confirmation of a nominated claim can retain, revise or narrow it according to prespecified evidence; narrowing is not the only permissible outcome. The apparatus is a synthetic illustrative scene.}
\label{fig:failure-confirmation}
\end{figure*}

Figure~\ref{fig:failure-confirmation} distinguishes trial interpretation from claim confirmation. Qualified preparation and adequate detection can make a low response a valid negative; an execution fault or unclear signal instead triggers failure or inconclusive status. Every attempt remains in history and can inform the full-record posterior (Eq.~\ref{eq:joint_belief}), even when excluded from qualified-response summaries. Valid positives, negatives and censored responses enter $N_{\mathrm{valid}}$, with censoring handled by interval likelihood; inconclusive outcomes remain separate. Fresh confirmation can retain, revise or narrow a nominated claim without changing this trial-status distinction.

\subsection{Minimum campaign record}\label{sec:minimum-record}

A reconstructable campaign record has three parts. Before collection, specify the scientific question, alternatives, endpoint and acceptance criterion; initial data, models and search space; allowed operations and physical preconditions; selector and executor versions; apparatus configuration, calibration and operating envelope; and the resource boundary, including instrument time, consumables, samples, human work, calendar time and parallel capacity.

For every attempt, retain commanded and realized conditions, specimen lineage, raw signals, calibration and detection limits, processing versions, failures, inconclusive outcomes, retries and censoring. Record human roles, interventions and authorizations, alongside the data schema, code versions, provenance and safety exceptions. These fields expose differences in physical execution, information and resources rather than attributing all variation to the selector.

For every claim, record the nomination set, independent confirmation rule, preparation and assay dependencies, follow-up status, held-out transfer unit and adaptation cost. The linked trial and claim records make it possible to reconstruct the experiment and distinguish a selected outcome from a finding that survives a fresh test.

\subsection{Adaptive stopping and prospective comparison}

Report each rate's population and follow-up status. Unfinished claim follow-up is $\sigma_k=\mathrm{pending}$, neither success nor failure; $\mathrm{censored}$ denotes only a valid interval response (Section~\ref{sec:formulation}). Searching noisy candidates favors optimistic maxima, and repeated threshold checks create optional-stopping error. Prespecify stopping and use independent confirmation or a policy-valid sequential procedure. A selected winner's posterior probability is not a confirmation rate. Match replication to the estimand: repeated readings measure repeatability, not between-preparation variation.

For a finite-look stopping design, prespecify $n_{\mathrm{look}}$ interim looks and valid per-look levels $\alpha(j)$, $j=1,\ldots,n_{\mathrm{look}}$, with $\sum_j\alpha(j)\le\alpha_{\mathrm{total}}$; reserve the remainder for the final test. A union-bound guarantee requires each look's test to remain valid under the campaign's selection policy, with any within-look multiplicity accounted for. Prespecifying a schedule alone cannot validate adaptively selected tests.

A prospective comparison tests whether selection or execution improves science. Before allocation, prespecify the question, endpoint, selector, executor, eligible experiments, initial information, resource vector, invalidity rules, stopping rule and confirmation procedure (Section~\ref{sec:minimum-record}). Randomize matched campaign replicates across the four selector-by-executor cells of Figure~\ref{fig:selector-executor}, hold endpoint, information and resources constant, charge confirmation trials to the same budget, and retain every attempt with realized conditions, raw signals, exceptions, decision trace and human work. Estimate selector, executor and interaction contrasts on valid-observation rate and confirmed findings per resource, with campaign-level uncertainty. Report each simple effect, the main effects and the difference-in-differences on the declared endpoint scale; if the endpoint is nonlinear, state the transformed scale. A rise in task completion or valid rate alone is an operational gain, not a scientific-evidence gain. Olympus can compare selectors on controlled problems, but a benchmark rank remains separate from a physical campaign outcome \citep{f2021olympus}; simulation can develop hypotheses but cannot establish mechanisms absent from the model.

\section{Reproducibility, safety and infrastructure}\label{sec:reproducibility}

\subsection{Provenance as part of the experimental model}

A physical provenance record is part of the observation model: without it, the reported measurement cannot be linked to the specimen and conditions that produced it. FAIR principles require findable, accessible, interoperable and reusable data \citep{fair2016}; embodied systems additionally need material lineage and realized conditions that a final property table cannot recover.

Each field addresses one distinct reconstruction failure, and it does so only when it is recorded for the trial it describes. A stable sample and batch identifier, with parent-child links for aliquoting, pooling, dilution, purification, consumption and replacement, lets a reader detect observations assigned to the wrong material and repeated readings mistaken for independent preparations. The protocol version and machine-readable units expose changed operations or unit conversions that a workflow name would otherwise hide. Device identity, configuration and operating envelope make hardware differences visible, so they are not attributed to the scientific method. Calibration certificates, reference standards and validity intervals let a later reader distinguish an instrument offset from a physical effect. Timestamps and intended as well as realized values expose drift, delays and deviations between command and intervention. Raw signals preserve information that summary values discard; processing code, version and parameters make analysis changes visible, not presented as new data. Exceptions, retries, censoring and failed recovery keep unsuccessful runs in the denominator. Human actions, decision trace, role and authorization distinguish expert intervention from unattended operation and identify who made an irreversible decision. These fields should be linked to each reported outcome, not collected as unstructured campaign notes.

Chemical programming and software abstractions separate reusable experimental intent from platform-specific execution; hardware interfaces and orchestration address instrument and workflow coordination \citep{rauschen2024,pylab2023,chemos2}. AlabOS provides reconfigurable workflows and resource reservation \citep{f2024alabos}. These mechanisms establish semantic comparability of named steps and fields, not physical equivalence. Two devices can compile the same protocol and still deliver different volumes, temperature histories or sample transformations because actuators, calibration and material histories differ. Physical equivalence therefore requires traceable calibration and comparison of realized conditions and outcomes on reference materials or matched preparations. Report replay on the same apparatus, repeatability across preparations and transfer across laboratories as separate tests; successful compilation alone cannot establish any of them.

The replay obligation is best read as three groups of records. The execution trace---scientific intent and protocol version, command and decision trace, realized conditions and deviations, and raw observations with trial status---establishes what was intended, commanded, realized and observed. Calibration and provenance---specimen lineage, device identity and calibration, units, processing versions and reference materials, and exceptions, retries and human authorizations---connect those records to calibration, material lineage and human responsibility. Cross-site replay---the shared schema and software versions, local adaptations and operator work, and held-out replay outcomes against a prespecified equivalence margin---tests whether another site can interpret and replay the workflow after declared adaptation. A replay claim is complete only when the relevant items are present and the endpoint, tolerance and equivalence margin are stated. The same source-linked coding covers the 28-workflow cohort described in Section~\ref{sec:applications} and Appendix~\ref{sec:appendix}.

\subsection{Uncertainty and model drift}

Drift must be detected from measurements, not inferred from an optimizer's confidence. Calibration is empirical \citep{guo2017}, and conformal coverage relies on assumptions such as exchangeability that adaptive selection and changing apparatus can violate \citep{conformal2021}. Interleave stable reference materials, blanks and controls with campaign samples, retain replicate measurements, and track calibration residuals and instrument state over time. Declare a trigger before the run, such as a control-chart limit or a residual shift. On a trigger, pause optimization, quarantine observations since the last valid check, recalibrate, and measure references again before resuming. Report which records were invalidated and whether endpoint conclusions change when they are excluded. Active oversight's second Gaussian-process model uses real-time raw-data assessment to constrain selection and allows human intervention for unexpected results \citep{f2025oversight}; this supports in-loop quality control, while reference measurements remain necessary to detect shifts the model does not represent. Since adaptive selection changes the sampled distribution, calibration and coverage must also be checked across relevant regimes or on held-out controls, not only summarized by aggregate confidence.

\subsection{Safety and human responsibility}

Equipment safety and biosecurity require separate controls because they prevent different harms. Equipment safety concerns physical hazards from apparatus, environment, materials and control: identify hazardous states and exposure pathways, then define prevention, monitoring, safe stop and recovery authority. Safe-SDL makes this an architectural requirement as accessible experiments expand \citep{leong2025}. Chemist Eye illustrates an independent perceptual channel for workspace monitoring; it must be evaluated as a detector, including false negatives and its control interface, because a monitor that misses a hazard does not make the apparatus safe \citep{f2026chemisteye}. Control barrier functions enforce modeled state constraints, but cannot cover hazards absent from the model or replace scientific-quality checks such as contamination and specimen viability \citep{cbf2017}.

Biosecurity instead concerns misuse of biological capabilities and information, including protocols that can be released without a robot executing them. A physical interlock cannot prevent harmful information release; the record must state in-scope requests, escalation to human review, reachable models and databases, release decisions and accountable roles. Keeping these risk models distinct prevents a laboratory's collision controls from being presented as a complete biosecurity policy. For both, human authority over goals, scope and ambiguous outcomes must be explicit, logged and included in the claimed operating envelope.

\subsection{Shared infrastructure and access}

A shared resource should publish a validated capability statement, not a generic robot endpoint. The statement must identify controllable and measurable scientific variables, validated ranges and uncertainty, preconditions, specimen requirements, calibration status, known failure modes, resource costs, endpoint definitions, applicable protocols and transfer evidence. It should also distinguish public metadata from restricted raw data or hazardous procedures, and specify access authorization, audit logs and responsibility for remote actions. These fields let a planner distinguish “no effect observed” from “this apparatus could not test the effect,” and let another laboratory decide whether its intervention and measurement meet the same scientific definition. Community infrastructure discussions identify access and coordination as practical issues, but remote availability alone does not establish reproducibility \citep{f2024communitysurvey}. Cross-site use must report local adaptations, operator time, failed runs and agreement against a prespecified equivalence margin on shared references or matched preparations. Without those checks, access expands reach but not the evidential scope of the result.

\section{Open problems and research directions}\label{sec:priorities}

\begin{figure*}[t]
\centering
\includegraphics[width=\textwidth,keepaspectratio]{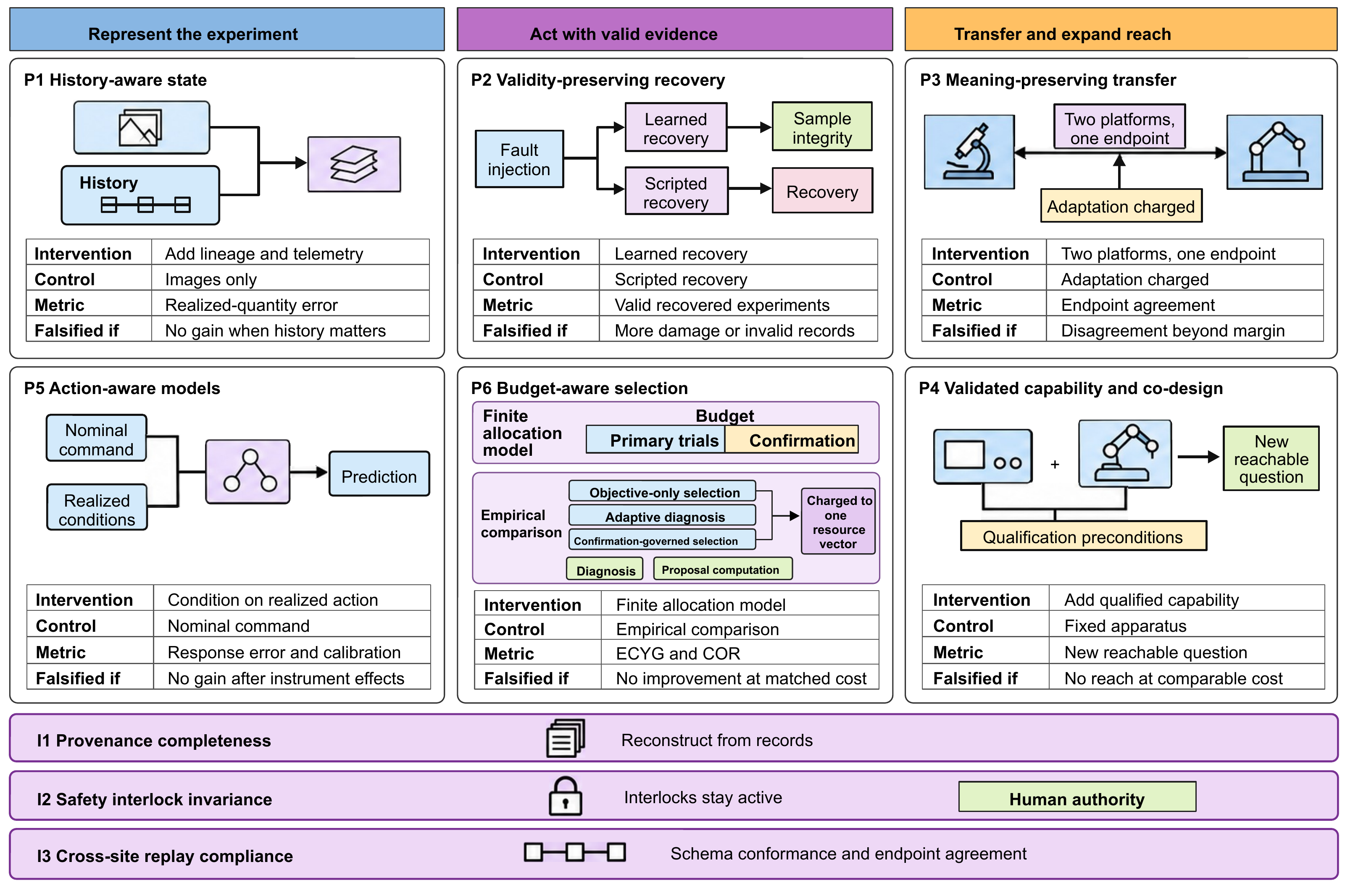}
\caption{Six empirical propositions grouped into three complementary research thrusts, with three cross-cutting normative invariants. P1 and P5 concern histories and realized actions; P2 and P6 concern validity-preserving recovery and matched-budget selection, diagnosis and confirmation; P3 and P4 concern endpoint-preserving transfer and qualified expansion of experimental reach. Provenance completeness, safety with human authority, and cross-site replay compliance must hold across all propositions. Grouping does not imply a temporal or causal order, predicted gains or established findings. Scientific miniatures are conceptual and apparatus imagery is synthetic.}
\label{fig:roadmap}
\end{figure*}

The research agenda is organized into six falsifiable empirical propositions (P1--P6) and three normative architectural and reporting invariants (I1--I3) (Table~\ref{tab:table7}). Each proposition specifies an independent intervention, a measurable dependent metric from the evaluation tables, a matched resource boundary and a falsification condition. The invariants specify properties that a scientific workflow must preserve when its selector, executor or deployment site changes. The 26-workflow physical synthesis, with two software-boundary comparators, provides an empirical benchmark grounding the research agenda; the propositions specify comparisons that can be adapted to domain-specific endpoints, while the invariants specify reporting and architectural requirements. Figure~\ref{fig:roadmap} groups the agenda into three complementary research thrusts: representing the experiment (P1, P5), acting with valid evidence (P2, P6), and transferring meaning while expanding reach (P3, P4). These are not chronological forecasts. The three invariants apply across all six propositions; Table~\ref{tab:table7} specifies their interventions, metrics and falsification or violation conditions.

\begin{table*}[t]
\caption{Controlled research comparisons and cross-cutting invariants.}
\label{tab:table7}
\centering\fontsize{8.1}{9.8}\selectfont
\setlength{\tabcolsep}{4pt}\renewcommand{\arraystretch}{1.13}
\begin{tabularx}{\textwidth}{@{}>{\raggedright\arraybackslash}p{.7cm}YYY@{}}
\toprule
\rowcolor{Light}\textbf{Item} & \textbf{Intervention / comparison} & \textbf{Dependent measurements} & \textbf{Falsification / violation}\\
\midrule
P1 & Images alone vs. telemetry vs. lineage/history; same robot, selector and capacity; hold out batches. & Valid rate, realized-quantity error, held-out endpoint agreement. & Precise estimates exclude the prespecified useful history gain.\\[3pt]
\addlinespace[1pt]
P2 & Scripted vs. learned recovery under matched faults, starts and total budget. & Valid recovered trials, time, sample loss, human burden, ECYG. & Similar recovery with greater damage, invalidity or burden, or lower confirmed yield.\\[3pt]
\addlinespace[1pt]
P3 & Two nonidentical platforms; matched preparations and calibration; charge adaptation. & Endpoint agreement, adaptation data/labor/calibration, COR. & Endpoint difference exceeds the prespecified equivalence margin after adaptation.\\[3pt]
\addlinespace[1pt]
P4 & Existing vocabulary/apparatus vs. one qualified capability or co-designed configuration. & New reachable question, valid endpoint, transfer and total cost. & No useful reach or valid endpoint at comparable cost, or simulated benefit fails physical validation.\\[3pt]
\addlinespace[1pt]
P5 & Intended-condition vs. realized-condition vs. measurement-aware models; matched data/capacity; hold out interventions. & Physical-response error, alternative discrimination, uncertainty calibration and horizon. & No gain after realized actions and instrument effects are controlled.\\[3pt]
\addlinespace[1pt]
P6 & First isolate finite primary/confirmation allocation; then compare objective-only, diagnostic and confirmation-aware policies with all costs charged. & ECYG, COR, modeled ICY, unresolved outcomes, infeasible proposals and full resource vector. & No prespecified evidence/cost gain; confirmation harms the specimen or biases blinded assessment.\\[3pt]
\addlinespace[1pt]
I1 & Replay each selector/executor variant from protocol, device, calibration, lineage, raw signals, exceptions and claim links. & Required-field coverage, replay success and traceable-claim fraction. & Missing fields prevent reconstruction of realized experiments or claim support.\\[3pt]
\addlinespace[1pt]
I2 & Change policy, executor, model or site while retaining interlocks and human authorization. & Interlock violations, safe-stop latency and authorized escalation. & Any change disables an interlock, bypasses escalation or leaves an unsafe state.\\[3pt]
\addlinespace[1pt]
I3 & Replay at a second site with a shared schema and matched preparations or held-out references; disclose local adaptation. & Schema conformance, endpoint agreement and adaptation cost. & Missing fields, unreported changes or endpoint disagreement beyond the margin.\\[3pt]
\addlinespace[1pt]
\bottomrule
\end{tabularx}
\par\vspace{3pt}\begin{minipage}{\textwidth}\fontsize{7.7}{9.3}\selectfont P1--P6 are empirical propositions; I1--I3 are normative invariants. Metric definitions and denominator requirements appear in Table~\ref{tab:metrics}. These are proposed tests, not established gains.\end{minipage}
\end{table*}

\subsection{P1: Represent the physical history needed for inference}

The first proposition is that including specimen lineage and execution history improves inference when nominally identical-looking observations arise from different physical histories. CrossFormer trains across 20 embodiments, showing that heterogeneous robot data can support a shared policy; the laboratory test additionally asks whether the state contains the history needed to interpret a specimen \citep{doshi2025crossformer}. Compare images alone, images plus telemetry, and those inputs plus specimen lineage, holding robot, selector and model capacity fixed while withholding complete batches. Measure valid-observation rate, realized-quantity error and held-out-batch endpoint transfer. If a sufficiently precise comparison excludes the prespecified useful gain, the proposed benefit is falsified for the tested task domain.

\subsection{P2: Learn recovery that preserves sample validity}

Learned recovery is useful only if it restores an experiment without damaging the specimen or concealing the failure. Failure-detection benchmarks and policies trained from deployment experience or expert corrections show that recognition and correction can be studied as capabilities \citep{duan2025aha,zhang2026foresight,chen2025conrft}. A laboratory recovery test adds sample integrity, audit-adjudicated validity and claim-level yield to task completion. Compare replayed or scripted recovery with a learned recovery policy under matched injected faults, holding executor and starting conditions fixed. Measure valid recovered experiments, interruption time, sample loss, human intervention and downstream ECYG. If recovery success is similar but learned recovery increases burden, damage or invalid records, the proposition is falsified.

\subsection{P3: Transfer experiments across hardware and laboratories}

Scientific transfer means preserving an endpoint and its interpretation across a declared change, not merely moving a policy to another body. CrossFormer and OpenVLA-OFT study cross-embodiment representations or adaptation and report robot-task performance; they motivate a laboratory transfer test with the scientific endpoint, calibration and equivalence margin held explicit \citep{doshi2025crossformer,kim2025openvlaoft}. Run the same scientific question and endpoint on two nonidentical platforms, specify an equivalence margin before data collection, and charge adaptation data, expert labor and calibration. Measure endpoint agreement and all adaptation costs separately. Agreement inside the margin supports transfer only for that changed platform; disagreement beyond it after adaptation counts against the transfer proposition. If the task is preserved but measurement semantics change, the comparison is not a transfer success.

\subsection{P4: Expand the experiment vocabulary through validated capabilities and co-design}

A new capability can add value by reaching a previously inaccessible question or by improving the quality or cost of an existing measurement; the test should specify which gain is claimed. A callable tool can increase interface breadth without changing the set of reachable interventions or interpretable measurements. Qualify one new intervention or measurement by documenting preconditions, uncertainty and failure modes, then compare the system with and without it on a question selected before the test. Jointly designing apparatus and experiment can improve reach when the optimized configuration solves a physical constraint that the fixed setup cannot meet. AI-designed-physics work formulates hardware configuration under practical constraints \citep{physics2026}; a prospective co-design test must include fabrication, calibration and physical endpoint assessment. Measure reachable questions, valid observations, endpoint reproducibility, transfer and total validation and operating cost. The proposition is falsified if the new interface or co-designed apparatus produces no new reachable question or valid endpoint at comparable cost, or if the apparent gain depends on an unvalidated measurement.

\subsection{P5: Couple scientific models to realized physical actions with transparent assumptions}

An action-aware scientific model should predict measured response from the intervention actually executed. A nominal command or a plausible-looking visual trajectory is not a substitute. Robot world models and action-conditioned video models predict future states or images, while sparse equation discovery offers a structured representation of response; these strands motivate models that separate apparatus transition, specimen evolution and measurement \citep{hafner2025dreamerv3,cen2025worldvla,sindy2016}. The comparison should measure concentration, temperature history or another declared physical response under held-out interventions. Compare models conditioned on intended conditions, realized conditions and measurement configuration using matched data and capacity, then hold out interventions. Measure physically meaningful response error, discrimination between alternatives, uncertainty calibration and horizon. If action-aware models do not outperform after instrument effects are controlled, the claimed benefit of coupling action to the scientific model is falsified.

\subsection{P6: Allocate a matched budget across diagnosis, confirmation and selection}

Adaptive diagnosis should improve confirmed evidence per resource when a calibration, blank, remeasurement or recovery decision can resolve ambiguity before another experiment is spent. Active oversight uses real-time raw-data assessment and human intervention for unexpected outcomes, while AHA detects and reasons about manipulation failures using perturbed examples \citep{f2025oversight,duan2025aha}. We first formulate an analytical two-way confirmation allocation model, subsequently extending the analysis to adaptive diagnosis, proposal computation, and sequential selection under matched resource constraints.

Let $N\in\mathbb N_{>0}$ be a fixed positive total budget in trial-equivalent units, $K\in\{0,\ldots,N\}$ the number of paid confirmation attempts, and $N-K$ the number of primary trials. Let $\gamma\in[0,1]$ be the probability that a primary trial nominates an interpretable candidate, with independent candidate generation in this analytical model. Then $C_K\sim\mathrm{Binomial}(N-K,\gamma)$. Without oracle knowledge of which nominations are true, select $M_K=\min(C_K,K)$ candidates uniformly for independent confirmation; each attempted confirmation costs one unit whether its result is positive, negative or invalid. Let $p_{\mathrm{drift}}\in[0,1]$ be the probability a selected candidate is invalid under the declared claim criterion, $s_{\mathrm{conf}}\in[0,1]$ the confirmation sensitivity on non-invalid candidates, and $\varphi\in[0,1]$ the false-pass probability on invalid candidates. Assume independent, homogeneous candidate validity and confirmation outcomes, with true-pass probability $\mathrm{TP}=(1-p_{\mathrm{drift}})s_{\mathrm{conf}}$ and false-pass probability $\mathrm{FP}=p_{\mathrm{drift}}\varphi$. A status-positive confirmation is therefore a reported orthogonally-confirmed claim, but not necessarily a true claim.

The finite-sample expectation of the number of candidates that can actually be attempted is
\[
\begin{aligned}
\mathbb E[M_K]
&=\sum_{c=0}^{N-K}\min(c,K)\binom{N-K}{c}\gamma^c(1-\gamma)^{N-K-c},\\
Y_{\mathrm{status}}(K)
&=\frac{\mathrm{TP}+\mathrm{FP}}{N}\,\mathbb E[M_K],\\
Y_{\mathrm{true}}(K)
&=\frac{\mathrm{TP}}{N}\,\mathbb E[M_K].
\end{aligned}
\]
Here $Y_{\mathrm{status}}$ is expected status-confirmed yield per total budget, the model counterpart of ECYG at the level of reported status rather than true validity; $Y_{\mathrm{true}}$ is genuine true-confirmed yield per total budget. The binomial sum is the exact expectation: replacing it by $\min\{\mathbb E[C_K],K\}$ would be only an upper bound, not an equality. No selected candidate is assumed known to be a true survivor before confirmation.

For a common counterfactual benchmark, define ideal-confirmation yield (ICY) as the true potential yield if every pre-confirmation nomination under either allocation received a free, independent assay with perfect sensitivity and no false passes. Only valid candidates then survive, so the sensitivity $s_{\mathrm{conf}}$ and the false-pass probability drop out. This is a model benchmark, not an operational claim state and not an estimate from unlabeled online data:
\begin{align*}
\mathrm{ICY}_{\mathrm{unconf}}&=\gamma\,(1-p_{\mathrm{drift}}),\\
\mathrm{ICY}_{\mathrm{conf}}(K)&=\gamma\frac{N-K}{N}\,(1-p_{\mathrm{drift}}).
\end{align*}
ICY isolates the opportunity cost of spending $K$ trial-equivalent units on confirmation instead of primary screening; it neither charges the realized confirmation attempts nor proves that confirmation-governed selection improves candidate quality. Actual reported outcomes use ECYG, COR and claim-level adjudication. Let $R_K$ be the number of status-positive confirmation reports. If $\mathrm{TP}+\mathrm{FP}>0$ and $P(R_K>0)>0$, the expected false-discovery fraction conditional on at least one report is $\mathrm{FP}/(\mathrm{TP}+\mathrm{FP})$ under the stated homogeneous independent-pass model; the ratio of expected false reports to expected reports is also $\mathrm{FP}/(\mathrm{TP}+\mathrm{FP})$ here, though the two estimands need not agree under heterogeneous candidate-specific pass probabilities. If $\mathrm{TP}+\mathrm{FP}=0$ or the event $R_K>0$ has probability zero, the conditional fraction is undefined and reported as not applicable. This model-based false-discovery fraction is distinct from COR: COR compares claims presented as confirmed with their adjudicated $\sigma_k$, whereas empirical invalid fractions require declared blinded claim-level references or privileged truth labels.

Choose the confirmation allocation by the exact finite integer optimization
\[
K^*\in\arg\max_{K\in\{0,\ldots,N\}}Y_{\mathrm{true}}(K).
\]
When $\mathrm{TP}>0$, the factor $\mathrm{TP}$ is constant in $K$, so both $\arg\max_K Y_{\mathrm{true}}(K)$ and $\arg\max_K Y_{\mathrm{status}}(K)$ equal $\arg\max_K \mathbb E[M_K]$: the optimum depends on the candidate-nomination rate $\gamma$ and budget $N$ alone, while $p_{\mathrm{drift}}$, $s_{\mathrm{conf}}$ and $\varphi$ scale the yield level without moving $K^*$. Estimate $\gamma$ on an independent pilot or over a prespecified grid; break ties by the smallest $K$. The optimum is conditional on this allocation model. The two criteria coincide only under this homogeneous model; whether heterogeneous candidate validity separates them is not tested here.

The empirical P6 comparison extends beyond this stylized two-way allocation. Compare objective-only selection, fixed-period controls and adaptive diagnosis under injected faults and drift; charge primary trials, confirmation attempts, diagnosis, online proposal computation, infeasible proposals, recovery, samples, instrument time and human work to the same declared resource vector. Reference materials, blanks and controls provide observations of apparatus state; when a prespecified residual or control-chart trigger fires, records since the last valid check are quarantined, calibration is repeated and the selector resumes on a segmented history. Measure actual ECYG, COR, CSU, unresolved outcomes, valid negatives mislabeled as faults and recovery burden. Physical preparation can fail because of execution limits, fluid properties or contamination; record these outcomes separately. Compare language-model-guided batch selection, amortized design policies and standard Gaussian-process BO on the same physical campaign. The broader proposition is falsified if diagnosis, confirmation-aware selection, language-model guidance or amortization does not improve the prespecified trade-off between confirmed evidence and invalid reported evidence per total cost, after accounting for invalidated proposals and follow-up. Policy comparisons require empirical evidence at the declared endpoint.

\subsection{I1: Preserve provenance completeness}

Provenance is an architectural invariant across all empirical propositions. Every record should link scientific intent, protocol version, device and calibration state, specimen lineage, commanded and realized conditions, raw observations, validity adjudication, trial status, human actions and claim-level status. A workflow satisfies I1 when another reader can reconstruct the realized experiment and the evidential path from the stored record. The invariant is violated when missing fields prevent reconstruction, even if the reported endpoint is favorable.

\subsection{I2: Preserve safety interlocks and human authority}

Mandatory exposure, collision, containment and biosecurity controls remain active when a selector, executor, model version, recovery policy or resource allocation changes. Because a physical interlock cannot stop harmful information release, I2 also requires that release decisions and accountable roles remain in the record. I2 is violated by any policy or budget change that disables an interlock, bypasses escalation or leaves a hazardous state without a safe stop.

\subsection{I3: Require cross-site replay schema compliance}

Cross-site replay uses a shared schema for units, versions, calibration, lineage, exceptions and claim links, while local adaptations and operator work are recorded separately. A second site should replay the declared workflow on held-out references or matched preparations and report endpoint agreement against a prespecified margin. Schema conformance and endpoint agreement are compliance metrics, not substitutes for scientific confirmation. I3 is violated by missing schema fields, unreported local changes or endpoint disagreement beyond the declared margin.

\subsection{Methodological discipline and claim calibration}

Self-driving-laboratory accounts stress that autonomy claims depend on what the platform can reach, measure and recover from, and that surrounding human work is part of the system \citep{seifrid2022}. The propositions assume that the endpoint is measurable, the resource boundary can be matched and confirmation can be specified; exploratory campaigns should state when one cannot. Methodological rigor requires guarding against characteristic failure modes where operational proxies substitute for verified physical evidence: an apparent speedup reverses when valid trials are replaced by confirmed claims; omitted failed attempts inflate reported reliability rates; benchmark leaderboard rankings are conflated with independent experimental replication; protocol compilation is mistaken for physical execution fidelity; valid negatives are mislabeled as execution faults; simulation rollouts are treated as physical measurements; and uncalibrated safety monitors are reported as formal guarantees. Explicitly reporting the proxy, its associated denominator, and untested execution steps preserves auditability and scientific reproducibility.

\section{Conclusions}\label{sec:conclusions}

Embodied AI for science connects operational, epistemic and evidential closure around one partially observed physical experiment. The reviewed robot-learning work predominantly evaluates control and prediction, whereas experiment-selection work evaluates allocation and response. Scientific claims become interpretable when these results are connected to realized conditions, audit-adjudicated validity and claim-specific confirmation.

Five conclusions follow from the evidence examined.

\begin{enumerate}
\item \emph{The scientific question should determine the method.} Property optimization, response mapping, hypothesis discrimination, sequential processing and adaptive observation have different terminal losses. Bayesian optimization, active learning, experimental design and control are therefore conditional choices: a study should name the terminal decision first and select the family that optimizes it. Language-model planners add proposal and tool interfaces, but their uncertainty calibration remains a separate evaluation task that a study must report \citep{chaloner1995,f2025anubis,f2024scienceagentbench}.
\item \emph{Realized conditions and audit-adjudicated validity are part of the model.} Commands do not determine specimen state or what an instrument observes. A model that distinguishes commanded, realized and observed quantities, audit-adjudicated validity $\hat v_t$ and trial status $c_t$ can represent competing specimen, apparatus and measurement explanations; separating them still requires discriminating sensors, controls or interventions. A command-conditioned model cannot separate delivery from response when the same command yields different realized volumes (Section~\ref{sec:formulation}).
\item \emph{General robot competence is an enabling result.} Vision-language-action policies, learned world models and tactile or force representations expand what can be physically attempted, but their reported endpoints are task success and prediction quality. Their suitability for a laboratory operation depends on delivered quantities, sample integrity and assay validity, which require operation-specific physical evaluation \citep{pi0,hafner2025dreamerv3,quijano2024pipetting}. Contact models illustrate the distinction: representations transfer across sensors and tasks \citep{higuera2025sparsh,feng2025anytouch}, yet a laboratory needs a calibrated force readout against a force--torque reference across wear, because contact is the realized condition of a Prepare step and an uncalibrated embedding cannot distinguish a valid insertion from a near-miss that consumes a sample (Section~\ref{sec:tactile-fm}). The consequence for evaluation is therefore specific: a tactile or force model should be counted as a laboratory instrument only when its readout is checked against a calibrated reference over the operating life of the sensor, and when it separates valid insertions from near-misses on held-out trials.
\item \emph{Application evidence supports claim-specific conclusions.} Across chemistry, materials, biology, microscopy and field science, workflows differ in whether they reach execution, valid measurement or confirmation, and in how much human work completes the loop. The A-Lab record illustrates why a claim must carry its boundary: the original report describes 36 compounds obtained from 57 targets; the 2026 correction finds the original conclusion correct for 36 of 40 reported successes, with four compounds inconclusive after reanalysis, and narrows novelty to the prediction platform \citep{alab2023,alabcorr2026}. Heterogeneous endpoints are therefore interpreted within their declared workflow and resource boundaries.
\item \emph{Evaluation must separate units and attribute gains.} Actions, valid experiments, campaigns and confirmed claims have different denominators; a selector-by-executor design separates allocation from execution; and confirmation requires evidence not used for selection \citep{volk2024}. The same separation applies inside the selector. Batch and language-model-guided proposals are evaluated on regret or predictive improvement \citep{gonzalez2016batch,liu2024llambo,f2024biodiscoveryagent}, while a laboratory additionally reports the probability that a proposed design is feasible and valid after preparation, ECYG, COR and CSU, with invalid attempts and human work charged to the same declared budget (Sections~\ref{sec:llm-bo} and~\ref{sec:evaluation}). The P6 ideal-confirmation yield (ICY) is a counterfactual true-yield benchmark for the opportunity cost of trial allocation, not an operational confirmed-claim count; actual status-confirmed yield is reported by ECYG and claim-level adjudication.
\end{enumerate}

The 28-workflow evidence map (Appendix~\ref{sec:appendix}), covering 26 physical workflows and two software-boundary components, grounds this synthesis across representative implementations; claim-level evaluation remains attached to each indexed claim object in $\mathfrak C_t$, with $U_t^{\mathrm{claim}}$ denoting its index set.

The record-linked evidence cycle is Record $\rightarrow$ Diagnose $\rightarrow$ Calibrate $\rightarrow$ Confirm, treated here as analytic functions with feedback. The research agenda tests whether history-aware state, recovery, transfer, validated capability expansion, action-aware models and budget-aware selection increase valid observations and independently assessed findings under matched costs. Provenance completeness, safety interlock invariance and cross-site replay schema compliance remain architectural and reporting invariants across those comparisons.

\appendix
\begingroup
\widowpenalty=1000
\clubpenalty=1000
\brokenpenalty=1000

\section{Literature selection and evidence interpretation}

The literature corpus was assembled through domain-targeted retrieval spanning embodied intelligence, AI for science, self-driving laboratories, robot learning, experimental design, adaptive scientific instruments, autonomous discovery agents, and laboratory infrastructure safety. Primary sources include peer-reviewed journal articles, conference proceedings, preprints, and official publisher repositories. Bibliographic fields were checked against Crossref, OpenAlex, arXiv, and publisher records where available. The original search strings, database hit counts and screening exclusions were not recorded reproducibly; subsequent metadata verification is not a reconstruction of that selection process. This is a structured narrative review, not a systematic review or an exhaustive field census.

\begin{figure*}[p]
\centering
\includegraphics[width=\textwidth,height=0.88\textheight,keepaspectratio]{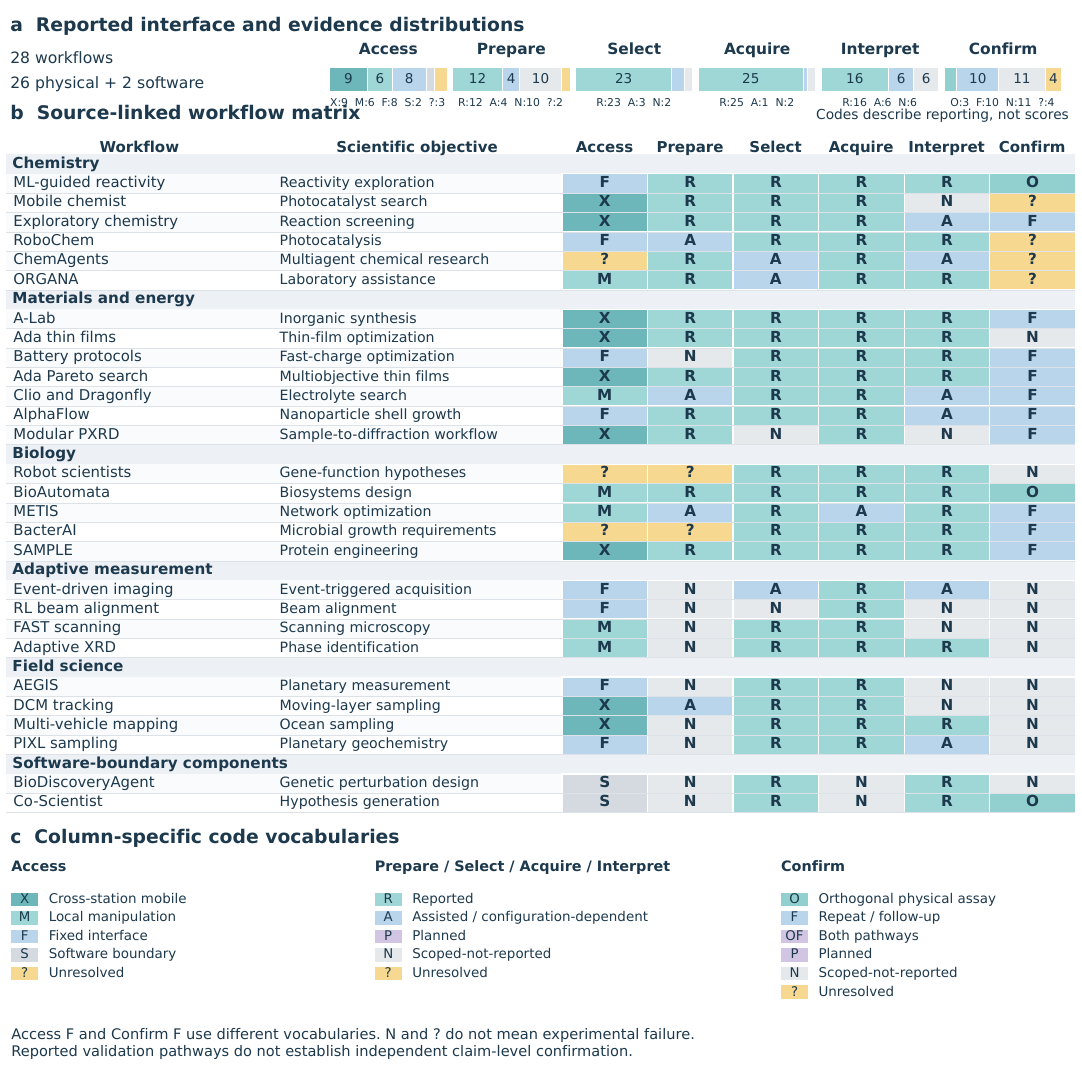}
\caption{Evidence atlas of 26 physical workflows and two software-boundary components. (a) Within-column code counts. (b) Every workflow, its recorded scientific objective and the six decision codes. (c) Separate vocabularies for Access, the four implementation columns and Confirm. Access uses X for cross-station mobile routing, M for local manipulation, F for fixed-interface or instrument-local control, S for software-boundary components and ? for unresolved routing. For Prepare, Select, Acquire and Interpret, R denotes reported implementation, A assisted or configuration-dependent operation, P planned capability, N scoped-not-reported and ? unresolved. In Confirm, O denotes an orthogonal physical assay, F repeat or follow-up testing, OF both, with P, N and ? carrying the same reported-status meaning. The Access F code (fixed interface) and the Confirm F code (follow-up) are different vocabularies. N and ? indicate absent or unresolved reporting, not experimental failure. The cohort includes two software-boundary components, so not all 28 rows are independent physical executions. Source rechecks retained the displayed codes, including ? where the relevant step remains unreported or unresolved. Full annotations and provenance are in the supplementary machine-readable dataset (\texttt{f07\_evidence\_map\_data.json} and the adjudication ledger) supplied with the source dataset.}
\label{fig:evidence-map}
\end{figure*}

\label{sec:appendix}

Application workflows were selected under three explicit inclusion criteria: (i) a defined scientific objective; (ii) physical embodiment interfacing with specimen or instrument states; and (iii) closed-loop adaptive feedback governing execution, acquisition, or experiment selection. Non-adaptive automation, purely computational methods, and general manipulation benchmarks were categorized as enabling capabilities or boundary comparators. Substantive claims are cited to original reports and author-published corrections \citep{alab2023,alabcorr2026}. Outcomes are interpreted at the demonstrated workflow level, preserving human operational responsibilities and follow-up requirements in the record.

Comparison fields cover scientific objective, physical interface, feedback-driven decision scope, and demonstrated boundary conditions. The cell codes were assigned by the authors from the cited sources, most carried over from an earlier coding pass; inter-rater agreement has not been assessed, so the map is a source-linked coding, not an independently validated measurement. They describe reported workflow properties rather than composite scores; unreported capabilities are designated as unrecorded rather than negative outcomes. For numerical findings, the sample population, operating duration, and evaluation metric are retained with the reported value.

For Figure~\ref{fig:evidence-map}, the 26 physical workflows and two software-boundary components are evaluated across six operational and decision dimensions: Access, Prepare, Select, Acquire, Interpret, and Confirm. Access is categorized by physical interface and routing modality: X = cross-station mobile routing (multi-station mobile manipulators, robotic transport, cross-instrument handover); M = single-station local manipulation (robotic handling, automated loading, localized actuation); F = fixed-interface or instrument-local control (inline flow cells, fluidic manifolds, microplate readers); S = software-boundary comparators (computational proposal components evaluated alongside physical follow-up); and ? = unresolved from available documentation. The operational decisions Prepare, Select, Acquire, and Interpret are coded as reported (R), assisted or configuration-dependent (A), planned (P), scoped-not-reported (N), or unresolved (?). Confirm categorizes the modality of reported follow-up evidence: O = orthogonal physical assay; F = repeat or follow-up testing; OF = both orthogonal and follow-up validation; P = planned; N = scoped-not-reported; and ? = unresolved. Claim-level verification ($\sigma_k = \mathrm{orthogonally\_confirmed}$) imposes the additional independence criteria formulated in Section~\ref{sec:formulation}. Complete machine-readable data, source locators, and adjudication ledgers are maintained in the accompanying supplementary files.


\endgroup

\FloatBarrier
\bibliographystyle{unsrtnat}

\bibliography{references}

\end{document}